\documentclass{article}
\PassOptionsToPackage{numbers, compress}{natbib}

\usepackage[final]{neurips_2022}

\usepackage[T1]{fontenc}    
\usepackage{hyperref}       
\usepackage{url}            
\usepackage{amsfonts}       
\usepackage{nicefrac}       
\usepackage{microtype}      
\usepackage{xcolor}         
\usepackage{framed}
\newtheorem{definition}{Definition}
\usepackage{algorithm}
\usepackage{algorithmicx}
\usepackage{algpseudocode}
\usepackage{bbm}
\usepackage[utf8]{inputenc}
\usepackage{subcaption}
\usepackage{amssymb}
\usepackage{graphicx,wrapfig,lipsum}
\usepackage[export]{adjustbox}
\usepackage{amsmath}
\usepackage{comment}
\usepackage{setspace}

\usepackage{booktabs,tabularx,enumitem,ragged2e}
\newcolumntype{Y}{>{\RaggedRight\arraybackslash}X}
\newcolumntype{T}{>{\raggedright\arraybackslash}p{4.5cm}}

\title{The Impact of Task Underspecification in \\ Evaluating Deep Reinforcement Learning}
\author{%
  Vindula Jayawardana \\
  MIT\\
  \texttt{vindula@mit.edu} \\
  \And
  Catherine Tang \\
  MIT\\
  \texttt{cattang@mit.edu} \\
  \And
  Sirui Li \\
  MIT\\
  \texttt{siruil@mit.edu} \\
  \AND
  Dajiang Suo \\
  MIT\\
  \texttt{djsuo@mit.edu} \\
  \And
  Cathy Wu \\
  MIT\\
  \texttt{cathywu@mit.edu} \\
}
\begin{document}

\maketitle

\begin{abstract}
Evaluations of Deep Reinforcement Learning (DRL) methods are an integral part of scientific progress of the field. Beyond designing DRL methods for general intelligence, designing task-specific methods is becoming increasingly prominent for real-world applications. In these settings, the standard evaluation practice involves using a few instances of Markov Decision Processes (MDPs) to represent the task. However, many tasks induce a large family of MDPs owing to variations in the underlying environment, particularly in real-world contexts. For example, in traffic signal control, variations may stem from intersection geometries and traffic flow levels. The select MDP instances may thus inadvertently cause overfitting, lacking the statistical power to draw conclusions about the method's true performance across the family. In this article, we augment DRL evaluations to consider parameterized families of MDPs. We show that in comparison to evaluating DRL methods on select MDP instances, evaluating the MDP family often yields a substantially different relative ranking of methods, casting doubt on what methods should be considered state-of-the-art. We validate this phenomenon in standard control benchmarks and the real-world application of traffic signal control. At the same time, we show that accurately evaluating on an MDP family is nontrivial. Overall, this work identifies new challenges for empirical rigor in reinforcement learning, especially as the outcomes of DRL trickle into downstream decision-making.

\end{abstract}

\section{Introduction}
Deep reinforcement learning research has progressed rapidly in recent years, achieving super-human level performance in many applications. At the core of DRL research lies the need to engage in rigorous experimental design for conducting evaluations. The lack of rigorous experimental design could lead to profound implications. Researchers may inadvertently mislead themselves and draw incorrect conclusions about DRL including what factors contribute to the success of a method~\cite{engstrom2020implementation, ilyas2018closer}, what factors make the results of a method reproducible~\cite{henderson2018deep}, or whether a method successfully solves one task~\cite{whiteson2011protecting} or multiple tasks~\cite{agarwal2021deep, jordan2020evaluating}. As evidenced by these findings, strong empirical rigor is crucial for research progress as it allows the research community to confidently assess the current state of the field.

The real world induces many complexities for control tasks, and one major complexity is the existence of multiple instances of the same task. Consider the case of a traffic signal control task where the goal is to design a signal control strategy for an intersection. To reliably claim a DRL method solves the traffic signal control problem, one needs to show that the proposed method sufficiently works for a considerable majority of the signalized intersection instances~\cite{attendlight}. We see this requirement of evaluating on a family of instances as an emerging requirement in general, not just limited to traffic signal control, specifically as DRL trickles into real-world applications. We refer to such evaluations as assessing the \textit{algorithmic generalization} of DRL methods \textit{within a task}. 

However, many studies that evaluate algorithmic generalization of DRL methods within a task ignore this requirement. In traffic signal control, the use of a few select intersection instances is popular for evaluations~\cite{ault2021reinforcement, Wei_2019, interpretable_TSC, mas, wei2018intellilight}. Similar discrepancies can also be seen in other application areas such as in autonomous driving~\cite{vinitsky2018benchmarks, jayawardana2022learning}, healthcare~\cite{ Zhao2009ReinforcementLD}, and in chemistry~\cite{Zhou2017OptimizingCR}. Such practices may fail to guard against evaluation overfitting~\cite{whiteson2011protecting}, in which the DRL method under evaluation records misleadingly high performance by overspecializing to the evaluation instance. As DRL trickles into real-world implementations and critical use cases, not having such practices can have serious implications, ranging from the opportunity cost of employing lower quality methods to hurting confidence in the field.

\begin{figure*}[h]
\centering
\begin{subfigure}{0.19\linewidth}
  \centering
  \includegraphics[width=\linewidth]{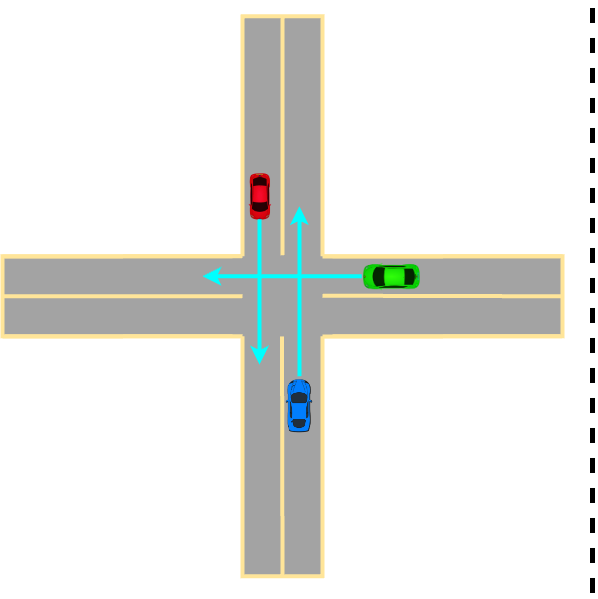}
  \caption{Default MDP}
  \label{int-example-a}
\end{subfigure}
\begin{subfigure}{0.79\linewidth}
  \centering
  \includegraphics[width=\linewidth]{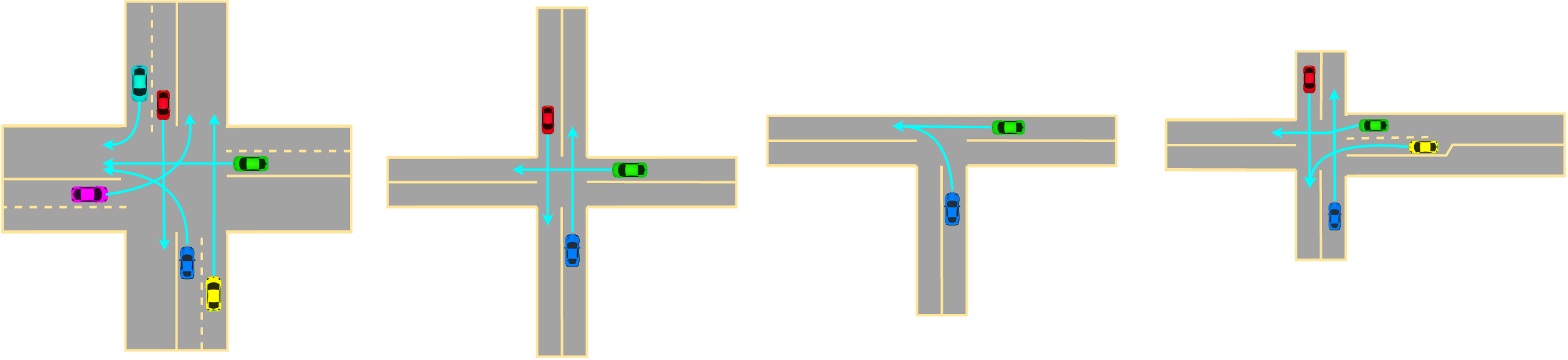}
  \caption{MDP Family}
  \label{int-example-b}
\end{subfigure}
\caption{An illustrative example of a). a default point MDP vs b). a family of point MDPs specified using some environment parameters $\phi$ for evaluations of traffic signal control. }
\label{int-example}
\end{figure*}

To formalize, consider the traffic signal control task, which is typically represented as a single MDP (Figure~\ref{int-example-a}). At the same time, the task is \textit{underspecified} and could easily be represented by a family of MDPs (Figure~\ref{int-example-b}), each specified using some environment parameters $\phi$ such as the number of lanes, turn configurations, and traffic inflow levels. Standard evaluation practice for assessing algorithmic generalization of DRL methods within the task of traffic signal control is to train and evaluate DRL methods on a \textit{point MDP} (or a subset of MDPs) chosen from an implicit family of MDPs (e.g., selecting a specific $\phi$). We refer to such evaluation approaches as \textit{point MDP-based evaluations}\footnote{We note that \textit{point MDP-based evaluations} are not just limited to MDPs but applicable to all variants of MDP like partially observable MDPs. We use \textit{point MDP} as a general terminology to refer to all such cases.}. Unfortunately, such point MDPs often appear to be selected arbitrarily or in a way that inadvertently simplifies the task. It is well known that the performance of a DRL method is sensitive to the underlying MDP~\cite{reda2020learning, henderson2018deep, andrychowicz2020matters}. In this work, \textbf{we hypothesize that the coupling of an arbitrary selection of point MDPs for evaluation and the general sensitivity of DRL methods can result in a substantial error in the true performance of DRL methods} over the implicit MDP family of a task. This, in turn, harms empirical rigor in DRL. We refer this phenomenon as the \textit{task underspecification problem} in DRL evaluations.

In this work, we take a deep dive into this matter. Our main contribution is identifying an emerging problem motivated by the real-world application of RL: when a point MDP is used in evaluating DRL methods within a task, the point MDP does not exhibit adequate statistical power to draw conclusions about the corresponding MDP family. We demonstrate that this phenomenon exists not only in real-world applications such as traffic signal control but also in standard control benchmarks, which indicates that it may be prevalent across the field. Our methodology consists of augmenting DRL evaluations with appropriately parameterized families of MDPs. We show that in comparison to applying DRL methods to the point MDP, evaluating the MDP family often yields a substantially different relative ranking of methods, which could lead researchers to draw incorrect conclusions about the performance of some methods over others. We demonstrate this phenomenon experimentally in a case study of traffic signal control. We show that DRL methods which were originally reported as outperforming traditional traffic signal control methods, significantly underperform when a family of MDPs is used for evaluation, causing a substantial change in the ranking of methods.  

Unlike standard DRL benchmark suites, in which the suite designer has full control over how many MDP instances are included in the suite~\cite{Kaiser2020ModelBasedRL, Tassa2018DeepMindCS}, in real-world applications, the domain dictates how many point MDPs are in a task. Whereas benchmark suites often have dozens of MDP instances, we estimate that a task can easily have hundreds or thousands of point MDPs. This thus illuminates new computational and reporting obstacles for evaluating DRL methods within a task. We provide some initial studies on the statistical power of evaluating DRL methods by sampling from the MDP family and use of performance profiles for reporting and conclude that it is a nontrivial problem owing to the sensitivity of today’s DRL methods.

\textbf{\textit{Important distinctions:}}
To put our contributions into the context of standard DRL practices, in this work we neither address the evaluations of algorithmic generalization across task families (standard DRL evaluations using task suites) nor evaluations of policy generalization within a task family (as addressed in multi-task, robust RL). Second, we focus on problems where a separate individual DRL \textit{model} can be trained for each MDP (\textit{i.e.,} to achieve overall better performance). Another parallel line is in which a single agent is designed to perform well on all MDPs of the family (\textit{e.g.,} multi-task learning in robotics). However, in the scope of this work, we do not consider such problems. 

In the following sections, unless otherwise stated, by \textit{evaluation}, we mean evaluations in DRL when evaluating algorithmic generalization of DRL methods within a task.


\section{Shortcomings of Point MDP-based Evaluations}
\label{pitfalls}

We illustrate the shortcomings of point MDP-based evaluation by considering an experiment derived from popular DRL benchmarks. We use three popular control tasks (Pendulum, Quad~\cite{Inala2020SynthesizingPP}, and Swimmer) as example underspecified tasks. For each task, we augment the nominal task to induce a family of MDPs. A summary of the tasks is given in Table~\ref{task-mdp}, along with their augmentations. For each task family, we then specify five point MDPs. More details of the tasks and their corresponding modifications are given in the Appendix~\ref{example-task-detailed}. We consider three popular DRL algorithms (PPO~\cite{Schulman2017ProximalPO}, TRPO~\cite{Schulman2015TrustRP} and TD3~\cite{Fujimoto2018Addressing}) for evaluation on the tasks. We have verified that each modified point MDP is not under-actuated and that given actuator limits, all are solvable. 

\begin{center}
\noindent
\captionof{table}{Control task and the MDP families. } 
\label{task-mdp}
\begin{tabularx}{\textwidth}{ @{} c X T }
  \toprule
  Task & Task description & MDP family \\
  \midrule
  Quad & 
  \begin{tabular}[t]{ @{\makebox[0.5em][l]{\textbullet}} p{\dimexpr\linewidth-0.9em} p{\dimexpr\linewidth-0.4em} }
    2D quadcopter in an obstacle course.\\
    \textbf{Goal}: maneuver a 2D quadcopter through an obstacle course using its vertical acceleration as the control action. 
  \end{tabular} 
  & 
  \begin{tabular}[t]{ @{\makebox[0.5em][l]{\textbullet}} p{\dimexpr\linewidth-0.9em} p{\dimexpr\linewidth-0.4em} }
    MDPs with varying obstacle course lengths (upper obstacle length and lower obstacle length).
    \end{tabular} \\
  \midrule
  Pendulum & 
  \begin{tabular}[t]{ @{\makebox[0.5em][l]{\textbullet}} p{\dimexpr\linewidth-0.9em} p{\dimexpr\linewidth-0.4em} }
    A pendulum that can swing.\\
    \textbf{Goal}: swing the pendulum upright.
  \end{tabular} 
  & 
  \begin{tabular}[t]{ @{\makebox[0.5em][l]{\textbullet}} p{\dimexpr\linewidth-0.4em} p{\dimexpr\linewidth-0.4em} }
    MDPs with varying masses and lengths of the pendulum.
  \end{tabular} \\
  \midrule
  Swimmer & 
  \begin{tabular}[t]{ @{\makebox[0.5em][l]{\textbullet}} p{\dimexpr\linewidth-0.4em} p{\dimexpr\linewidth-0.4em} }
    MuJoCo 3-link swimming robot in a viscous fluid. \\
    \textbf{Goal}: make the robot swim forward as fast as possible by actuating the two joints. 
  \end{tabular} 
  & \begin{tabular}[t]{ @{\makebox[0.5em][l]{\textbullet}} p{\dimexpr\linewidth-0.4em} p{\dimexpr\linewidth-0.4em} }
    MDPs with varying capsule sizes for the segments comprising the robot swimmer.
    \end{tabular} \\
  \bottomrule
\end{tabularx}

\end{center}

\textbf{Observation 1: Reporting evaluations based on point MDPs can be misleading}.

Previous works commonly use three criteria for selecting a point MDPs for evaluations. 1) \textit{random MDP}: a random MDP is used to model a somewhat arbitrary selection of an MDPs from the family (\textit{e.g.,} a random selection of a signalized intersection for training a traffic signal control agent~\cite{ault2021reinforcement}) 2) \textit{generic MDP}: an MDP that represents the key characteristics of the MDPs in the family (\textit{e.g.,} the use of a generic cancer progression model for training a chemotherapy designing agent~\cite{Zhao2009ReinforcementLD}) and 3) \textit{simplified MDP}: an MDP that simplifies the modeling (\textit{e.g.,} zero-error instrument modeling when training an agent for chemical reaction optimization~\cite{Zhou2017OptimizingCR}).

\begin{figure*}[!h]
\centering
\begin{subfigure}{0.33\linewidth}
  \centering
  \includegraphics[width=\linewidth]{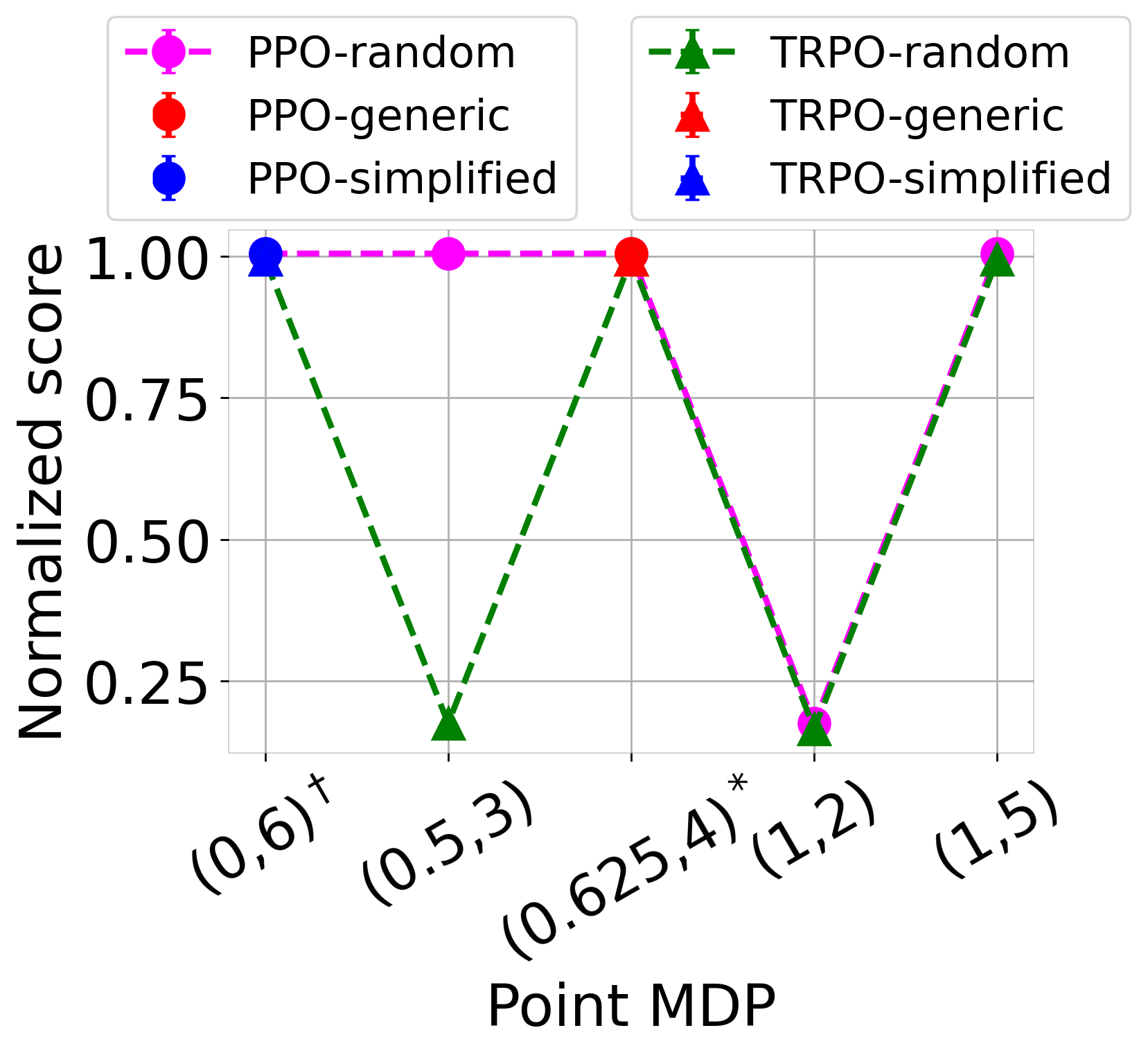} 
  \caption{Quad}
  \label{fig11a}
\end{subfigure}
\begin{subfigure}{0.31\linewidth}
  \centering
  \includegraphics[width=\linewidth]{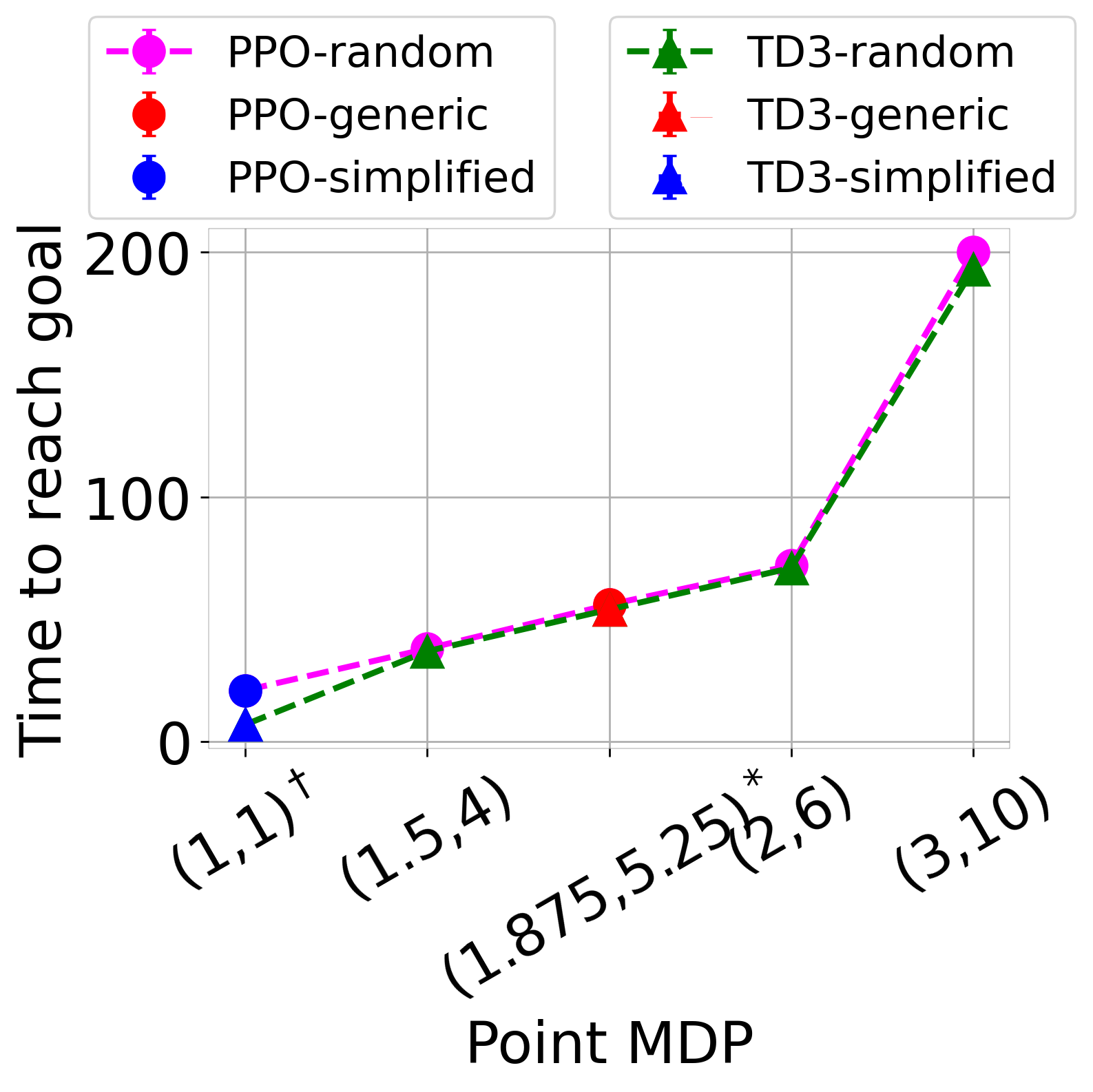} 
  \caption{Pendulum}
  \label{fig11b}
\end{subfigure}
\begin{subfigure}{0.32\linewidth}
  \centering
  \includegraphics[width=\linewidth]{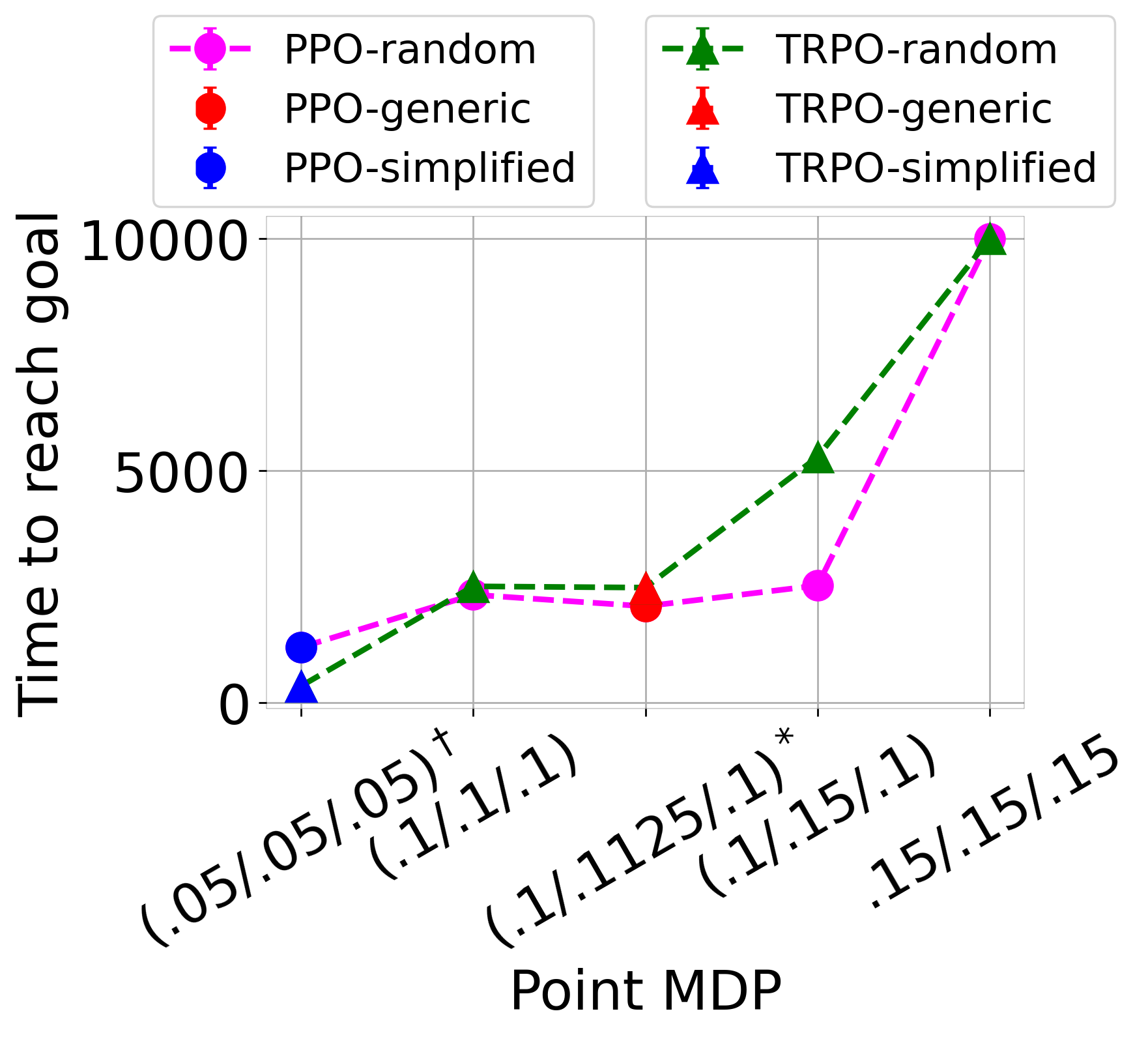} 
  \caption{Swimmer}
  \label{fig11b}
\end{subfigure}
\caption{Performance comparison of point MDP-based evaluations of the three control tasks. The x-axis represents the five point MDPs. For Quad, performance is a normalized score of the distance the quadcopter traveled before crashing or reaching the goal with respect to the total distance required to travel (higher the better). For the pendulum, it is the time to swing the pendulum upright (lower the better), and for the swimmer, it is the time to reach the goal (lower the better). }
\label{Issue-one}
\end{figure*}

In Figure~\ref{Issue-one}, we report the evaluations of the DRL methods on each point MDP described in Table~\ref{task-mdp} for the three control tasks. That is, each method is trained and evaluated on each individual point MDP. Given an MDP family, all child MDPs are considered possible random MDPs. For each task, $\star$ denotes the generic MDP, and $\dagger$ represents the simplified MDP. The simplified MDP is chosen as the ``easiest'' MDP, i.e., simplifies the transition dynamics. The generic MDP is the parametric mean MDP of the other four MDPs in the family.

We observe a few interesting phenomena (1) the same method yield significant differences in performance under different point MDPs for all three tasks, (2) simplified MDPs generally achieve better performance than other point MDPs in the family, and (3) comparing DRL methods based on point MDPs can provide conflicting conclusions (e.g., conflicting relative performance benefits) based on which point MDP is used for evaluations. For example, for Quad, point MDP (0,6) indicates both PPO and TRPO are equally well-performing. However, under point MDP (0.5,3), we see that PPO outperforms TRPO with approximately 0.75 point difference. These observations highlight the variability of point MDP-based evaluations and the uncertainties involved. Such evaluations could mislead the community to incorrectly conclude that one method is better than another and thereby hinder the scientific progress of the field. 

\textbf{Observation 2: DRL training can be sensitive to the selected point MDP properties}. It is generally known that DRL training is sensitive to the underlying MDP. Therefore, selecting a point MDP from a family can demonstrate a training impact that does not generalize to other MDPs in the family. Subsequently, the performance of the DRL method could be overestimated or underestimated.

\begin{figure*}[!h]
\centering
\begin{subfigure}{0.32\linewidth}
  \centering
  \includegraphics[width=\linewidth]{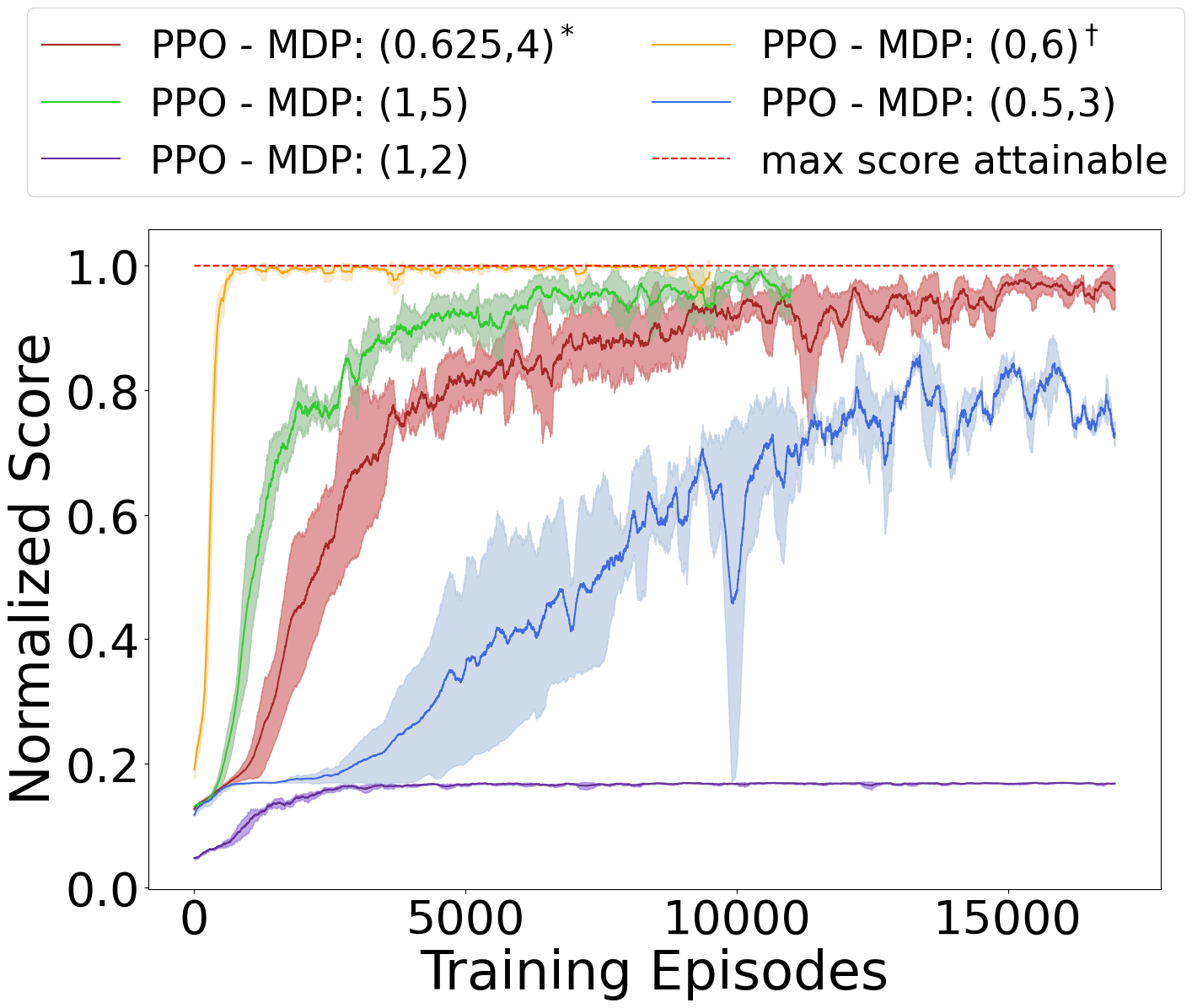}
  \caption{Quad}
  \label{quad-1-a}
\end{subfigure}
\begin{subfigure}{0.32\linewidth}
  \centering
  \includegraphics[width=\linewidth]{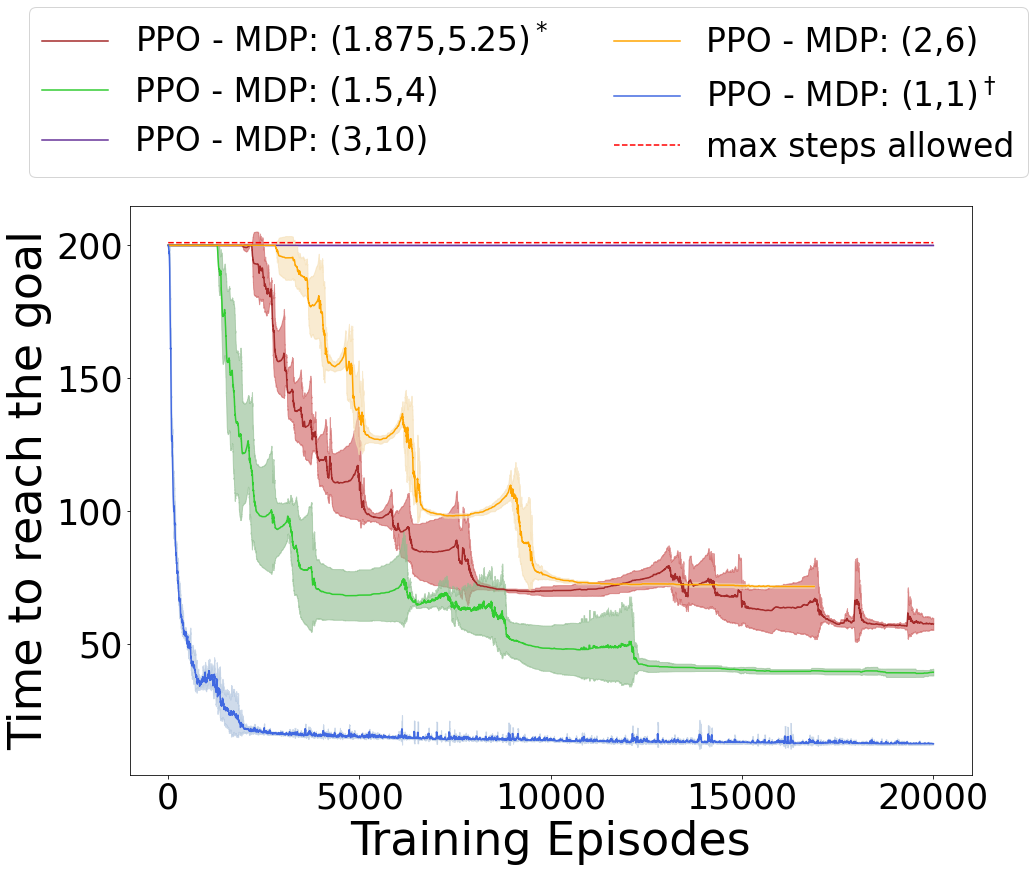}
  \caption{Pendulum}
  \label{pen-1-b}
\end{subfigure}
\begin{subfigure}{0.32\linewidth}
  \centering
  \includegraphics[width=\linewidth]{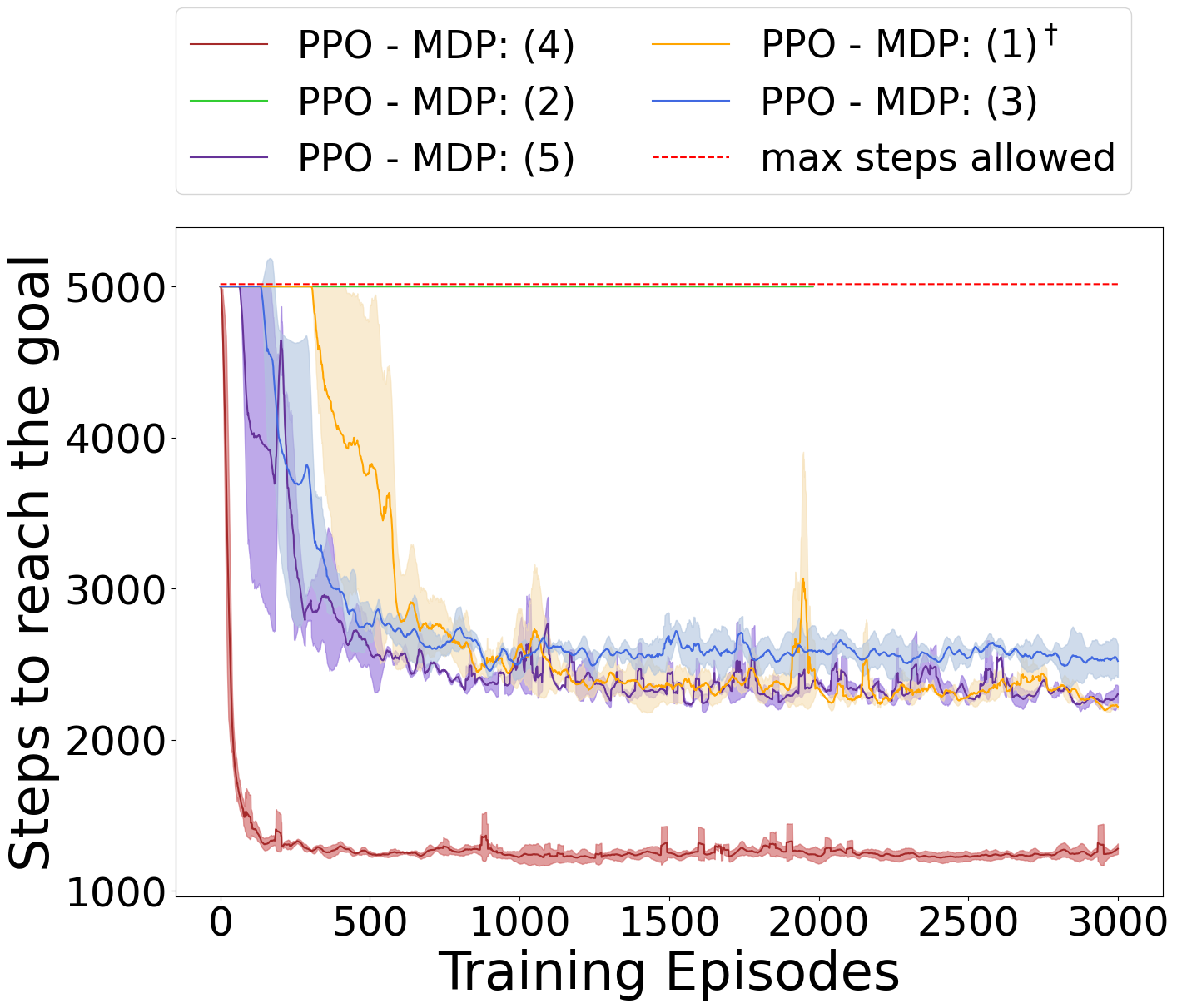}
  \caption{Swimmer}
  \label{swim-1-c}
\end{subfigure}
\caption{Training progress of each task for different point MDPs. For Quad, performance is a normalized score that indicates the distance the quadcopter traveled before crashing or reaching the goal with respect to the total distance required to travel (higher the better). In the pendulum, the training performance is measured as the time to swing the pendulum upright (lower the better). For the swimmer, the training performance is measured as the time to reach the goal (lower the better). }
\label{Issue-two}
\end{figure*}

We demonstrate this phenomenon using our three control tasks in Figure~\ref{Issue-two} when using the PPO algorithm for training\footnote{Under some point MDPs, not all parallel runs succeed. For point MDPs where some runs succeed, we only plot the runs that succeed. For point MDPs where all runs fail, we plot them as it is. We fix the number of total training steps to 5M for swimmer and 2M for quad and pendulum. Curves are truncated for better visibility. }. First, in Quad (Figure~\ref{quad-1-a}), reaching the goal means achieving a normalized score of 1. However, only under a few of the point MDPs, the agent reaches the goal during training. Specifically, 2 out of 5 settings converge to a local minimum (crashing on the same obstacle as training progress). Similarly, in Figure~\ref{pen-1-b} for the pendulum, under one of the MDPs the agent does not achieve the goal (stuck at the 200 steps mark) while under the other four, the agent achieves the goal. Finally, we see similar behavior in Swimmer task in Figure~\ref{swim-1-c} where under one point MDP the agent fails to achieve the goal (stuck at the 5000 steps) during training and succeeds under others. Further examples of training complications under other DRL algorithms are given in Appendix~\ref{i2-more-examples}.  

We hypothesize one of the root causes of this phenomenon is the complexity differences in point MDPs. For example, in Quad, making the obstacle-free course narrower requires the DRL agent to explore some specific actions different from what it would otherwise need to explore when the course is wider. This effectively makes the training process harder for point MDPs with narrower paths. Another potential cause is MDP designers' over-fitting MDP design to point MDPs. For example, in the pendulum, the default reward function penalizes higher torque values while encouraging reaching the goal with a weighted composite reward. MDP designers may over-fit the weighting values to a selected point MDP which does not generalize to the entire family. This can result in point MDPs with higher pendulum mass failing if the weight on the torque limits is higher.

\section{Notation and Formalism}
\label{formalism}
Given a sequential decision-making task $T$ and a reinforcement learning method $R$ to be evaluated, we consider the setting where there are $M$ possible point MDPs in the family of MDPs. In general, despite the illustrative examples in the previous section, $M$ can be quite large. Let $N$ denote a computational budget in terms of the number of models that can be trained\footnote{Most cloud service providers charge users based on the time they use the services. Therefore, approximating the average number of models that can be trained given a pricing budget can be done with rough estimates of how long it takes to train one model on a single point MDP.}. We assume that $M \gg N$; that is, we cannot evaluate the given RL method on all MDPs in the family.

As we demonstrated in Section~\ref{pitfalls}, the performance of a DRL method $R$ can significantly depend on the choice of the point MDP. As suggested by previous work~\cite{agarwal2021deep}, it is therefore warranted to model the performance of $R$ as a real-valued random variable $X_{R}$. This means a normalized \textit{performance score $s_{R,i}$} for a given $R$ and a point MDP $i$ is a realization of the random variable $X_{R}$. We normalize point MDP performance scores by linearly rescaling scores based on a given baseline. For example, scores in Atari games are typically normalized with respect to an average human~\cite{mnih2015human, agarwal2021deep}.

Given a family of MDPs, a point MDP $i$ may be more important or common than another point MDP $j$. This is a common requirement in the real world where practitioners have predefined performance requirements, such as the performance of a signalized intersection in an urban area may be more important than a signalized intersection in a sub urban area in traffic signal control~\cite{CHODUR201687}. This \textit{importance score $p_{T,i}$} of a point MDP $i$ on task $T$ can be considered as a realization of a real-valued random variable $Y_{T}$. This means depending on the importance scores of each point MDP, a distribution can be generated for the MDP family. For $\tau \in \mathbb{R}^n$ where $\tau$ represents the point MDP context, we therefore define the point MDP distribution as $F(\tau) = P(Y_T)$. 

\textbf{Assumption}: For a given task $T$, we assume $p_{T,i}$ is given for each MDP $i$. 

\begin{definition}
The overall performance of a DRL method $R$ on task $T$ is defined as $E^{T}_{R} = \mathop{\mathbb{E}}[X_R] = \sum_{i=1}^{|U|}s_{R,i} p_{T,i}$ where $U$ is the set of point MDPs.
\label{overall-performance-def}
\end{definition}

However, obtaining $E^{T}_{R}$ is not always possible because of the budget constraint $M \gg N$. Therefore, a potential solution is to select a subset $V$ of point MDPs from the MDP family to perform an evaluation. Accordingly, the estimated performance of the method $R$ on task $T$ is $\hat{E}^{T}_{R} = \sum_{i=1}^{|V|}s_{R,i} p_{T,i}$. Clearly, if not careful, selecting different subsets can greatly affect the accuracy of the evaluations. 

An intuitive approach to selecting a subset of point MDPs is to assess the \textit{contribution $c_i$} of each point MDP $i$ to overall evaluation. Contribution depends both on the importance score, which is given, and performance score, which incurs a computational cost to evaluate, and can be defined as $c_i = s_{R,i} p_{T,i}$. Thus, we seek to find the set of point MDPs that has the highest contributions to the overall evaluation. Given that we wish to estimate the contribution of individual point MDPs without assessing the overall performance $E^{T}_{R}$, this poses a chicken-and-egg problem. In the subsequent sections, we propose approximation techniques that one can employ to identify a subset $V$ based on the approximated contributions. 
\section{Case Study: Traffic Signal Control}
\label{case-study}
To validate the shortcomings of point MDP-based evaluations, we consider an established benchmark that exhibits a large implicit family of MDPs describing a single task and wherein there could be significant real-world implications. In particular, we consider the evaluations of DRL methods on the traffic signal control task, leveraging the \textit{RESCO} benchmark~\cite{ault2021reinforcement}. We use six algorithms from RESCO in our case study, namely: \textit{IDQN}, \textit{IPPO}~\cite{interpretable_TSC}, \textit{MPLight}~\cite{mplight}, \textit{MPLight$^*$}, \textit{Fine-tuned Fixed time} and \textit{Max pressure}~\cite{VARAIYA2013177}. Further details can be found in Appendix~\ref{case-study-methods}. 

Naturally, traffic signal control should be considered on multiple intersection geometries and vehicle flow levels, hence a family of MDPs. We base our importance score $p_{T,i}$ of each intersection on the frequency of occurrence within a geographic region and the performance score $s_{R,i}$ as the normalized per vehicle average delay. Scores are normalized based on an untuned yet sufficiently performant fixed time controller baseline. The importance scores and the intersections used to build the point MDP distribution of the intersections are taken from Salt Lake City in Utah. We use 164 unique intersections and refer the reader to Appendix~\ref{intersection-dist} for more details on building this distribution. 

\begin{figure}[!h]
  \includegraphics[width=0.99\textwidth]{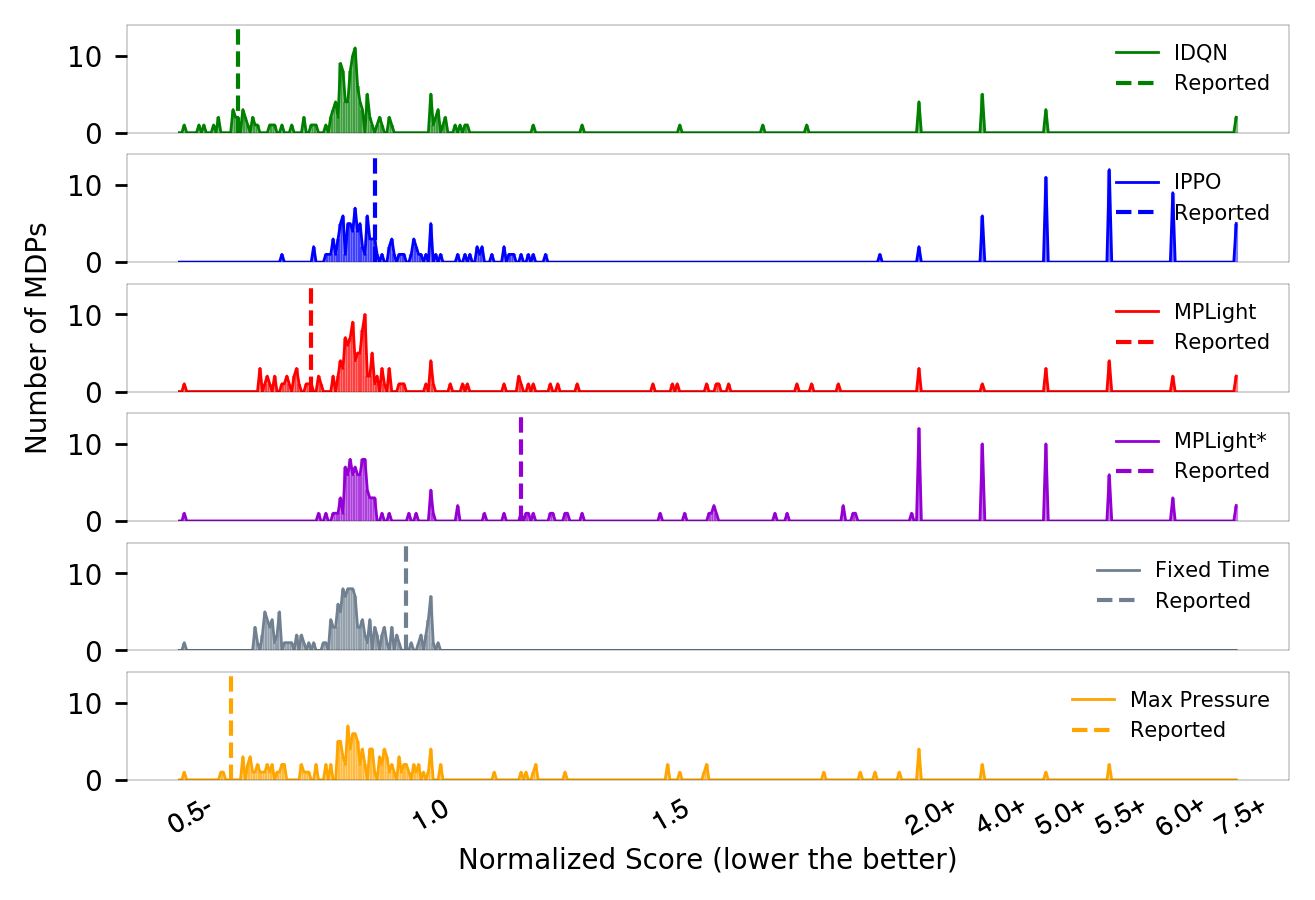}
  \caption{Performance vs the number of point MDPs that demonstrate the performance for all traffic signal control methods in RESCO. 164 unique point MDPs were considered for each method.} 
  \label{fig:DRL-performance-traffic-light}
\end{figure}

In Figure~\ref{fig:DRL-performance-traffic-light}, we report the performance of each of six methods on 164 unique point MDPs and the reported performance as per~\cite{ault2021reinforcement}\footnote{Reported performances are based on re-evaluations of the methods on Ingolstadt single intersection.}. First, we observe significant variations in the performance based on the point MDP used for evaluations. Specifically, all DRL methods demonstrate a significant variation while non-DRL methods demonstrate comparatively low yet considerable variations. Second, we observe that the reported performances in related literature can be significantly biased. As an example, IDQN, IPPO and MPLight performances are clearly overestimated. To quantitatively analyze the potential shortcomings, we denote the performance of each method over the entire MDP family in Figure~\ref{ts-results}.

\begin{minipage}[]{0.4\linewidth}
\includegraphics[width=1.0\linewidth, valign=T]{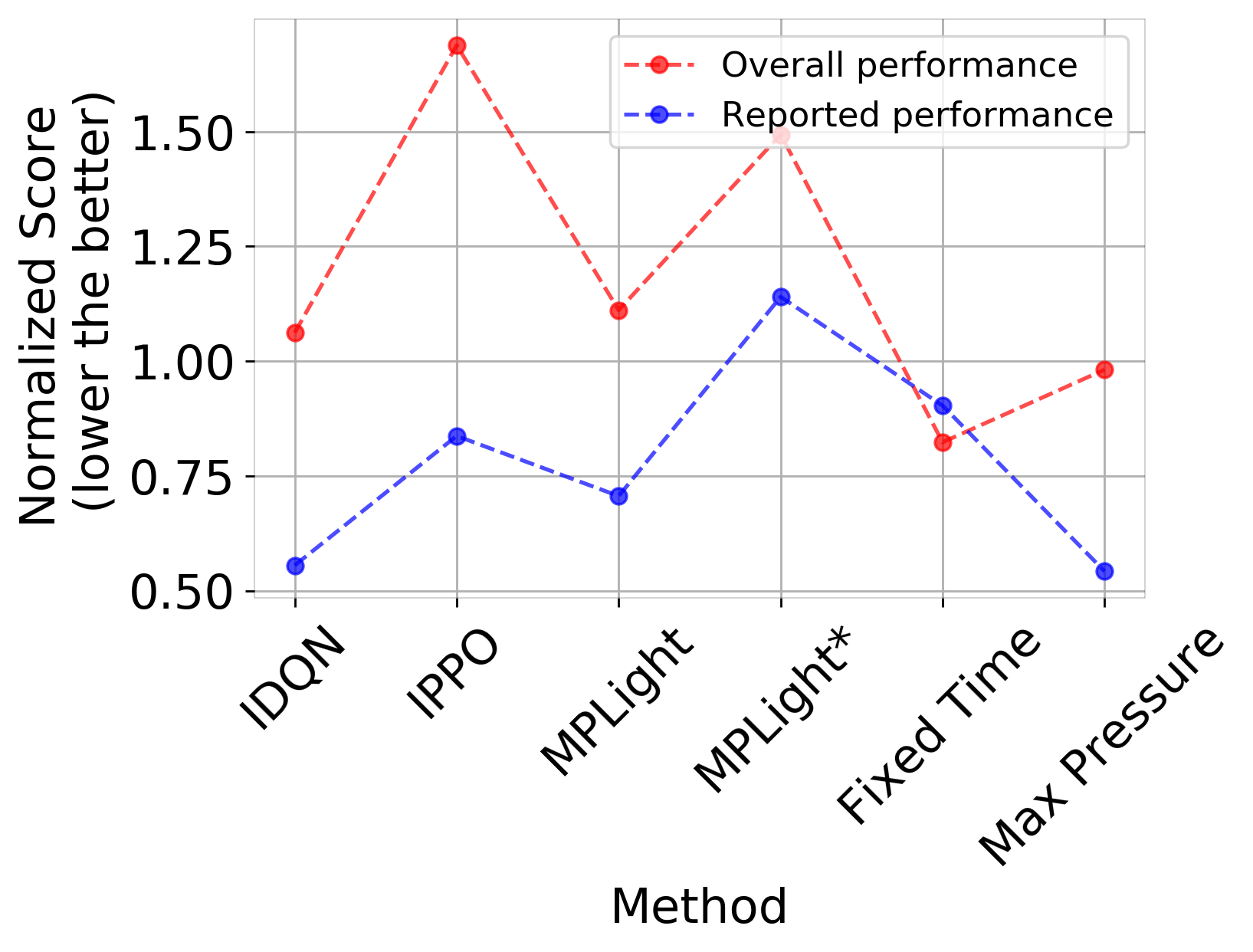}
\captionof{figure}{Overall performance (lower\\ the better)}
\label{ts-results}
\end{minipage}
\begin{minipage}[]{0.6\linewidth}
Interestingly, we see a significant result change from previously reported results. Although the Fixed time controller is regarded as an underperforming method as per reported performance in~\cite{ault2021reinforcement}, we find that under the MDP family-based evaluations, the non-RL Fixed time controller and Max pressure controller perform significantly better than all the four DRL controllers. We note that fine-tuning a non-RL Fixed time controller is simple enough that it does not pose a computational burden and can be done easily, even on a regular computer. While reported performances ranked IPPO as a well-performing model with a normalized score of 0.84, we see that, in fact, it is the lowest-performing method and that the revised normalized score is as high as 1.7 (even cannot
\end{minipage}
outperform an untuned fixed time controller). It is thus clear that point MDP-based evaluations can be misleading and may pose performance benefits that do not generalize to the MDP family. \footnote{Results reported in this work should not be illustrated as evidence against using DRL for traffic signal control and should only be used as evidence of shortcomings in point MDP-based evaluations. Further studies are encouraged to study the benefits of DRL for traffic signal control without the point MDP-based assumptions.} 

\section{Further Evidence on Shortcomings of Point MDP-based Evaluations}
\label{common-tasks}

\begin{figure*}[h]
\centering
\begin{subfigure}{0.32\linewidth}
  \centering
  \includegraphics[width=\linewidth]{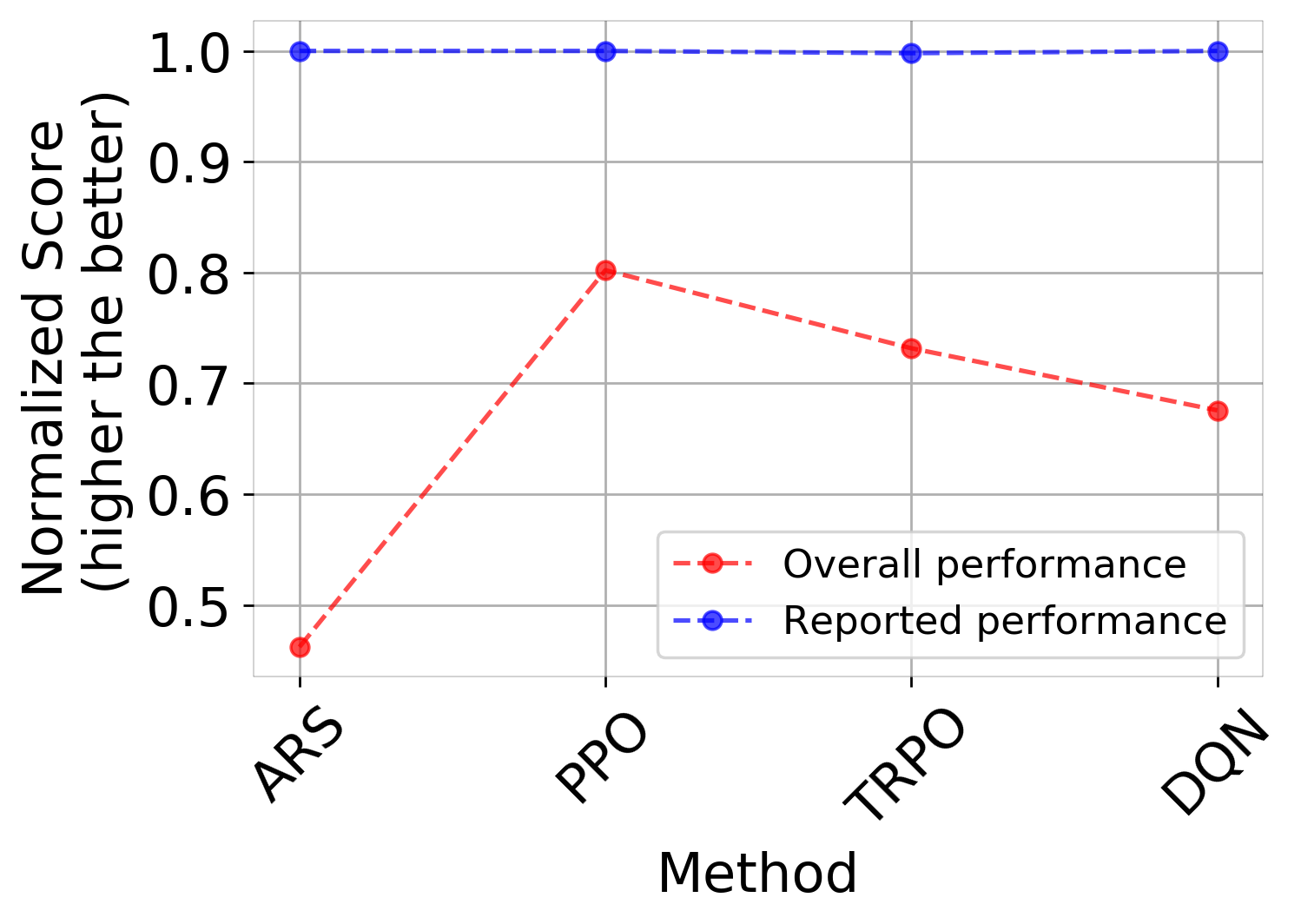}
  \caption{Cartpole (576 point MDPs)}
  \label{cartpole-a}
\end{subfigure}
\begin{subfigure}{0.32\linewidth}
  \centering
  \includegraphics[width=\linewidth]{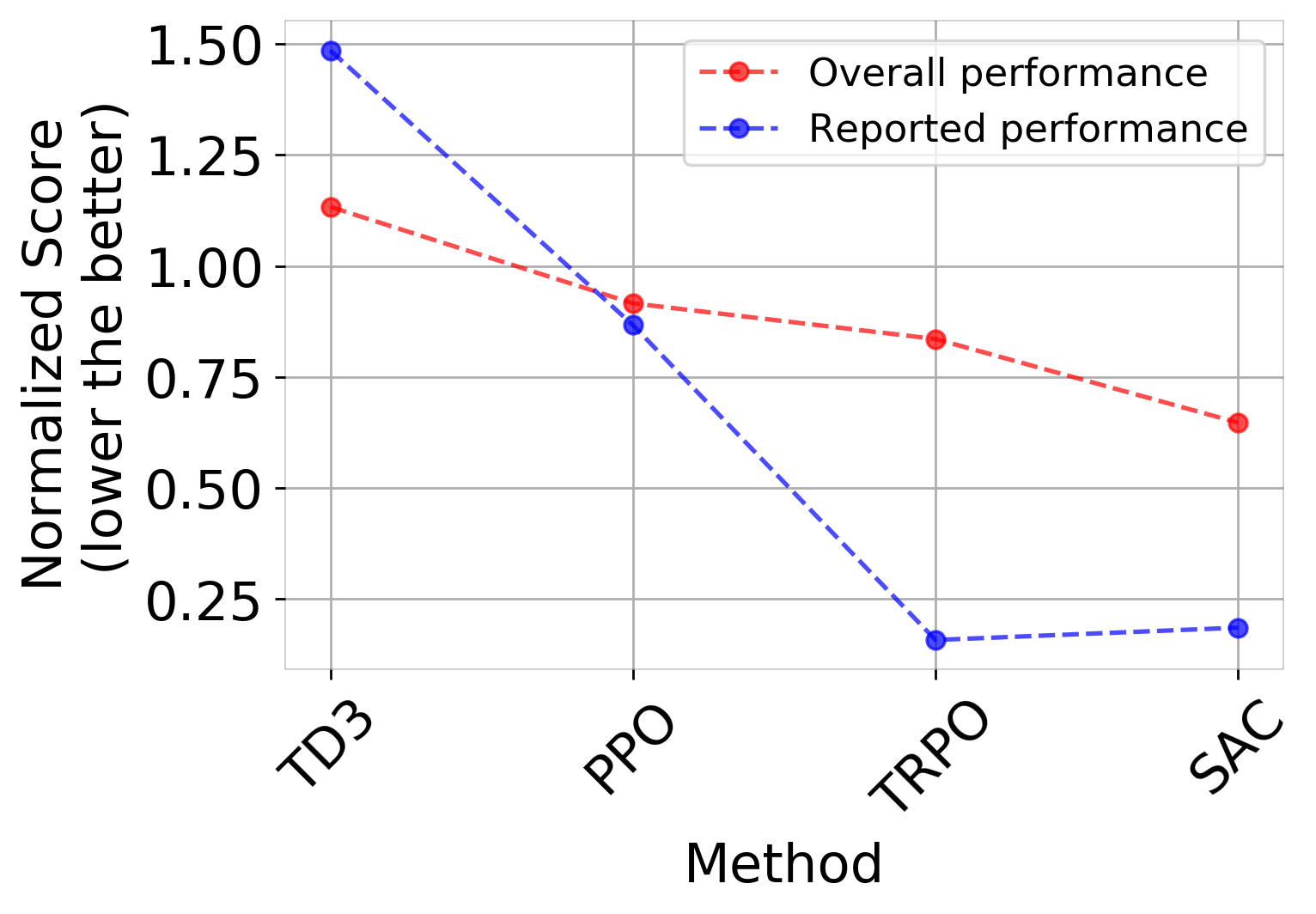}
  \caption{Pendulum (180 point MDPs)}
  \label{pendulum-a}
\end{subfigure}
\begin{subfigure}{0.32\linewidth}
  \centering
  \includegraphics[width=\linewidth]{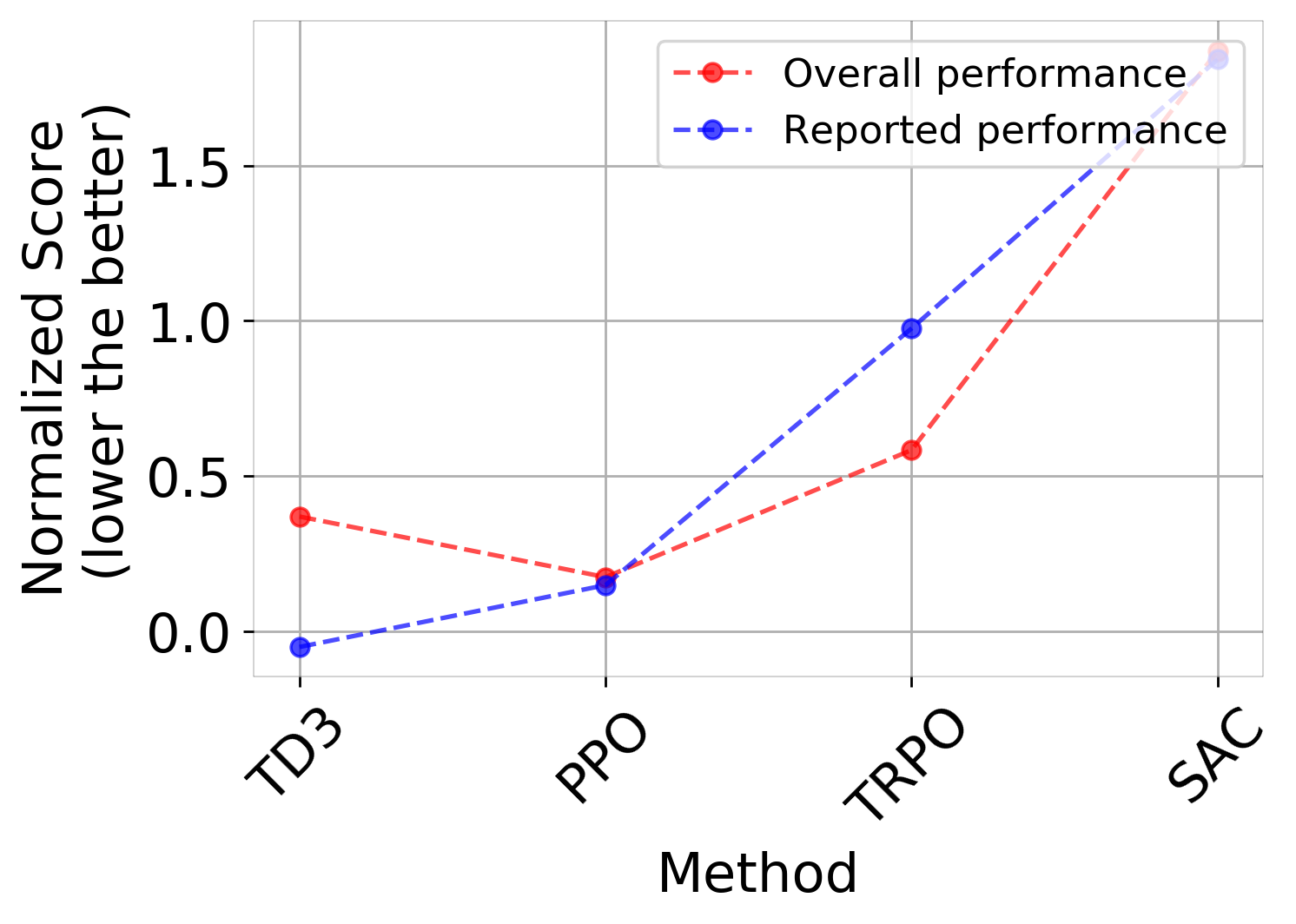}
  \caption{Half Cheetah (120 point MDPs)}
  \label{halfcheetah-a}
\end{subfigure}
\caption{Discrepancies between reported vs. overall performance of popular DRL methods in cartpole, pendulum and half cheetah when evaluated for algorithmic generalization within the task.}
\label{DRL-tasks-pitfalls}
\end{figure*}

To further validate the shortcomings of point MDP-based evaluations and the impact, we look at three popular DRL control tasks: cartpole (discrete actions), pendulum (continuous actions), and half-cheetah (continuous actions). For each task, we devise a family of MDPs as described in Table~\ref{drl-tasks-detailed-vv} in Appendix~\ref{drl-task-config}. Our MDP families consist of 576, 180, and 120 point MDPs for cartpole, pendulum, and half cheetah, respectively. Due to space limitations, we provide an in-depth analysis of performance variations in Appendix~\ref{drl_tasks_analysis} and only provide a summary of the analysis in this section. In Figure~\ref{DRL-tasks-pitfalls}, we show the significant discrepancies in overall performance and the reported performances. The reported performance of each task is measured by training and evaluating DRL methods on commonly used single point-MDP given in common benchmark suites.

We see significant result changes when evaluated on the point MDP family. For example, in cartpole (Figure~\ref{cartpole-a}), all reported performances achieve the best score of 1.0 while we see a significantly different overall performance. Although under reported performance, all four DRL methods for cartpole are ranked as equally well-performing, we see an interesting rank change as some methods  underperform when considering their overall performance. Similar insights can be seen in Pendulum (Figure~\ref{pendulum-a}) and in half cheetah (Figure~\ref{halfcheetah-a}). 

\section{Reliable Evaluations Within a Task}

In Section~\ref{pitfalls}, \ref{case-study} and~\ref{common-tasks}, we demonstrate the shortcomings of point MDP-based evaluations. In this section, we discuss the challenges that arise as a result of conducting MDP family-based evaluations. We present three main challenges and an initial set of recommendations to the research community as summarized in Table~\ref{summry-recommendatations}.

\begin{center}
\noindent
\captionof{table}{Summary of challenges and initial recommendations} 
\label{summry-recommendatations}
\begin{tabularx}{\textwidth}{ @{} T X }
  \toprule
  Challenge & Our recommendation\\
  \midrule
  Lack of benchmarks & 
  \begin{tabular}[t]{ @{\makebox[0.5em][l]{\textbullet}} p{\dimexpr\linewidth-0.7em} p{\dimexpr\linewidth-0.2em} }
    Create benchmarks that depict MDP families.\\
    Publish datasets of MDP families of control tasks including point-MDP distributions.\\
    Incentivize publication of such datasets and control task at leading conferences.
    \end{tabular} \\
  \midrule
  Large families of MDPs with limited computational budgets & 
  \begin{tabular}[t]{ @{\makebox[0.5em][l]{\textbullet}} p{\dimexpr\linewidth-0.9em} p{\dimexpr\linewidth-0.4em} }
    Adopt performance approximations using clustering and random sampling under a computational budget. \\
    Standardize the evaluations by making the selected point-MDPs public. 
  \end{tabular} \\
  \midrule
  Lack of emphasize on all point-MDP performances  & 
  \begin{tabular}[t]{ @{\makebox[0.5em][l]{\textbullet}} p{\dimexpr\linewidth-0.4em} p{\dimexpr\linewidth-0.4em} }
    Use performance profiles to show a detailed view of how overall performance changes with point MDPs \\
  \end{tabular} \\
  \bottomrule
\end{tabularx}

\end{center}

\subsection{Create data sets of MDP families to benchmark RL methods in evaluations}

The use of benchmark datasets for evaluations is well-established in supervised learning. From computer vision~\cite{dendorfer2020mot20} and biological data~\cite{biological-data} to natural language processing~\cite{peng-etal-2019-transfer}, standard datasets are being used for evaluations of deep learning methods. In DRL, the practice is different. In assessing algorithmic generalization of DRL methods within a task, instead of published datasets, researchers use standardized benchmark suites like RESCO for traffic signal control~\cite{ault2021reinforcement}, and Vinitsky et al. \cite{vinitsky2018benchmarks} for mixed autonomy traffic. 

However, we recognize that current state-of-the-art benchmarks alone are not sufficient to standardize the evaluations in DRL. The main limitation is that they only provide select point MDPs and do not consider the family of possible MDPs. As many real-world applications inherently demonstrate a family of MDPs, the current DRL standardization of evaluations may steer the research community in a vacuum, while realistically useful DRL methods may get rejected or not even developed. 

Therefore, our first recommendation is to create DRL benchmark suites which inherently demonstrate a requirement to incorporate a family of MDPs. However, which MDPs to include in a given family is task-dependent, and the expertise of domain experts may be needed. Second, we also encourage the research community to actively create MDP families for existing control tasks (e.g., different signalized intersections as traffic signal control MDP family) and publish them publically. Finally, we also encourage main artificial intelligence conferences with datasets and benchmark tracks like NeurIPS to encourage the community to publish such datasets and tasks and to include necessary check-ins in the paper submission checklists.

\subsection{Evaluate DRL methods on a family of MDPs instead of point MDPs} 

\label{approx_main-content}
In Section~\ref{pitfalls}, \ref{case-study} and \ref{common-tasks}, we demonstrated the shortcomings of point MDP-based evaluations. Therefore, our next recommendation is to encourage researchers to use families of MDPs instead of point MDPs in DRL evaluations. However, evaluating performance over an entire family of MDPs can often be computationally expensive to carry out in practice due to the large cardinality of the family. A solution is to conduct performance approximations. While more sophisticated methods like active learning approaches are possible, we resort to effective yet straightforward techniques to bolster the adoptability of the techniques within a wider community.

We present three techniques: (1) \textbf{M1}: random sampling with replacements from the point MDP distribution, (2) \textbf{M2}: random sampling without replacements, and (3) \textbf{M3}: clustering point MDPs using k-means and assigning probability mass of all point MDPs that belong to same cluster to its centroid. More details of each method and more analysis can be found in Appendix~\ref{performance approximations}. 

\begin{figure*}[h]
\centering
\begin{subfigure}{0.32\linewidth}
  \centering
  \includegraphics[width=\linewidth]{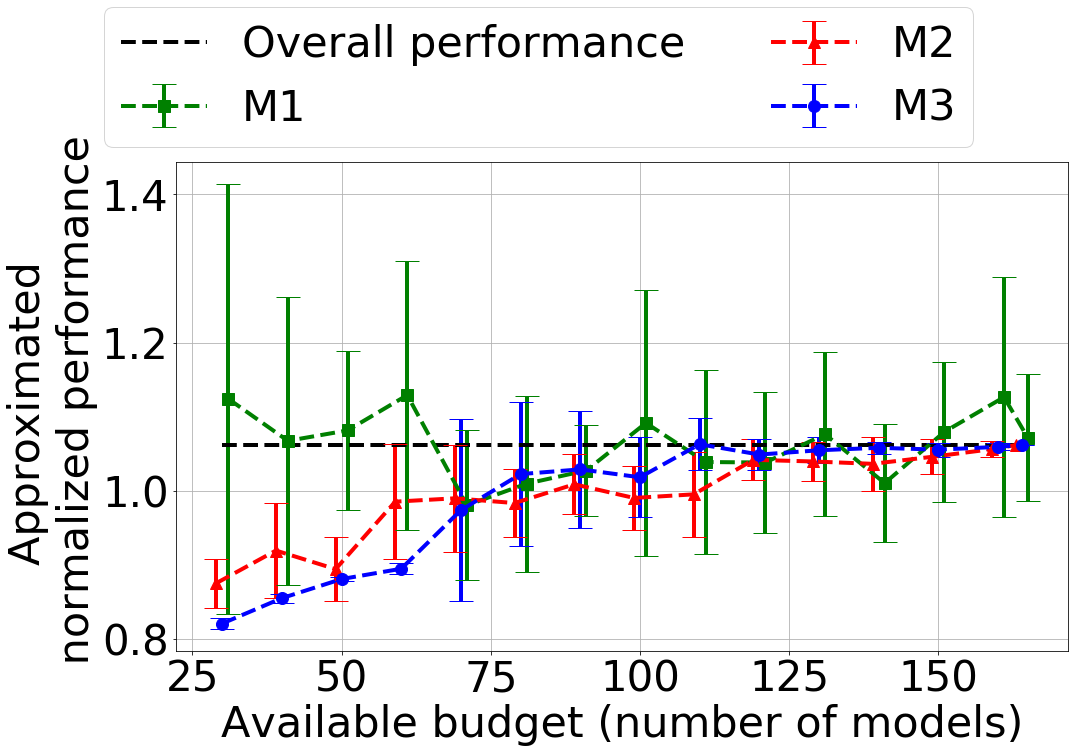}
  \caption{IDQN}
  \label{fig11a}
\end{subfigure}
\begin{subfigure}{0.32\linewidth}
  \centering
  \includegraphics[width=\linewidth]{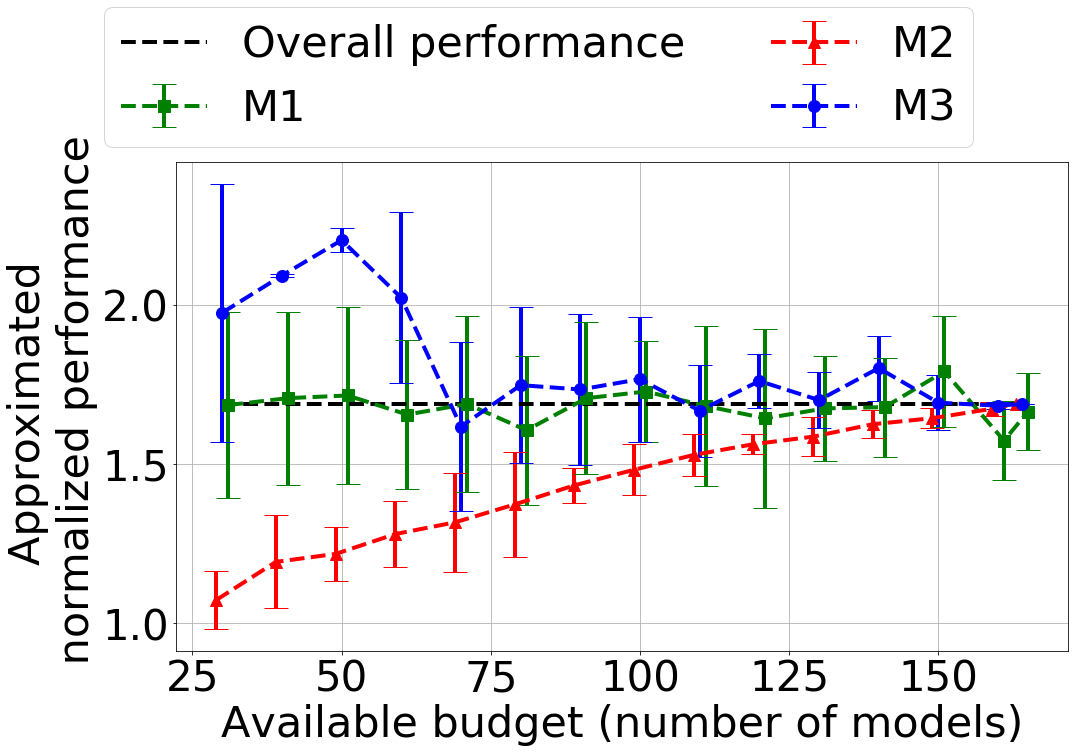}
  \caption{IPPO}
  \label{fig11b}
\end{subfigure}
\begin{subfigure}{0.32\linewidth}
  \centering
  \includegraphics[width=\linewidth]{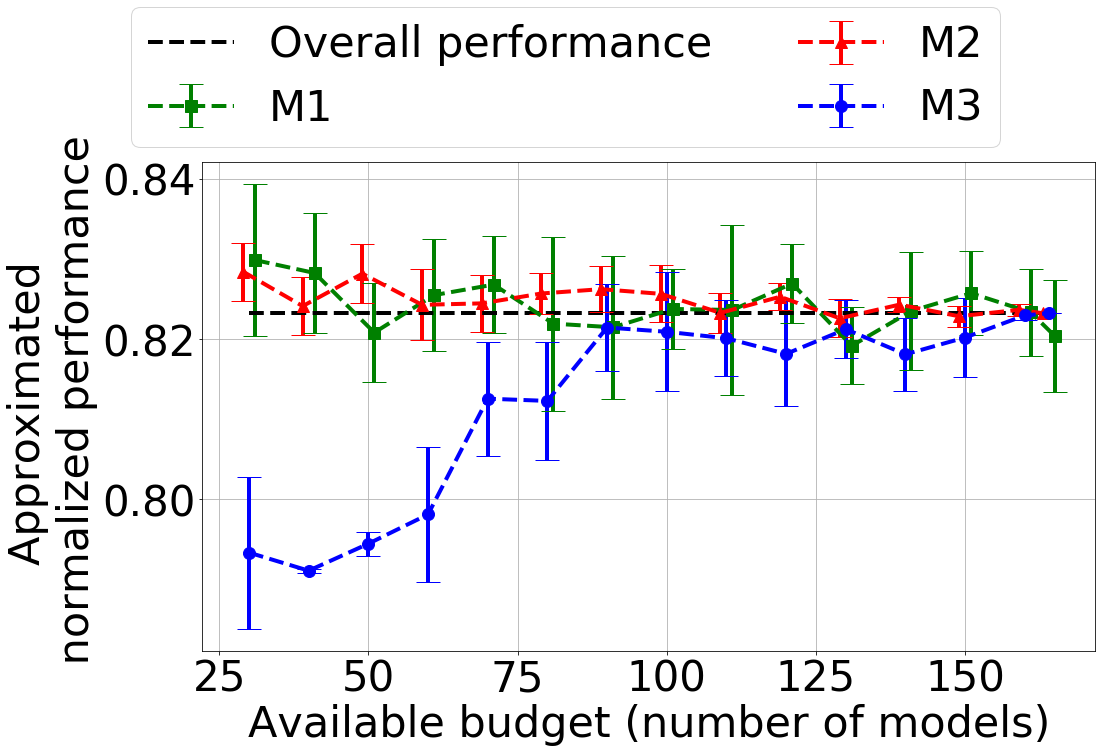}
  \caption{Fixed Time}
  \label{fig11b}
\end{subfigure}
\caption{Approximated performance evaluations over a family of MDPs in traffic signal control with varying budget limits. Closer to the overall performance the better.} 
\label{sampling}
\end{figure*}

In Figure~\ref{sampling}, we denote the approximate evaluations of IDQN, IPPO and Fixed time methods in traffic signal control in comparison to the ground-truth performance while varying the budget size. 
In general, k-means clustering-based approximation produces better estimates than the other two techniques with smaller standard deviations. Specifically, if the budget size is half the total MDP family size, k-means clustering demonstrates reasonably accurate performance estimates. Also, publishing the selection of point MDPs as data sets can enable reproducibility and standardization of the evaluations of the tasks. A sensitivity analysis of the proposed techniques to the underlying point MDP distribution is also given in Appendix~\ref{extra-approximations}.

\textbf{Remarks.}
We acknowledge that there are other factors to consider for a computational budget, including hyperparameter tuning~\cite{jordan2020evaluating} and training using multiple random seeds within each point MDP~\cite{agarwal2021deep}. Here, we focus on the computational budget allotted for the task underspecification issue.

\subsection{Use performance profile of the MDP family instead of point MDP performance}

Inspired by the idea of performance profiling for DRL evaluations~\cite{agarwal2021deep} and in optimization software~\cite{Dolan2002BenchmarkingOS}, our final recommendation is to report the performance of a DRL method over a family of MDPs as a performance profile. Although the overall performance of a DRL method over a family of MDPs can yield more reliable evaluations than evaluating on a point MDP, it may encapsulate further insights into the method's performance. By representing the performance of a method as a performance profile, a more detailed visualization of the performance over the point MDPs can be illustrated. 

In Figure~\ref{fig:performance-profile}, we present an example performance profile for k-means clustering-based performance approximation with a budget size of 80 models in the traffic signal control task.

\begin{minipage}[]{0.6\linewidth}
\includegraphics[width=0.95\textwidth]{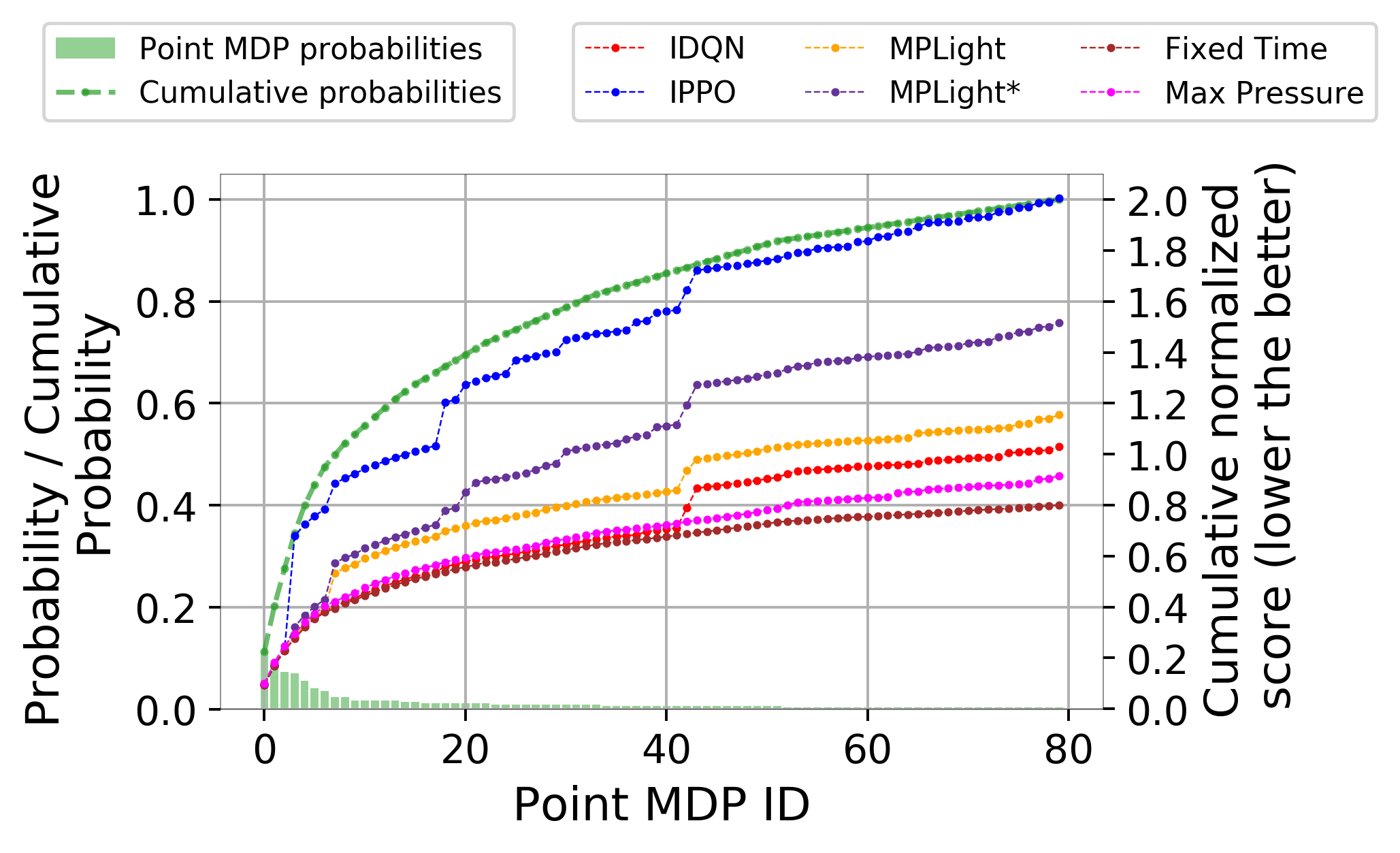}
\captionof{figure}{Performance profile of k-means clustering \\based performance approximation with budget size \\ of 80 models in traffic signal control task.  }
\label{fig:performance-profile}
\end{minipage}
\begin{minipage}[]{0.4\linewidth}

A performance profile illustrates what point MDPs are most probable in the distribution and how each of the point MDP contributes to the final estimate of the performance. It enables direct comparison of methods. In particular, if the cumulative performance curve of method A is strictly below method B, method A is said to \textit{stochastically dominate} method B (lower the score the better)~\cite{agarwal2021deep}. Furthermore, point MDPs in which a method seemingly under-performs is easily visible, giving a clear overview of where the limitations and strengths are originating from.
\end{minipage}

\section{Related Work}

Deep reinforcement learning algorithms have notoriously high variance, resulting in reliability and reproducibility issues when applying them to real-world applications~\cite{agarwal2021deep, jordan2020evaluating, chan2019measuring, bjorck2021high}. Designing sound methodologies for conducting performance evaluations is, therefore, critical. To our knowledge, Falkenauer~\cite{falkenauer1998method} first identified overfitting a design to a selected problem instance as a cause for concern. Whiteson et al.~\cite{whiteson2011protecting} later used the same motivation to argue similar overfitting can happen, particularly in reinforcement learning. We look at the algorithmic generalization of DRL methods within a task, which poses different requirements and properties compared to previous works. 

The use of a family of problem instances is not new. It has been used in combinatorial auctions~\cite{leyton2000towards} and in reinforcement learning~\cite{bhatnagar2007incremental, kalyanakrishnan2009empirical}. Recently, MDP families have been used to achieve better generalization in DRL. Benjamins et al.~\cite{Benjamins2022ContextualizeM} argue that generalization in DRL is held back by factors stemming in part from a lack of problem formalization. Benchmarks suites such as CARL~\cite{Benjamins2021CARLAB} are proposed to use MDP families to study generalization. Eimer et al.~\cite{Eimer2021HyperparametersIC} further show that such DRL problems demonstrate challenges that the DRL community has not looked at carefully. 

The family of MDPs has been formally modeled in multiple ways in the literature. The most recent and general method is to model the problem as a Contextual Markov Decision Processes (cMDP)~\cite{Benjamins2022ContextualizeM} where all MDPs in the family share the same MDP configuration except for the transition function and the reward function. In Hidden Parameter MDPs~\cite{DoshiVelez2016HiddenPM}, only the transition function changes but keep the reward function fixed over the family of MDPs. Epistemic POMDPs introduced by Ghosh et al.~\cite{Ghosh2021WhyGI} are a special case of a cMDP where the context is assumed to be unobservable. They show that there is implicit partial observability under generalization to unseen test conditions from a limited number of training conditions. This phenomenon translates even a fully observed MDP into POMDP. In comparison to these formulations, we do not restrict what components of an MDP can or should change in evaluations. We let that be defined by the domain of the task.

\section{Conclusion and Future Work}
\label{conclusion}
In this work, we identify an important yet overlooked issue of task underspecification in DRL evaluations---the reliance of reporting outcomes on select \textit{point} MDPs. We experimentally demonstrate that evaluating the MDP family often yields a substantially different relative ranking of methods compared to evaluating on select MDPs. Moreover, evaluating on a family of MDPs is not trivial and is faced with multiple challenges. One exciting avenue for future work is to explore if similar shortcomings occur when evaluating the algorithmic generalization of DRL methods across tasks. Furthermore, our recommendations for the related challenges when conducting reliable evaluations with a family of MDPs are only a starting point for more focused research. Future research can shed light on these directions, including designing efficient yet effective methods that can produce good approximate performance estimates. Overall, we intend for our findings to raise awareness of task underspecification that impacts the empirical rigor of DRL and aim to help move the needle toward a more disciplined science overall. 

\section{Broader Impact}

Although there is not a single definition of responsible machine learning, 
the Institute for Ethical AI \& Machine Learning has developed a series of eight principles to guide the responsible development of machine learning systems. \textit{Practical accuracy} is one of the principles, which emphasizes that \textit{accuracy and cost metric functions are aligned to the domain-specific applications}. This article contributes to bolstering the practical accuracy of RL, when employed for a complex downstream decision (e.g. whether to adopt an RL method for a societal system), by highlighting some limitations of current standard RL research practices and proposing to explicitly consider a family of MDPs that constitutes the complex decision. On the other hand, even with more empirically rigorous ML practices, there are still subjective aspects of the task distribution (analogous to the family of MDPs in RL), so it remains important to not over-index on RL-based evaluations.

\section{Acknowledgments}

This work was supported by the MIT Amazon Science Hub, the US DOT's Federal Highway Administration and Utah Department of Transportation under project number F-ST99(783), the MIT-IBM Watson AI Lab, a gift from Mathworks, and the National Science Foundation (NSF) under grant number 2149548. The authors acknowledge MIT SuperCloud and the Lincoln Laboratory Supercomputing Center for providing computational resources supporting the research results in this paper. The authors are also grateful for insightful discussions with Jiaqi Zhang and the constructive suggestions from the anonymous reviewers.


\bibliographystyle{plain}
\bibliography{references}

\newpage
\section*{Checklist}

\begin{enumerate}

\item For all authors...
\begin{enumerate}
  \item Do the main claims made in the abstract and introduction accurately reflect the paper's contributions and scope?
    \answerYes{}
  \item Did you describe the limitations of your work?
    \answerYes{See Section~\ref{conclusion}}.
  \item Did you discuss any potential negative societal impacts of your work?
    \answerYes{}
  \item Have you read the ethics review guidelines and ensured that your paper conforms to them?
    \answerYes{}
\end{enumerate}

\item If you are including theoretical results...
\begin{enumerate}
  \item Did you state the full set of assumptions of all theoretical results?
    \answerNA{}
	\item Did you include complete proofs of all theoretical results?
    \answerNA{}
\end{enumerate}

\item If you ran experiments...
\begin{enumerate}
  \item Did you include the code, data, and instructions needed to reproduce the main experimental results (either in the supplemental material or as a URL)?
    \answerYes{See Table~\ref{task-mdp-detailed-v} and Section \ref{case-study-appendix} in Appendix. 
    }
  \item Did you specify all the training details (e.g., data splits, hyperparameters, how they were chosen)?
    \answerYes{Provided the details in the respective sections where they are discussed.}
	\item Did you report error bars (e.g., with respect to the random seed after running experiments multiple times)?
    \answerYes{}
	\item Did you include the total amount of compute and the type of resources used (e.g., type of GPUs, internal cluster, or cloud provider)?
    \answerNA{}
\end{enumerate}

\item If you are using existing assets (e.g., code, data, models) or curating/releasing new assets...
\begin{enumerate}
  \item If your work uses existing assets, did you cite the creators?
    \answerYes{}
  \item Did you mention the license of the assets?
    \answerYes{Attribution-NonCommercial-ShareAlike 2.0 Generic (CC BY-NC-SA 2.0)}
  \item Did you include any new assets either in the supplemental material or as a URL?
    \answerNA{}
  \item Did you discuss whether and how consent was obtained from people whose data you're using/curating?
    \answerNA{}
  \item Did you discuss whether the data you are using/curating contains personally identifiable information or offensive content?
    \answerNA{}
\end{enumerate}

\item If you used crowdsourcing or conducted research with human subjects...
\begin{enumerate}
  \item Did you include the full text of instructions given to participants and screenshots, if applicable?
    \answerNA{}
  \item Did you describe any potential participant risks, with links to Institutional Review Board (IRB) approvals, if applicable?
    \answerNA{}
  \item Did you include the estimated hourly wage paid to participants and the total amount spent on participant compensation?
    \answerNA{}
\end{enumerate}

\end{enumerate}
\newpage
\LARGE
\appendix{Appendix}
\normalsize
\section{Introduction}

\subsection{Task underspecification in common benchmark DRL tasks}

In Table~\ref{underspecification-table}, we present some of the commonly used DRL benchmark tasks for evaluations and some ways they may be underspecified. We note that the ways in which a task is underspecified may depend on the application context, and thus the table is meant to be illustrative rather than exhaustive.

\begin{center}
\noindent

\label{underspecification-table}
\begin{tabularx}{\textwidth}{ @{} T X }
  \toprule
  \textbf{Task} & \textbf{Underspecification}\\
  \midrule
  Cartpole & 
  \begin{tabular}[t]{ @{\makebox[0.5em][l]{\textbullet}} p{\dimexpr\linewidth-0.9em} p{\dimexpr\linewidth-0.4em} }
    Cart mass, pole mass, pole length, and gravity.\\
    \end{tabular} \\
  \midrule
  Mountain car & 
  \begin{tabular}[t]{ @{\makebox[0.5em][l]{\textbullet}} p{\dimexpr\linewidth-0.9em} p{\dimexpr\linewidth-0.4em} }
    Heights of the mountains, the trigonometric curves defining the mountains, and gravity.
  \end{tabular} \\
  \midrule
  Lunar lander  & 
  \begin{tabular}[t]{ @{\makebox[0.5em][l]{\textbullet}} p{\dimexpr\linewidth-0.4em} p{\dimexpr\linewidth-0.4em} }
    Landing polygon size, lander leg heights, widths, and gravity changes.\\
  \end{tabular} \\
  \midrule
  Acrobot & 
  \begin{tabular}[t]{ @{\makebox[0.5em][l]{\textbullet}} p{\dimexpr\linewidth-0.4em} p{\dimexpr\linewidth-0.4em} }
     Masses and lengths of the two links connected linearly to form a acrobot chain.
  \end{tabular} \\
  \midrule
  Pendulum  & 
  \begin{tabular}[t]{ @{\makebox[0.5em][l]{\textbullet}} p{\dimexpr\linewidth-0.4em} p{\dimexpr\linewidth-0.4em} }
    Mass and length of the pendulum. \\
  \end{tabular} \\
  \midrule
  Swimmer  & 
  \begin{tabular}[t]{ @{\makebox[0.5em][l]{\textbullet}} p{\dimexpr\linewidth-0.4em} p{\dimexpr\linewidth-0.4em} }
    Capsule sizes for the segments comprising the robot. \\
  \end{tabular} \\
  \midrule
  Walker2D  & 
  \begin{tabular}[t]{ @{\makebox[0.5em][l]{\textbullet}} p{\dimexpr\linewidth-0.4em} p{\dimexpr\linewidth-0.4em} }
    Capsule sizes for the segments comprising the robot. \\
  \end{tabular} \\
  \midrule
    Breakout  & 
  \begin{tabular}[t]{ @{\makebox[0.5em][l]{\textbullet}} p{\dimexpr\linewidth-0.4em} p{\dimexpr\linewidth-0.4em} }
    Paddle length and friction of the paddle surface. \\
  \end{tabular} \\
  \midrule
  Pong  & 
  \begin{tabular}[t]{ @{\makebox[0.5em][l]{\textbullet}} p{\dimexpr\linewidth-0.4em} p{\dimexpr\linewidth-0.4em} }
    Paddle lengths.  \\
  \end{tabular} \\
  \bottomrule
\end{tabularx}

\end{center}
\section{Shortcomings of Point MDP-based Performance Evaluations}

\subsection{Example task configurations}
\label{example-task-detailed}

Table~\ref{task-mdp-detailed-v} provides the MDP configurations that was used in Section~\ref*{pitfalls} for elaborating on the potential shortcomings of point MDP-based evaluations. For each task, $\star$ denotes the generic MDP, and $\dagger$ represents the simplified MDP. The simplified MDP is usually designed to be what is supposed to be the easiest MDP or the one that simplifies the transition dynamics. The generic MDP is the parametric mean MDP of four other MDPs of the family. We verified that all the MDPs provided in Table~\ref{task-mdp-detailed-v} can be solved under the enforced actuator limits and that none of the settings are under-actuated.

\begin{center}
\noindent
\captionof{table}{Control task and the MDP families. For each task, $\star$ denotes the generic MDP and $\dagger$ represents the simplified MDP.} 
\label{task-mdp-detailed-v}
\begin{tabularx}{\textwidth}{ @{} c X T }
  \toprule
  Task & Task description & MDP family \\
  \midrule
  Quad & 
  \begin{tabular}[t]{ @{\makebox[0.5em][l]{\textbullet}} p{\dimexpr\linewidth-0.9em} p{\dimexpr\linewidth-0.4em} }
    \textbf{Goal}: maneuver a 2D quadcopter through an obstacle course using its vertical acceleration as the control action. \\
    \textbf{MDP family}: MDPs with varying the obstacle course lengths (upper obstacle length $u$ and lower obstacle length $l$)
  \end{tabular} 
  & 
  \begin{tabular}[t]{ @{\makebox[0.5em][l]{\textbullet}} p{\dimexpr\linewidth-0.9em} p{\dimexpr\linewidth-0.4em} }
    (u=0.0, l=6.0)$^\dagger$\\
    (u=0.5, l=3.0)\\
    (u=0.625, l=4.0)$^\star$\\
    (u=1.0, l=2.0)\\
    (u=1.0, l=5.0)\\
    \end{tabular} \\
  \midrule
  Pendulum & 
  \begin{tabular}[t]{ @{\makebox[0.5em][l]{\textbullet}} p{\dimexpr\linewidth-0.9em} p{\dimexpr\linewidth-0.4em} }
    \textbf{Goal}: swing up the pendulum. \\
    \textbf{MDP family}: MDPs with varying masses ($m$) and lengths ($l$) of the pendulum
  \end{tabular} 
  & 
  \begin{tabular}[t]{ @{\makebox[0.5em][l]{\textbullet}} p{\dimexpr\linewidth-0.4em} p{\dimexpr\linewidth-0.4em} }
    (m=1.0, l=1.0)$^\dagger$\\
    (m=1.5, l=4.0)\\
    (m=1.875, l=5.25)$^\star$\\
    (m=2.0, l=6.0)\\
    (m=3.0, l=10.0)\\
  \end{tabular} \\
  \midrule
  Swimmer & 
  \begin{tabular}[t]{ @{\makebox[0.5em][l]{\textbullet}} p{\dimexpr\linewidth-0.4em} p{\dimexpr\linewidth-0.4em} }
    MuJoCo 3-link swimming robot in a viscous fluid. \\
    \textbf{Goal}: make the robot swim forward as fast as possible by actuating the two joints.  \\
    \textbf{MDP family}: MDPs with varying capsule sizes for the segments comprising the robot swimmer ($a$, $b$ and $c$)
  \end{tabular} 
  & \begin{tabular}[t]{ @{\makebox[0.5em][l]{\textbullet}} p{\dimexpr\linewidth-0.4em} p{\dimexpr\linewidth-0.4em} }
    MDP 1 =(a=0.10, b=0.10, c=0.10)\\
    MDP 2 =(a=0.15, b=0.15, c=0.15)\\
    MDP 3 =(a=0.10, b=0.15, c=0.10)\\
    MDP 4 =(a=0.05, b=0.05, c=0.05)$^\dagger$\\
    MDP 5 =(a=0.10, b=0.1125, c=0.10)$^\star$\\
    \end{tabular} \\
  \bottomrule
\end{tabularx}

\end{center}

\subsection{Visual illustrations of MDP families}
\label{visual-mdp-families}

\begin{figure*}[h]
\centering
\begin{subfigure}{0.18\linewidth}
  \centering
  \includegraphics[width=\linewidth]{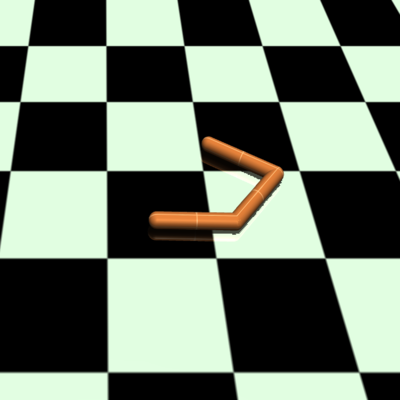}
  \caption{MDP 1}
  \label{fig11a}
\end{subfigure}
\begin{subfigure}{0.18\linewidth}
  \centering
  \includegraphics[width=\linewidth]{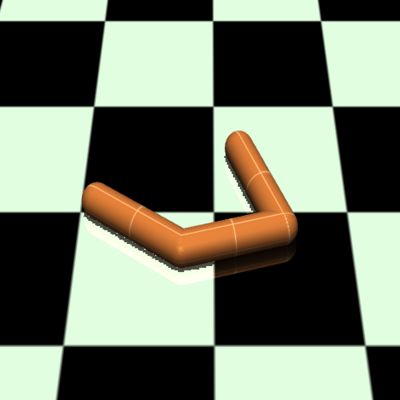}
  \caption{MDP 2}
  \label{fig11b}
\end{subfigure}
\begin{subfigure}{0.18\linewidth}
  \centering
  \includegraphics[width=\linewidth]{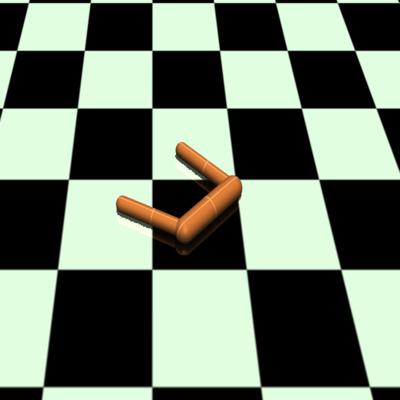}
  \caption{MDP 3}
  \label{fig11b}
\end{subfigure}
\begin{subfigure}{0.18\linewidth}
  \centering
  \includegraphics[width=\linewidth]{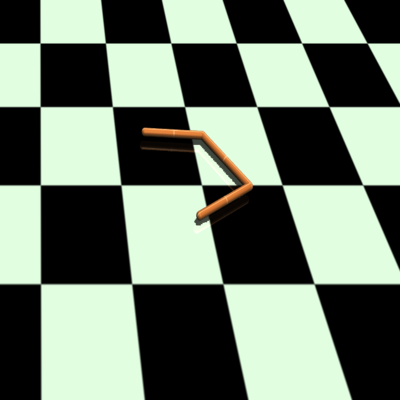}
  \caption{MDP 4}
  \label{fig11b}
\end{subfigure}
\begin{subfigure}{0.18\linewidth}
  \centering
  \includegraphics[width=\linewidth]{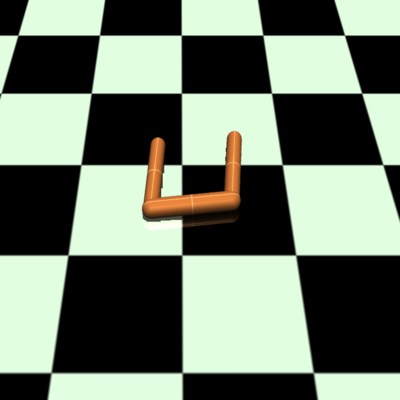}
  \caption{MDP 5}
  \label{fig11b}
\end{subfigure}
\caption{Visual illustration of family of point MDPs used in the Swimmer task}
\label{swimmer}
\end{figure*}

\begin{figure*}[h]
\centering
\begin{subfigure}{0.19\linewidth}
  \centering
  \includegraphics[width=\linewidth]{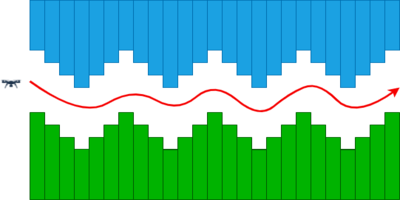}
  \caption{(u=1,l=5)}
  \label{fig11a}
\end{subfigure}
\begin{subfigure}{0.19\linewidth}
  \centering
  \includegraphics[width=\linewidth]{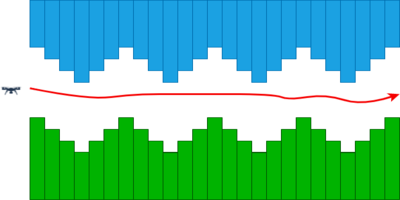}
  \caption{(u=0,l=6)}
  \label{fig11b}
\end{subfigure}
\begin{subfigure}{0.19\linewidth}
  \centering
  \includegraphics[width=\linewidth]{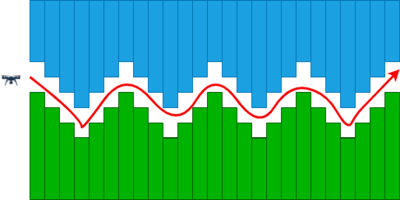}
  \caption{(u=0.5,l=3)}
  \label{fig11b}
\end{subfigure}
\begin{subfigure}{0.19\linewidth}
  \centering
  \includegraphics[width=\linewidth]{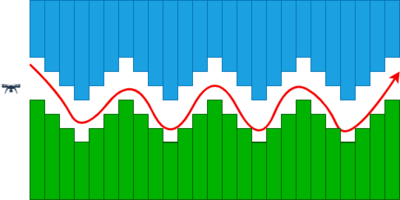}
  \caption{(u=0.625,l=4)}
  \label{fig11b}
\end{subfigure}
\begin{subfigure}{0.19\linewidth}
  \centering
  \includegraphics[width=\linewidth]{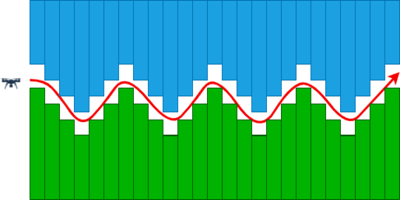}
  \caption{(u=1,l=2)}
  \label{fig11b}
\end{subfigure}
\caption{Visual illustration of family of point MDPs used in the Quad task}
\label{quad}
\end{figure*}

\begin{figure*}[h]
\centering
\begin{subfigure}{0.18\linewidth}
  \centering
  \includegraphics[width=\linewidth]{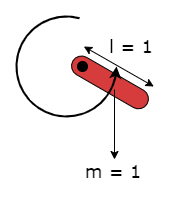}
  \caption{(m=1,l=1)}
  \label{fig11a}
\end{subfigure}
\begin{subfigure}{0.18\linewidth}
  \centering
  \includegraphics[width=\linewidth]{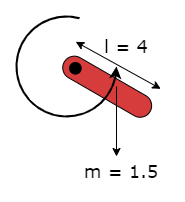}
  \caption{(m=1.5,l=4)}
  \label{fig11b}
\end{subfigure}
\begin{subfigure}{0.18\linewidth}
  \centering
  \includegraphics[width=\linewidth]{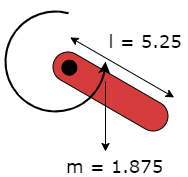}
  \caption{(m=1.875,l=5.25)}
  \label{fig11b}
\end{subfigure}
\begin{subfigure}{0.18\linewidth}
  \centering
  \includegraphics[width=\linewidth]{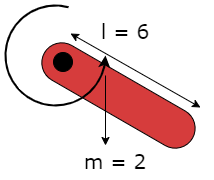}
  \caption{(m=2,l=6)}
  \label{fig11b}
\end{subfigure}
\begin{subfigure}{0.18\linewidth}
  \centering
  \includegraphics[width=\linewidth]{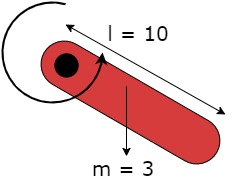}
  \caption{(m=3,l=10)}
  \label{fig11b}
\end{subfigure}
\caption{Visual illustration of family of point MDPs used in the Pendulum task}
\label{pendulum}
\end{figure*}

In Figures~\ref{swimmer}, \ref{quad} and \ref{pendulum}, we visually demonstrate the family of MDPs described in Table~\ref{task-mdp-detailed-v}.

\subsection{Further examples on illustrating DRL training can be sensitive to the selected point MDP properties.}
\label{i2-more-examples}

\begin{figure*}[hbt]
\centering
\begin{subfigure}{0.32\linewidth}
  \centering
  \includegraphics[width=\linewidth]{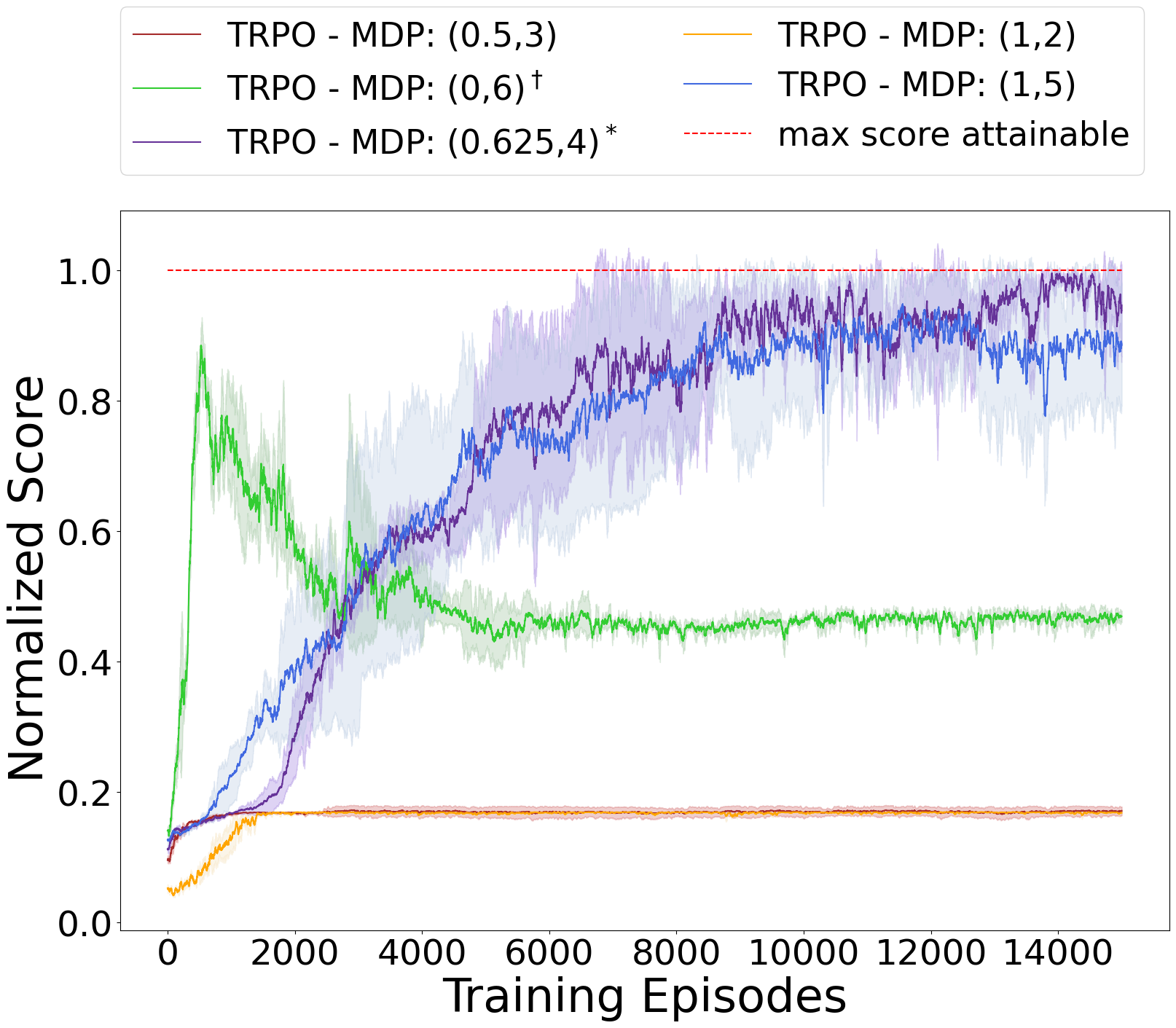}
  \caption{Quad TRPO training}
  \label{fig11a}
\end{subfigure}
\begin{subfigure}{0.32\linewidth}
  \centering
  \includegraphics[width=\linewidth]{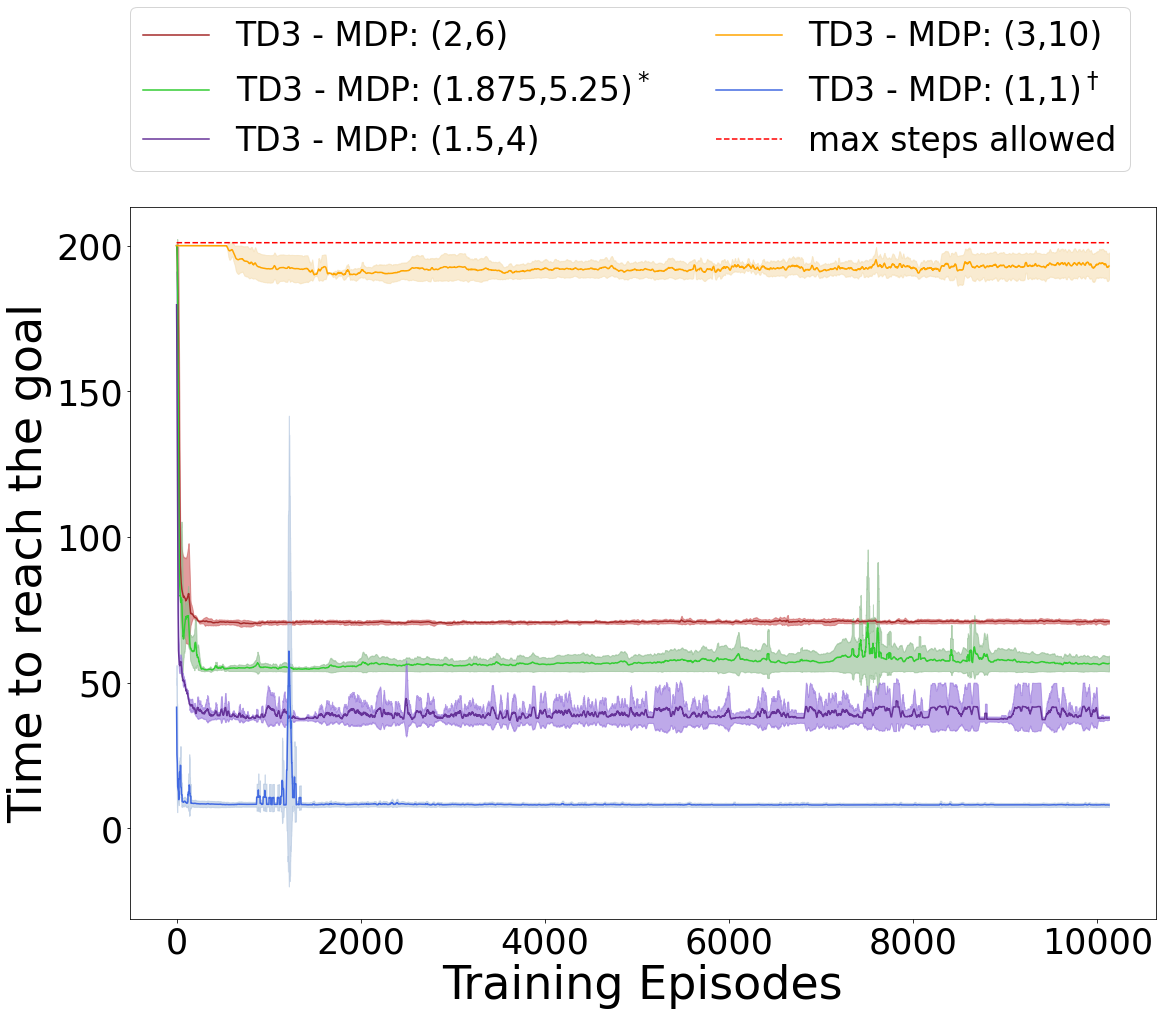}
  \caption{Pendulum TD3 training}
  \label{fig11b}
\end{subfigure}
\begin{subfigure}{0.33\linewidth}
  \centering
  \includegraphics[width=\linewidth]{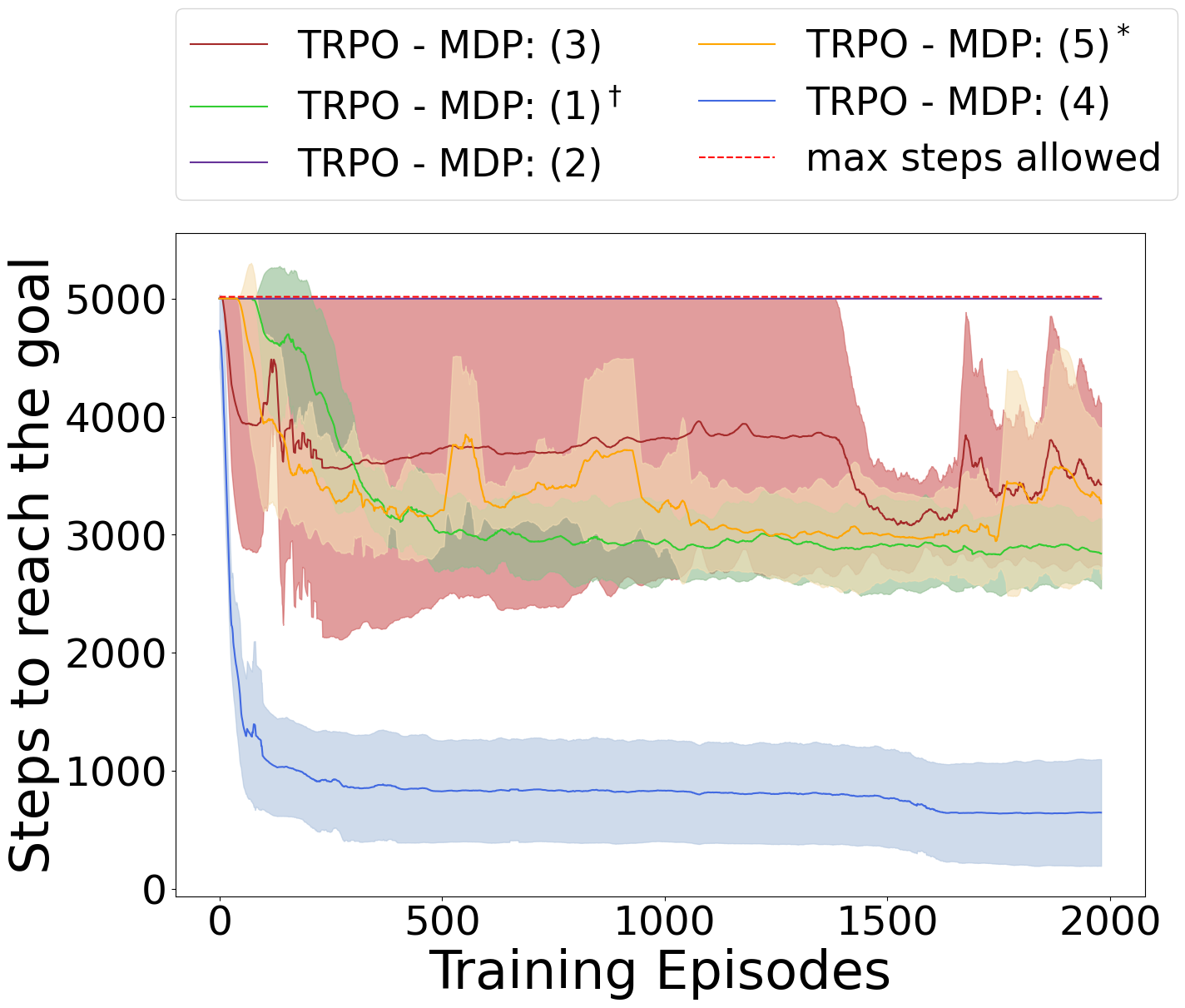}
  \caption{Swimmer TRPO training}
  \label{fig11b}
\end{subfigure}
\caption{Training progress of each task for different point MDPs. For Quad, performance is a normalized score that indicates the distance the quadcopter traveled before crashing or reaching the goal with respect to the total distance required to travel (higher the better). In the pendulum, the training progress is measured as the time to swing the pendulum up (lower the better). Finally, for the swimmer, the training progress is measured as the time to reach the goal (lower the better). TRPO was used in training Quad and Swimmer, whereas TD3 was used for the Pendulum.}
\label{Issue-two-further}
\end{figure*}

Figure~\ref{Issue-two-further} illustrates another example set of training curves for illustrating observation 2 of Section~\ref{pitfalls}. Here we use TRPO for training Quad and Swimmer, whereas TD3 was used for Pendulum. In summary, clearly, there are variations in the training depending on the choice of point MDP for all three control tasks as pointed out in Section~\ref{pitfalls}. 

\section{Case Study: Traffic Signal Control}
\label{case-study-appendix}

\subsection{RESCO traffic signal control benchmark}
\label{case-study-methods}
RESCO provides a standardized implementation of state-of-the-art DRL algorithms for traffic signal control that have become popular in recent years. It also provides non-RL baseline methods from the traffic engineering community.

We use six algorithms from RESCO in our case study, namely: (1) \textbf{IDQN}: a deep Q-learning approach, (2) \textbf{IPPO}~\cite{interpretable_TSC}: same as IDQN with a modified output layer, (3) \textbf{MPLight}~\cite{mplight}: scalable FRAP model~\cite{Zheng2019LearningPC} approach using the pressure concept, (4) \textbf{MPLight$^*$}: MPLight implementation with the addition of sensing information, (5) \textbf{Fixed time}: a fine tuned non-RL pre-defined controller where phases are enabled for a fixed duration following a fixed cycle, (6) \textbf{Max pressure}~\cite{VARAIYA2013177}: a non-RL controller where phase selection is based on the maximal joint pressure. 

\subsection{Intersection distribution}
\label{intersection-dist}
In this work, we use Salt Lake City intersection data for building the distribution due to its well-documented and advanced traffic network system. Our data for building the intersection distribution comes from a combination of open data sources. For most of the street network data which includes street geometry and layout, we use OpenStreetMaps (OSM) \cite{osm}. As for the traffic signal and demand data, we utilize data from the UDOT Open Data Portal~\cite{opendatautah} and the Automated Traffic Signal Performance Measures (ATSPM)~\cite{atspm2020components}. 

The first part of the data used consists of road networks obtained from OSM. We use the OSMnx Python package to manipulate the OSM data. We perform a data pre-processing that involves using mean substitution for NaN values (for example, the mean speed of 25 mph was used as a default) and the removal of motorways and motor links which are beyond the scope of this analysis. Next, we join the OSM data with UDOT Data Portal 2019 traffic demand data~\cite{aadt} by intersecting the two datasets based on the edge locations. This is a manual step done within ArcGIS Pro, which involves buffering the data to account for slight positional differences in the location of edges from the two datasets as well as calculating bearings to correct for any improperly intersected edges. The UDOT provides traffic demand data in terms of average annual daily traffic, a measure of the traffic volume of an entire year averaged over 365 days. We utilized this dataset to define the vehicle inflow rates (i.e., vehicles per hour) at the intersections. With that, we finally use six features to describe an intersection: number of lanes,  maximum allowed speed, lane length, traffic inflow, number of left turning lanes, and number of right turning lanes. In Table~\ref{table:features}, we summarize the mean and standard deviation of the selected six features.

\begin{table}
\begin{center}
\begin{tabular}{lccc} \toprule
Feature & Units & Mean & Standard Dev\\ \midrule
Lane Count & - & 3.8 & 1.37\\ \midrule
Speed & mph & 32.6 & 5.40\\ \midrule
Length of Lanes & meters & 260.8 & 193\\ \midrule
Vehicle inflow & vehicles/hour & 73.5 & 774\\ \midrule
Left Turns Count & - & 0.229 & 0.496 \\ \midrule
Right Turns Count & - & 0.100 & 0.298\\ \midrule
\end{tabular}
\caption{Overview of the features that describe the intersection point MDP distribution}
\label{table:features}
\end{center}
\end{table}

Next, we filter this network to take a subset of intersections (and their adjacent streets) that correspond geospatially to signalized intersections in Salt Lake City from the Automated Traffic Signal Performance Measures. Finally, we build the full distribution of intersections based on this dataset. In our analysis, we use 345 intersections. 

\section{Shortcomings of Point MDP-based Evaluations in Common DRL Tasks}
\label{drl-tasks-pitfalls}
In this section, we provide an in-detail view of the shortcomings of point MDP-based evaluations by taking three popular control tasks as examples: cartpole (discrete actions), pendulum (continuous actions), and half cheetah (continuous actions). For each task, we devise a family of MDPs as described in Table~\ref{drl-tasks-detailed-vv} using CARL benchmark suite~\cite{Benjamins2021CARLAB}. Our MDP families consist of 576, 180, and 120 point MDPs for cartpole, pendulum, and half cheetah, respectively. 

\subsection{MDP family configurations}
\label{drl-task-config}

\begin{center}
\noindent
\captionof{table}{Popular DRL control tasks and the context features defining the MDP families} 
\label{drl-tasks-detailed-vv}
\begin{tabularx}{\textwidth}{ @{} c X T }
  \toprule
  Task & Task description & Context features \\
  \midrule
  Cartpole & 
  \begin{tabular}[t]{ @{\makebox[0.5em][l]{\textbullet}} p{\dimexpr\linewidth-0.9em} p{\dimexpr\linewidth-0.4em} }
    \textbf{Goal}: balance a pole attached to an un-actuated joint of a cart by applying forces in the left and right direction on the cart.
  \end{tabular} 
  & 
  \begin{tabular}[t]{ @{\makebox[0.5em][l]{\textbullet}} p{\dimexpr\linewidth-0.9em} p{\dimexpr\linewidth-0.4em} }
    pole length (0.05, 0.5, 3, 5)\\
    mass of the cart (0.1, 1, 6, 10)\\
    mass of the pole (0.01, 0.1, 0.5, 1)\\
    force magnifier (1, 50, 100)\\
    gravity (0.1, 9.8, 19.6)\\
    \end{tabular} \\
  \midrule
  Pendulum & 
  \begin{tabular}[t]{ @{\makebox[0.5em][l]{\textbullet}} p{\dimexpr\linewidth-0.9em} p{\dimexpr\linewidth-0.4em} }
    \textbf{Goal}: apply torque on the free end of the pendulum to swing it into an upright position.
  \end{tabular} 
  & 
  \begin{tabular}[t]{ @{\makebox[0.5em][l]{\textbullet}} p{\dimexpr\linewidth-0.4em} p{\dimexpr\linewidth-0.4em} }
    mass of the pendulum (0.4, 1, 1.5, 2, 3, 4)\\
    length of the pendulum (0.5, 1, 2, 4, 7, 10)\\
    gravity (2, 5, 10, 12, 15)
  \end{tabular} \\
  \midrule
  Half-cheetah & 
  \begin{tabular}[t]{ @{\makebox[0.5em][l]{\textbullet}} p{\dimexpr\linewidth-0.4em} p{\dimexpr\linewidth-0.4em} }
    A 2-dimensional robot consisting of 9 links and 8 joints connecting them (including two paws). \\
    \textbf{Goal}: apply a torque on the joints to make the cheetah run forward as fast as possible.
  \end{tabular} 
  & \begin{tabular}[t]{ @{\makebox[0.5em][l]{\textbullet}} p{\dimexpr\linewidth-0.4em} p{\dimexpr\linewidth-0.4em} }
    gravity (-2, -5, -9.8, -12, -15)\\
    torso mass (0.5, 2, 5, 9.457, 12, 15)\\
    friction (0.1, 0.3, 0.6, 1.0)\\
    \end{tabular} \\
  \bottomrule
\end{tabularx}

\end{center}

In creating the MDP families, we generate point MDPs by generating all combinations of context features as specified in Table~\ref{drl-tasks-detailed-vv}. 

\subsection{Analysis of performance scores}
\label{drl_tasks_analysis}

\begin{figure*}[!h]
  \includegraphics[width=0.99\textwidth]{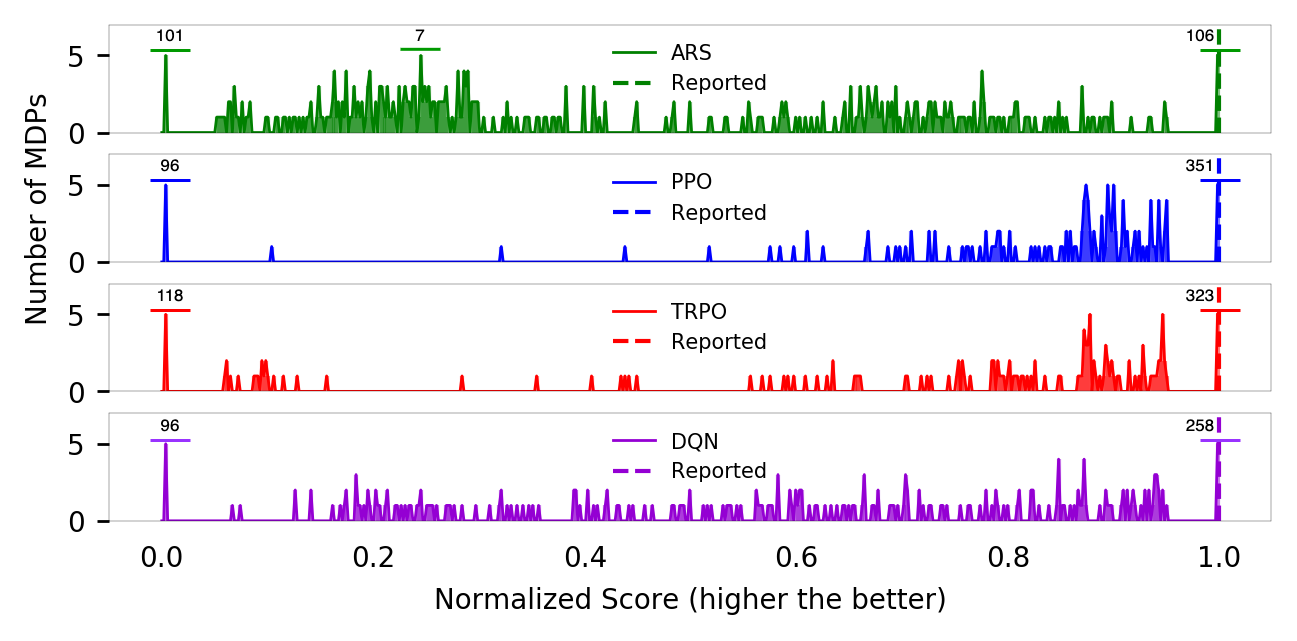}
  \caption{Performance vs. the number of point MDPs that demonstrate the performance using four popular DRL algorithms in Cartpole. 576 unique point MDPs were considered for each method. For better visibility, we limit the height of the y-axis to five point MDPs. If there are more than five point MDPs for a given performance score, we indicate the total number on the corresponding vertical plot line.}
  \label{fig:DRL-performance-cartpole}
\end{figure*}

\begin{figure*}[!h]
  \includegraphics[width=0.99\textwidth]{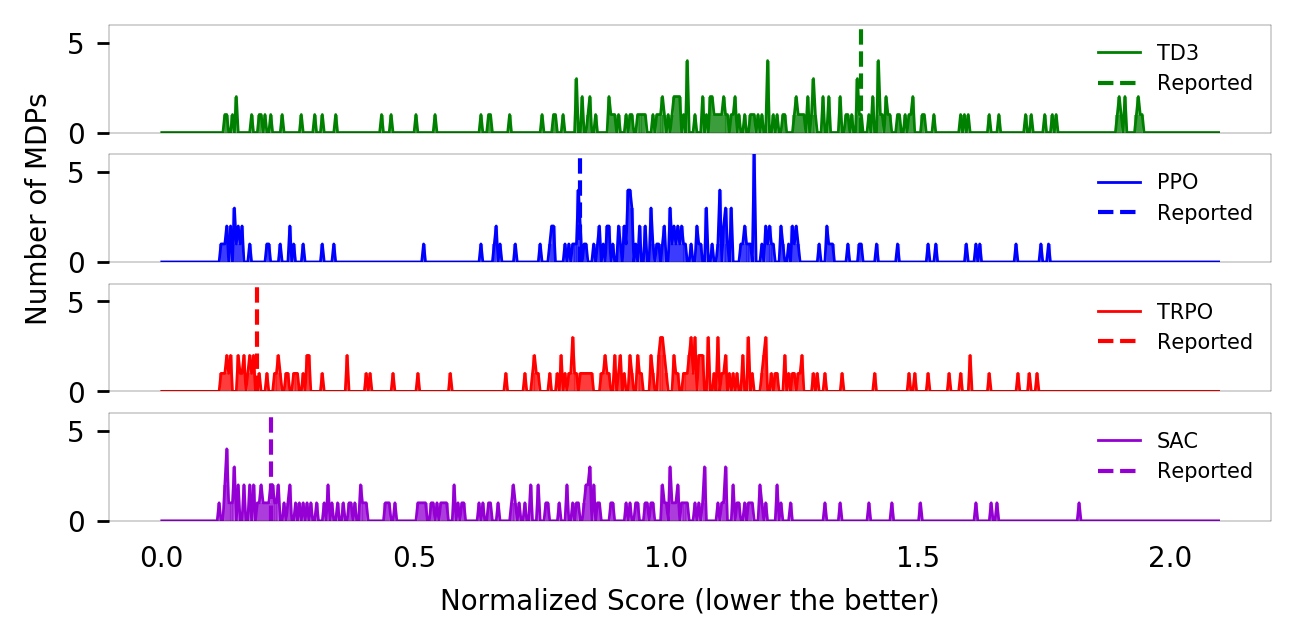}
  \caption{Performance vs. the number of point MDPs that demonstrate the performance using four popular DRL algorithms in the Pendulum. 180 unique point MDPs were considered for each method.}
  \label{fig:DRL-performance-pendulum}
\end{figure*}

\begin{figure*}[!h]
  \includegraphics[width=0.99\textwidth]{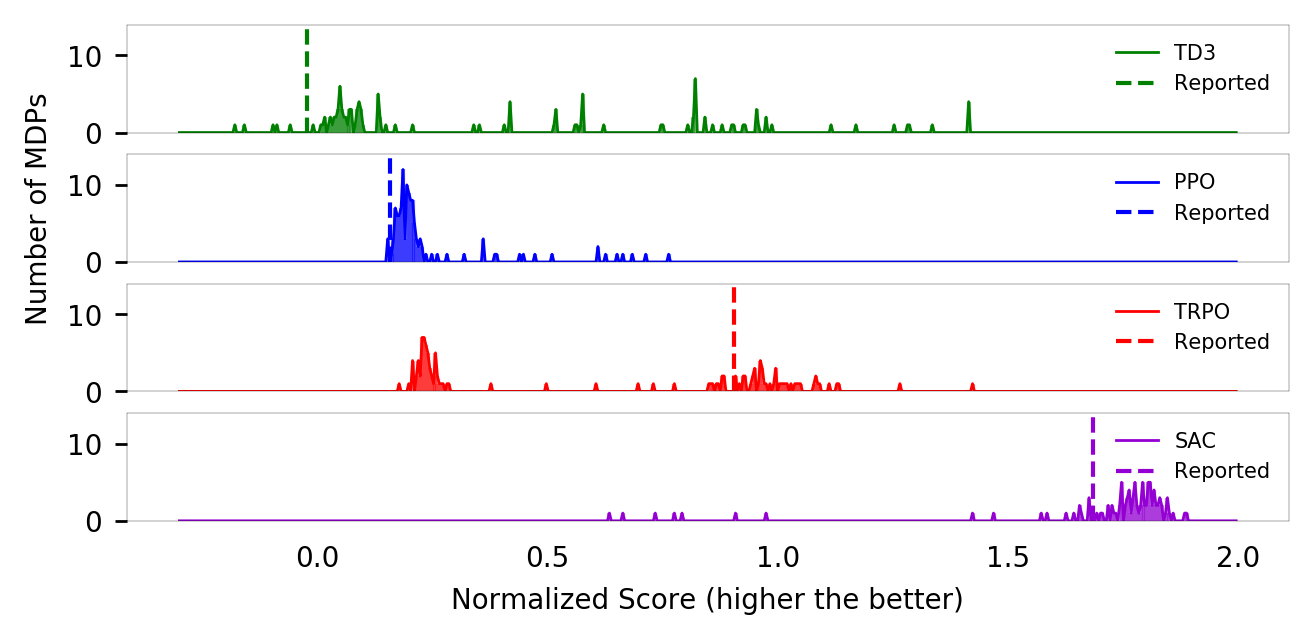}
  \caption{Performance vs. the number of point MDPs that demonstrate the performance using four popular DRL algorithms in Half Cheetah. 120 unique point MDPs were considered for each method.}
  \label{fig:DRL-performance-half-cheetah}
\end{figure*}

In Figures~\ref{fig:DRL-performance-cartpole},~\ref{fig:DRL-performance-pendulum} and \ref{fig:DRL-performance-half-cheetah}, we illustrate performance score variations in cartpole, pendulum and half-cheetah tasks in addition to what we illustrated in Section~\ref{common-tasks}. The reported performance of each task is measured by training and evaluating DRL methods on commonly used single point-MDP given in the benchmark suites (OpenAI gym for cartpole and pendulum and Brax for half cheetah). For measuring overall performance, we use uniform point MDP distribution (each point MDP is equally important).

Figure~\ref {fig:DRL-performance-cartpole} provided details insights into the performance of DRL methods in the cartpole task. As can be seen, all four DRL methods report a performance of 1.0, indicating optimally solving the cartpole task. However, when evaluated on the family, we see significant discrepancies. First, not all methods are equally well performing as reported. Specifically, methods like ARS and DQN underperform significantly compared to methods like PPO or TRPO. Second, all methods struggle to solve nearly 17\% of point MDPs, while PPO and TRPO have the highest success in optimally solving point MDPs (61\% and 56\%, respectively). This calls for a significant method ranking change. Although reported results would rank all methods as equally well-performing, our analysis indicates that the ranking of methods is PPO, TRPO, DQN, and then ARS. 

Similar insights can be seen in Figure~\ref{fig:DRL-performance-pendulum} for the pendulum task. The performance of TRPO and SAC are clearly overestimated under reported performances. Also, there is a significant variation in performance under all four DRL methods. Although one would tend to pick TRPO as the best-performing method under reported performance, our analysis shows that SAC is, in fact, the highest-performing method. (Section~\ref{common-tasks} Figure ~\ref{pendulum-a}). This again calls for a ranking change.  

Figure~\ref{fig:DRL-performance-half-cheetah} shows similar results for half cheetah task. However, we see slightly less variation in performances across MDPs. We believe if more contextual features are used for describing point-MDPs, we would see more variations, just as seen in other tasks. In terms of ranking of DRL methods, although in this case, the first and second performing methods (SAC and TRPO) stay the same, third and fourth places are swapped. According to the reported performances, PPO outperforms TD3, but we show that TD3 is in fact better than PPO when overall performance is used for ranking. Additionally, although TRPO’s rank stays the same, we see its performance has degraded by a significant margin. These phenomena again highlight the shortcomings of point MDP-based evaluations when used in evaluating algorithmic generalization within a task. 

\section{Reliable Evaluations Within a Task}
\label{futher-budget}
\subsection{Evaluate DRL methods on a family of MDPs instead of point MDPs}
\label{performance approximations}
Further details on the approximation algorithms used for performance approximations are presented below.

\textbf{M1: random sampling with replacements}: Under this method, we sample point MDPs from the distribution randomly with replacements. We sample exactly the same number of point MDPs as the maximum budget. One can decide to sample $N/n$ point MDPs where $N$ is the budget, and $n$ is the number of parallel runs used for the evaluation of each point MDP. In this work we set $n=1$. After sampling and evaluations on each point MDP, the overall performance is defined as $\sum_{i \in X}^{} s_{R,i}$ where $X$ denotes the index set of point MDPs that were sampled.  

\textbf{M2: random sampling without replacements}: This method is similar to M1 except that we do not perform replacements. The overall performance is defined as $\sum_{i \in X}^{} \hat{q_{T, i}} s_{R,i}$ where $\hat{q_{T,i}} = \frac{q_{T,i}}{\sum_{i \in X}^{}q_{T,i}}$ and $q_{T,i}$ is the probability of point MDP $i$ according to the distribution. $X$ denotes the index set of point MDPs that were sampled.

\textbf{M3: clustering point MDPs using k-means}: In this method, we cluster the point MDPs with k-means clustering giving each point MDP an equal weight. Next, we assign the sum of probabilities of all point MDPs that belong to the same cluster to its centroid. However, a point MDP that is represented by a centroid may not actually exist in the problem domain. Therefore, we match each centroid to the nearest actual point MDP from the family and conduct evaluations. The overall performance is then defined as $\sum_{i \in X}^{} q^*_{T, i} s_{R,i}$ where $q^*_{T,i}$ is the total probability assigned to each point MDP selected to represent a cluster centroid. $X$ denotes the index set of point MDPs that were selected based on centroids. 

\begin{figure}[!h]
\centering
\begin{subfigure}{0.4\linewidth}
  \centering
  \includegraphics[width=\linewidth]{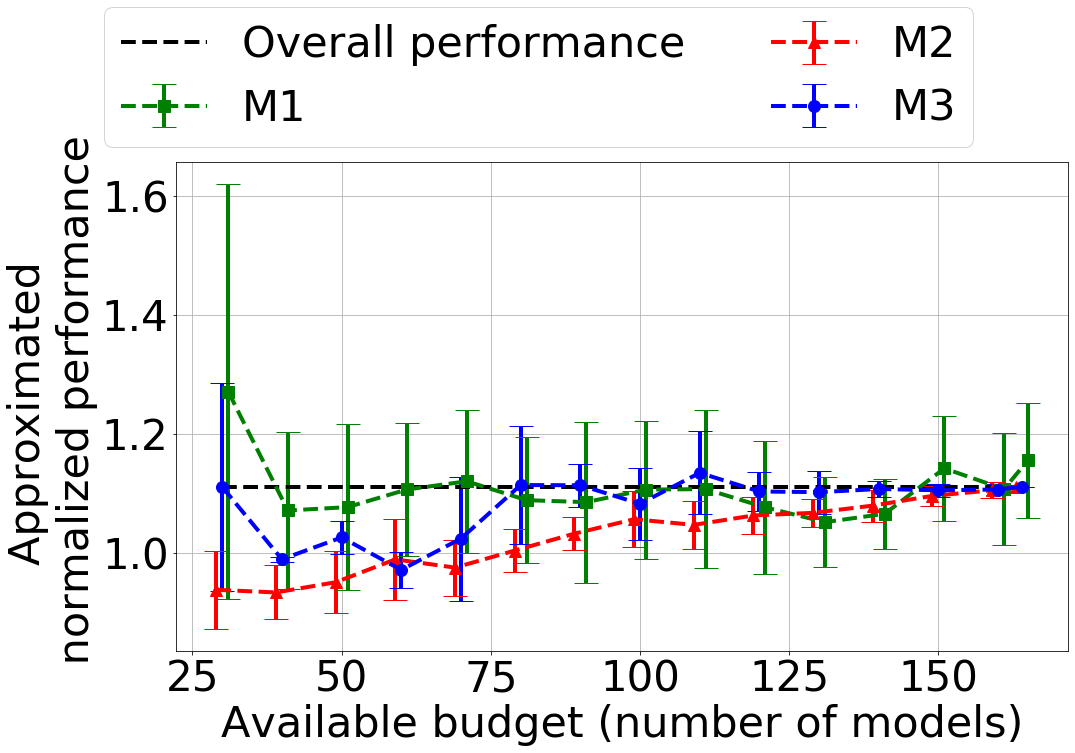}
  \caption{MPLight}
  \label{fig11a}
\end{subfigure}
\begin{subfigure}{0.4\linewidth}
  \centering
  \includegraphics[width=\linewidth]{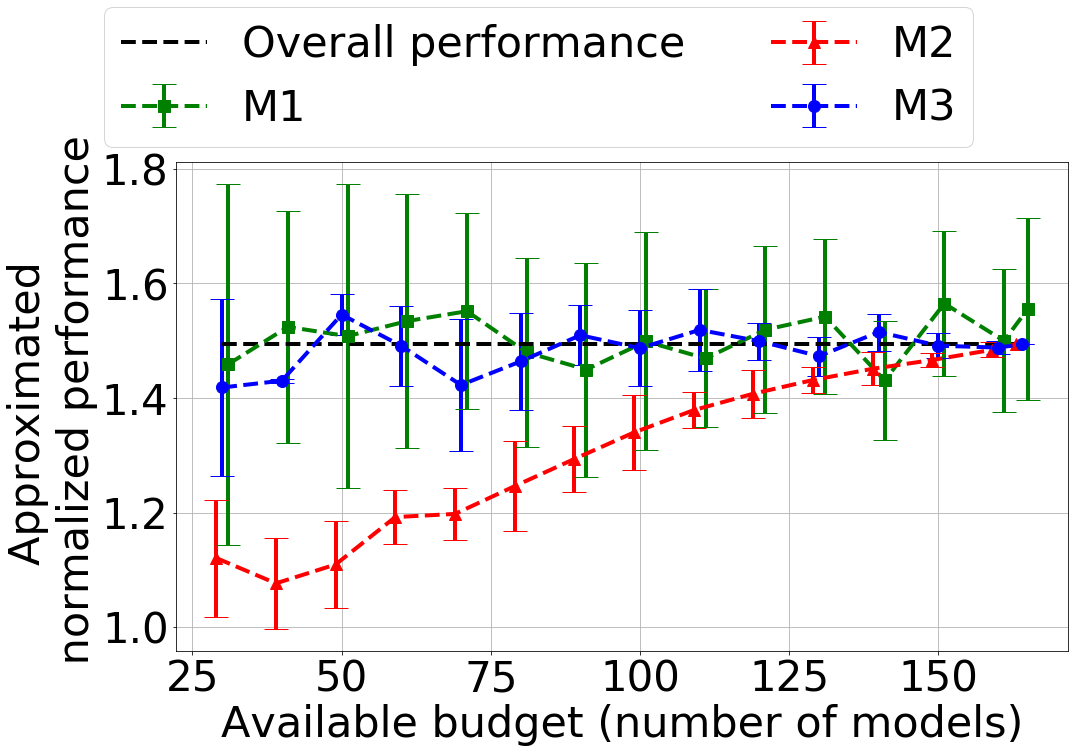}
  \caption{MPLight$^*$}
  \label{fig11b}
\end{subfigure}
\begin{subfigure}{0.4\linewidth}
  \centering
  \includegraphics[width=\linewidth]{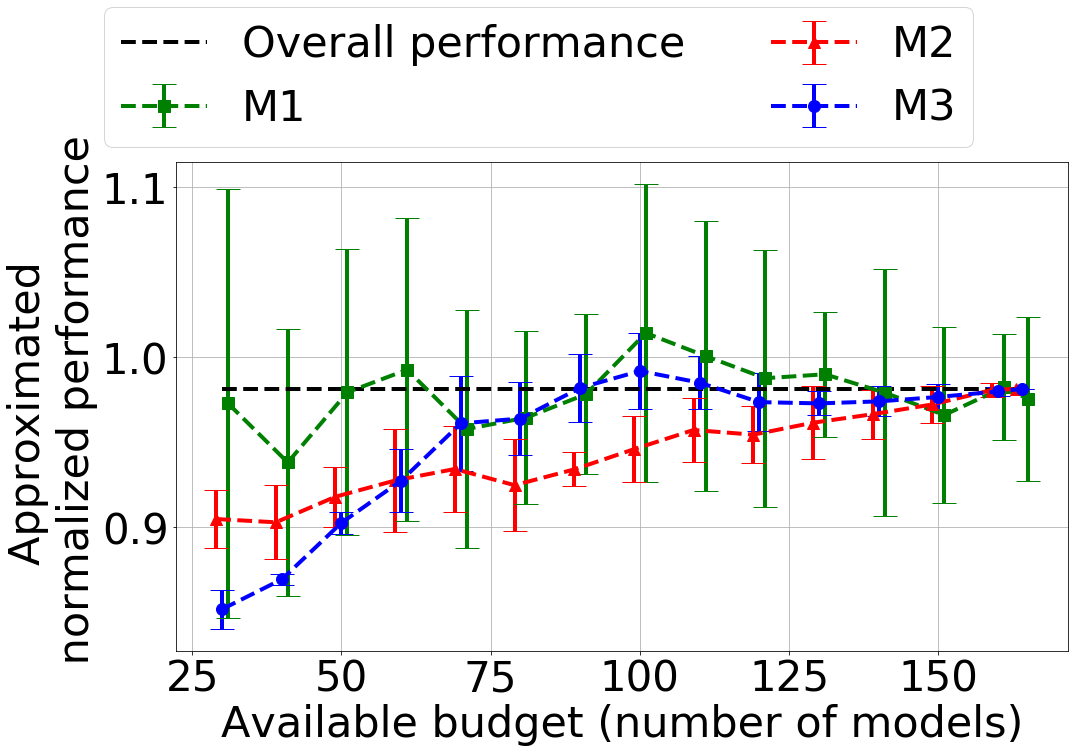}
  \caption{Max Pressure}
  \label{fig11b}
\end{subfigure}
\caption{Evaluations over a family of MDPs in traffic signal control with varying budget limits and sampling techniques. Closer to the overall performance the better.}
\label{sampling-extra}
\end{figure}

In Figure~\ref{sampling-extra}, we show the approximate evaluations of MPLight, MPLight$^*$ and Max Pressure methods in traffic signal control in comparison to the ground-truth performance while varying the budget size. Other related plots are given in Figure~\ref{sampling}.

\begin{figure}[!h]
\centering
\begin{subfigure}{0.4\linewidth}
  \centering
  \includegraphics[width=\linewidth]{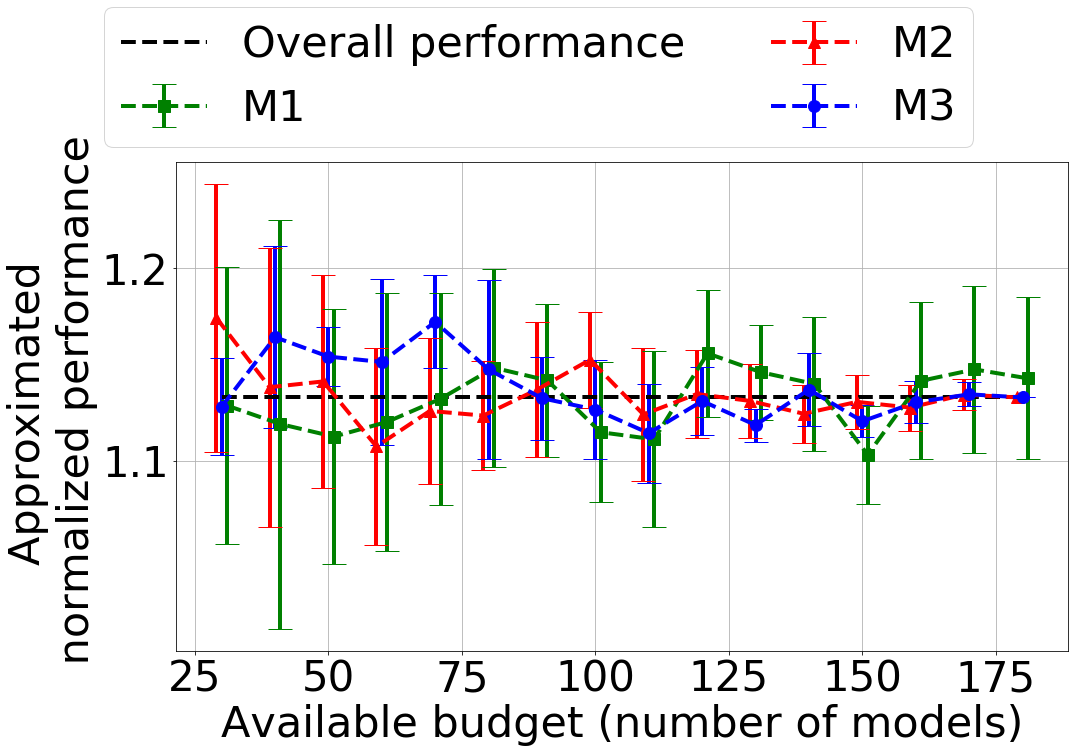}
  \caption{TD3}
  \label{fig11a}
\end{subfigure}
\begin{subfigure}{0.4\linewidth}
  \centering
  \includegraphics[width=\linewidth]{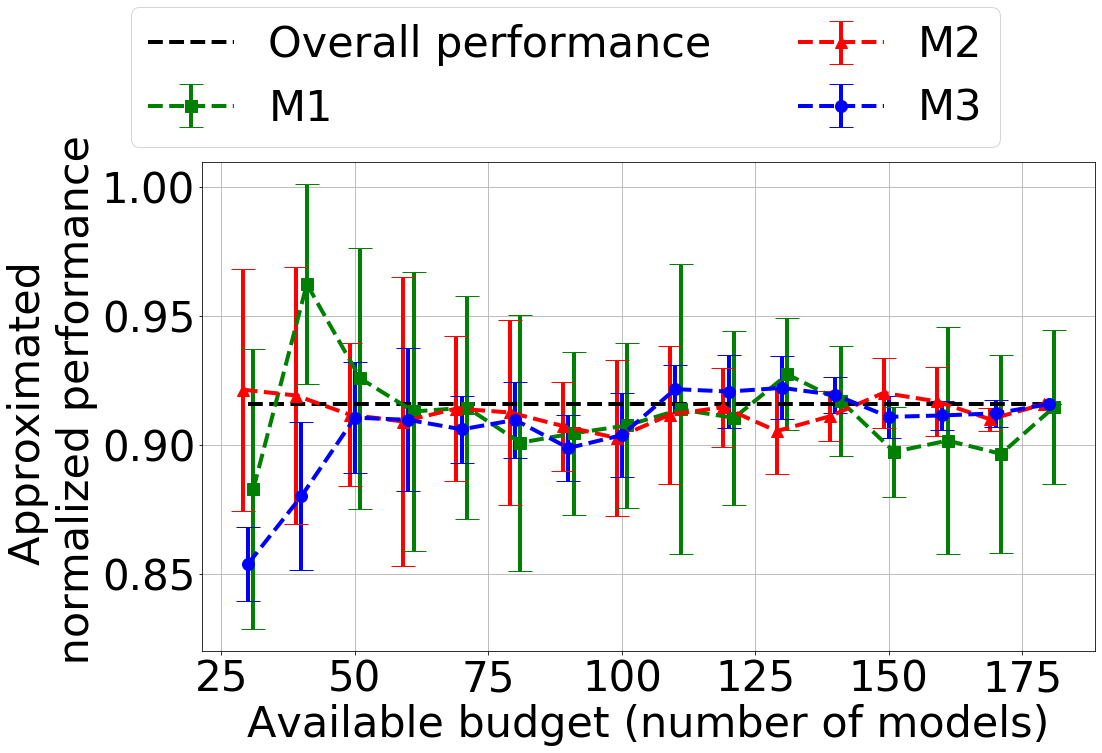}
  \caption{PPO}
  \label{fig11b}
\end{subfigure}
\begin{subfigure}{0.4\linewidth}
  \centering
  \includegraphics[width=\linewidth]{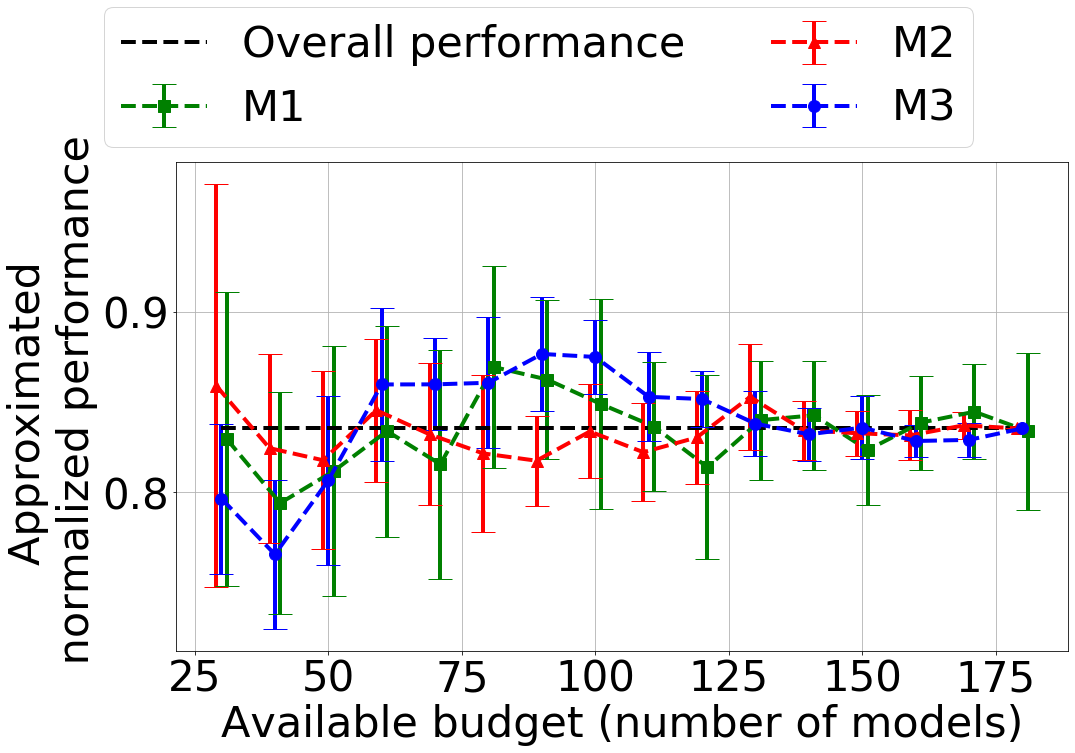}
  \caption{TRPO}
  \label{fig11b}
\end{subfigure}
\begin{subfigure}{0.4\linewidth}
  \centering
  \includegraphics[width=\linewidth]{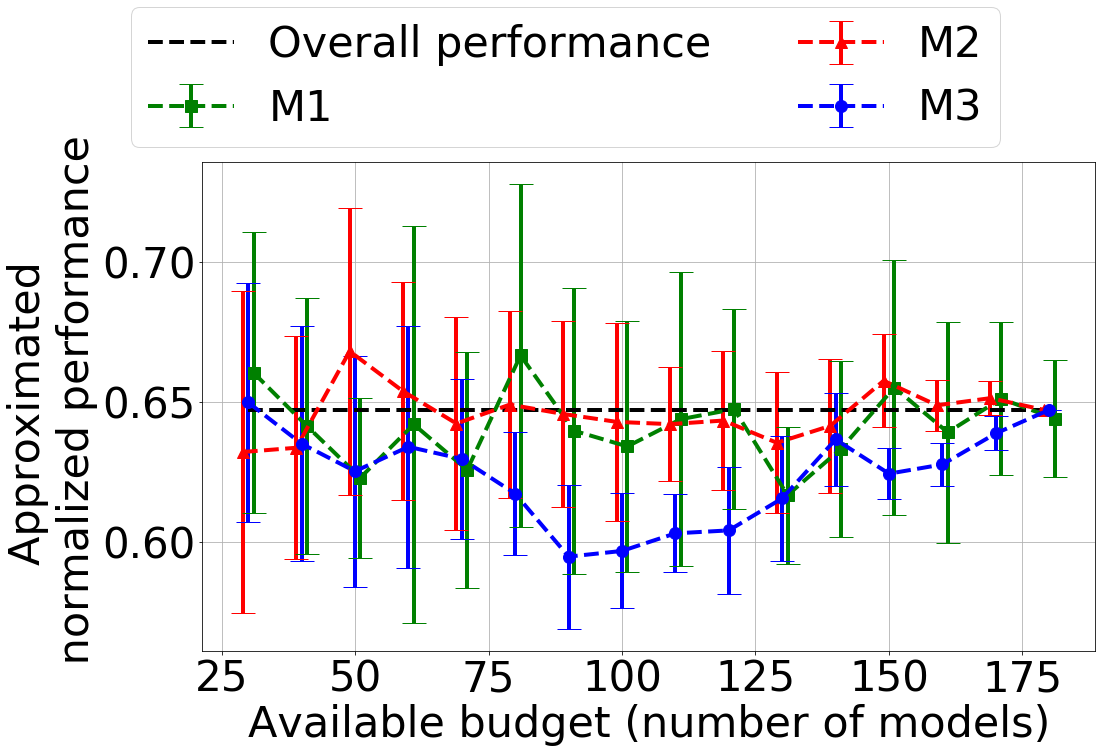}
  \caption{SAC}
  \label{fig11b}
\end{subfigure}
\caption{Evaluations over a family of MDPs in pendulum task with varying budget limits and sampling techniques. Closer to the overall performance the better.}
\label{sampling-pendulum}
\end{figure}

\begin{figure}[!h]
\centering
\begin{subfigure}{0.4\linewidth}
  \centering
  \includegraphics[width=\linewidth]{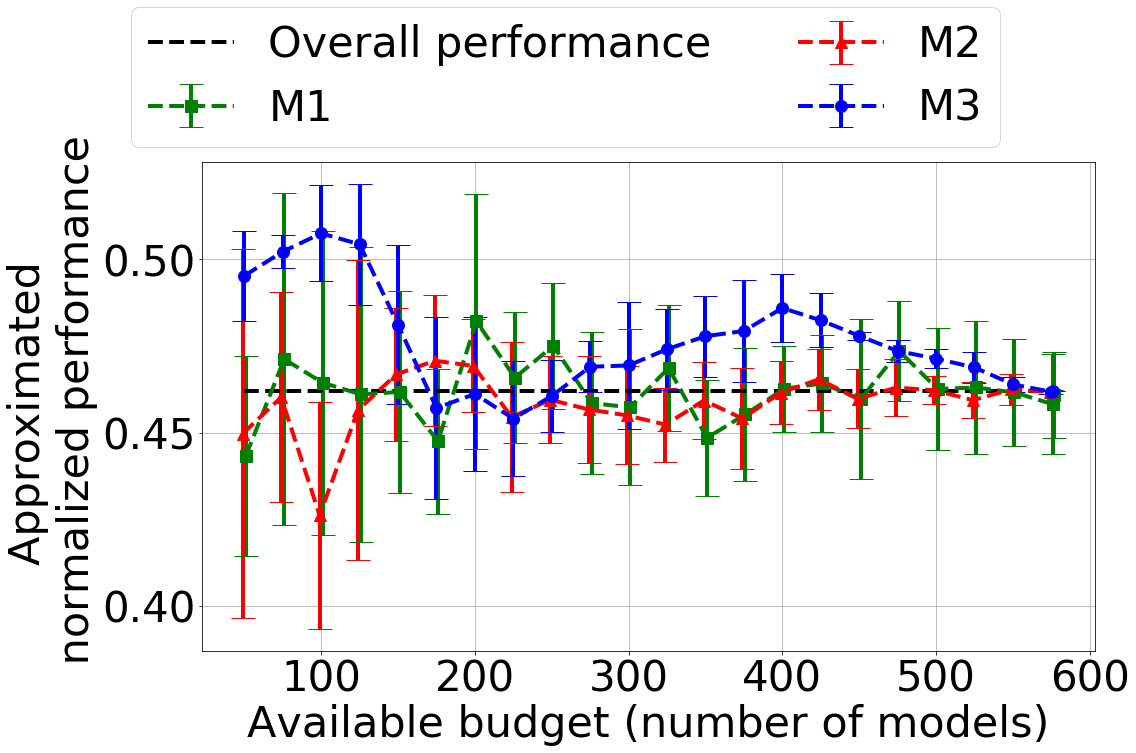}
  \caption{ARS}
  \label{fig11a}
\end{subfigure}
\begin{subfigure}{0.4\linewidth}
  \centering
  \includegraphics[width=\linewidth]{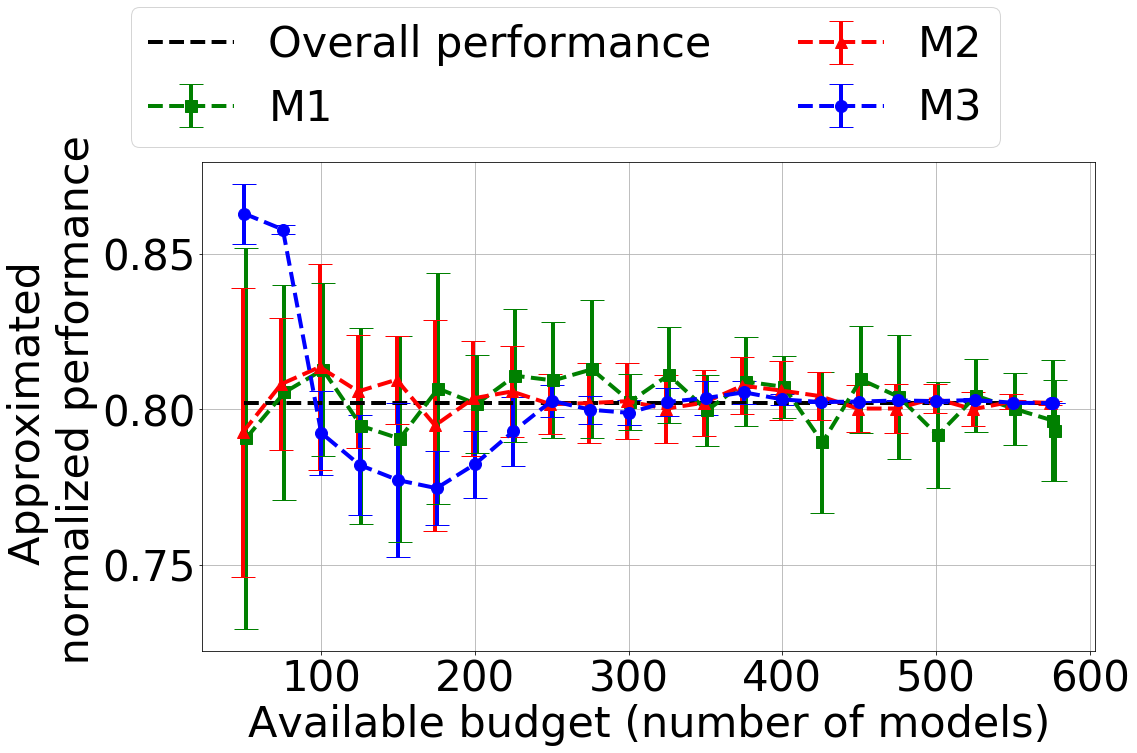}
  \caption{PPO}
  \label{fig11b}
\end{subfigure}
\begin{subfigure}{0.4\linewidth}
  \centering
  \includegraphics[width=\linewidth]{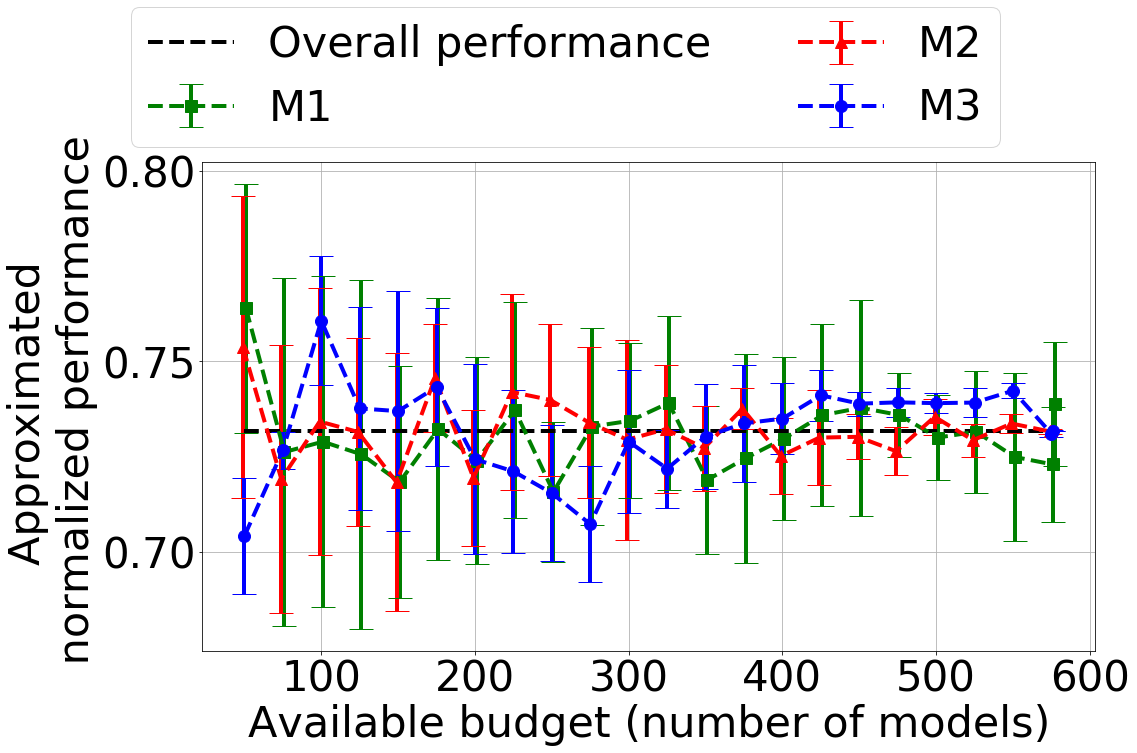}
  \caption{TRPO}
  \label{fig11b}
\end{subfigure}
\begin{subfigure}{0.4\linewidth}
  \centering
  \includegraphics[width=\linewidth]{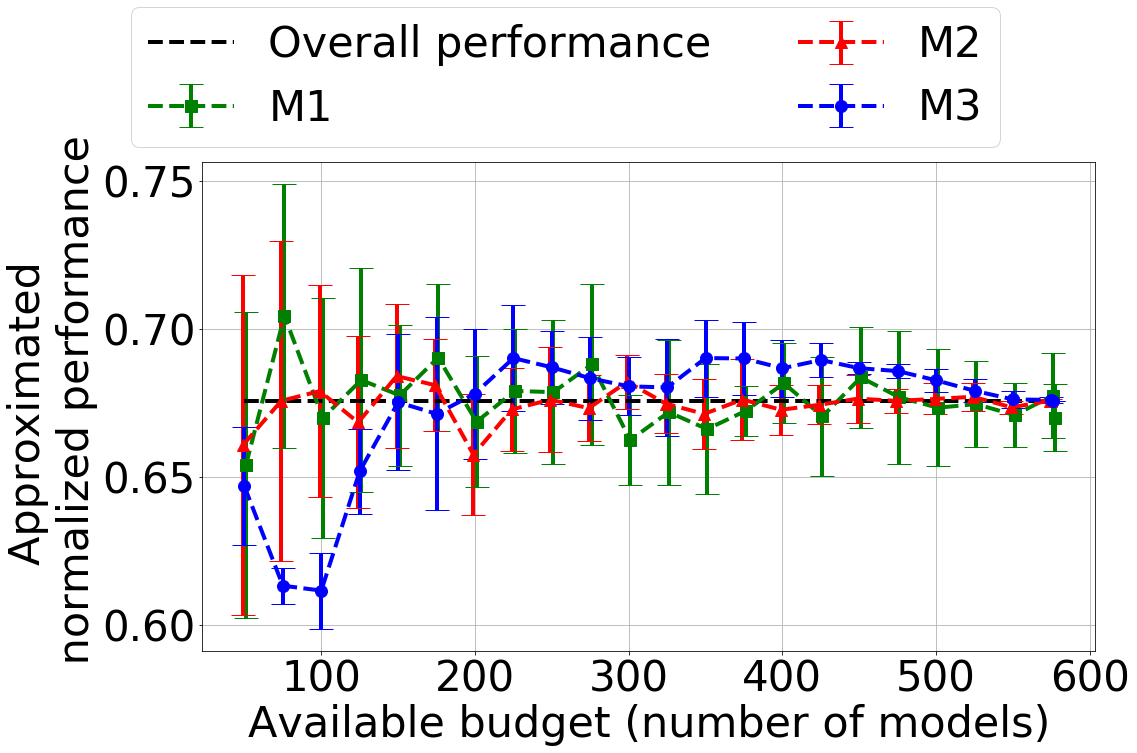}
  \caption{DQN}
  \label{fig11b}
\end{subfigure}
\caption{Evaluations over a family of MDPs in cartpole task with varying budget limits and sampling techniques. Closer to the overall performance the better.}
\label{sampling-cartpole}
\end{figure}

\begin{figure}[!h]
\centering
\begin{subfigure}{0.4\linewidth}
  \centering
  \includegraphics[width=\linewidth]{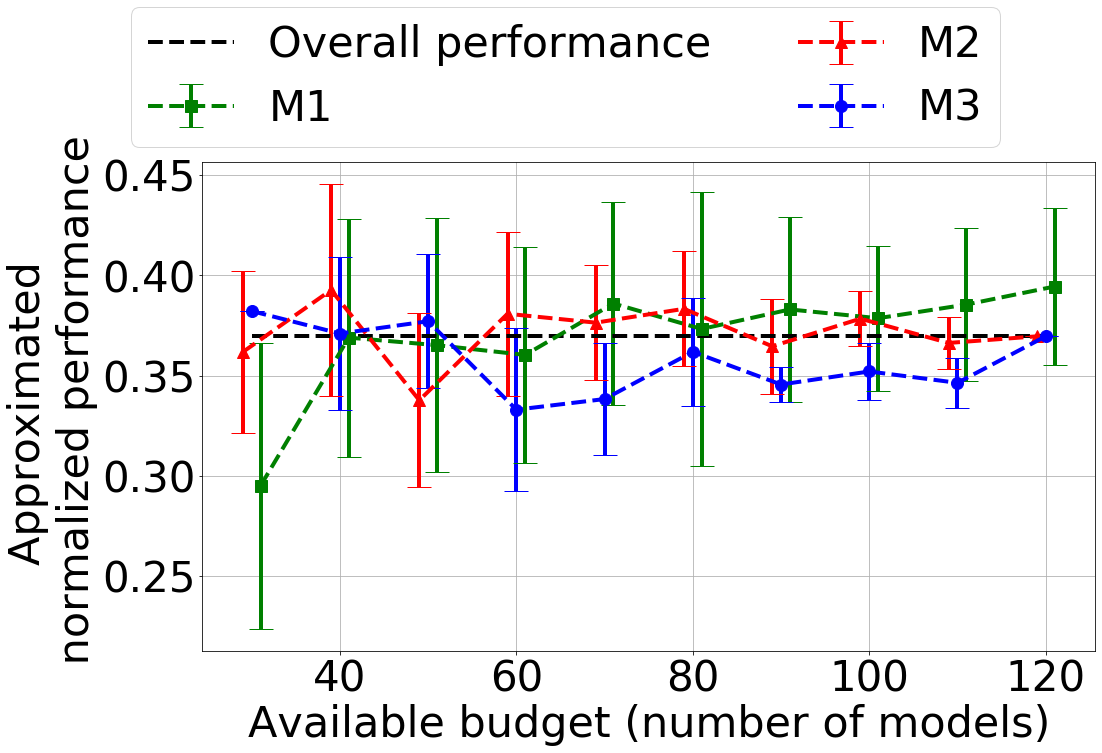}
  \caption{TD3}
  \label{fig11a}
\end{subfigure}
\begin{subfigure}{0.4\linewidth}
  \centering
  \includegraphics[width=\linewidth]{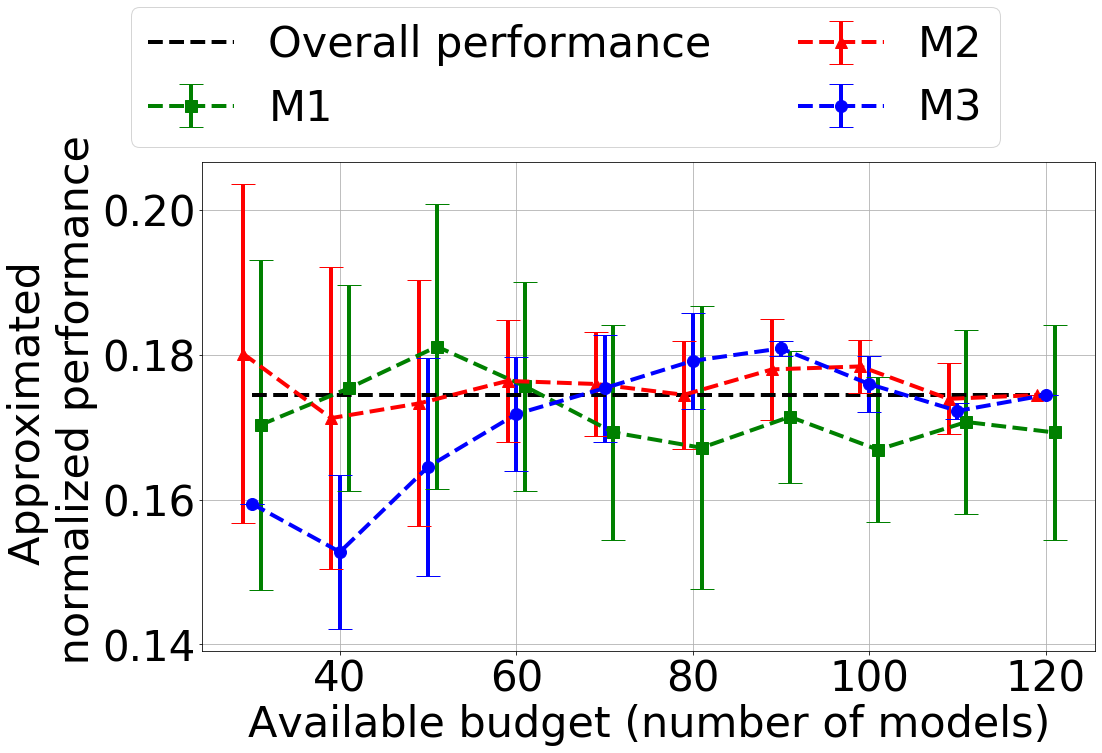}
  \caption{PPO}
  \label{fig11b}
\end{subfigure}
\begin{subfigure}{0.4\linewidth}
  \centering
  \includegraphics[width=\linewidth]{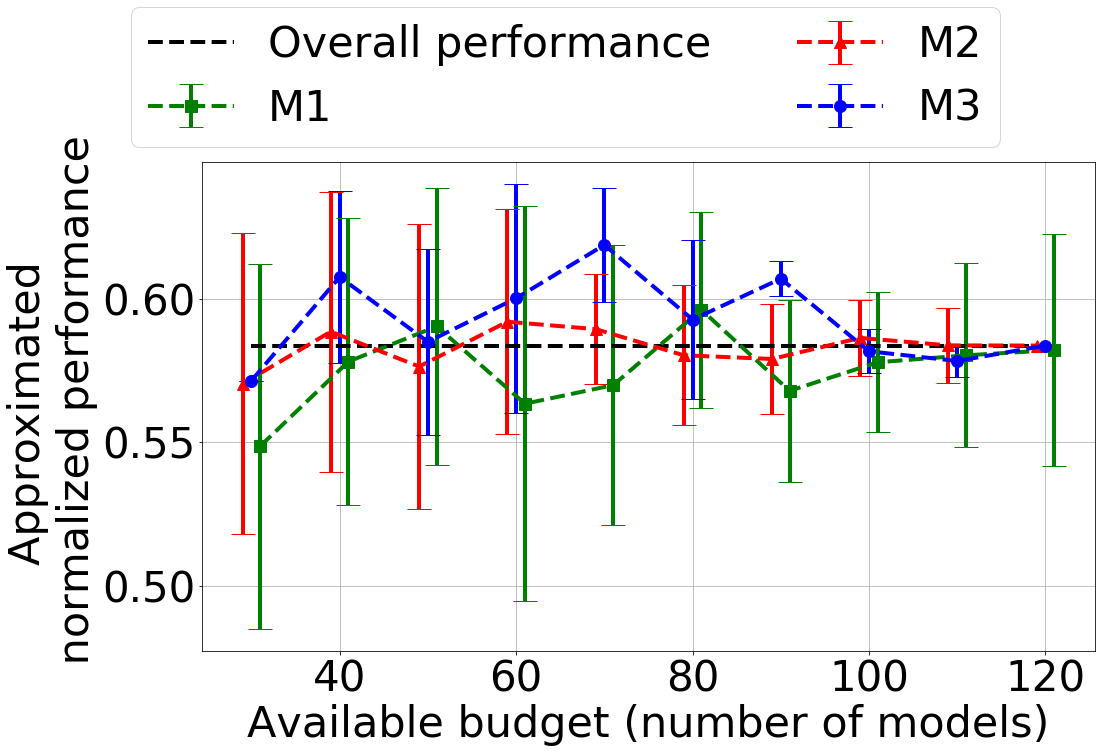}
  \caption{TRPO}
  \label{fig11b}
\end{subfigure}
\begin{subfigure}{0.4\linewidth}
  \centering
  \includegraphics[width=\linewidth]{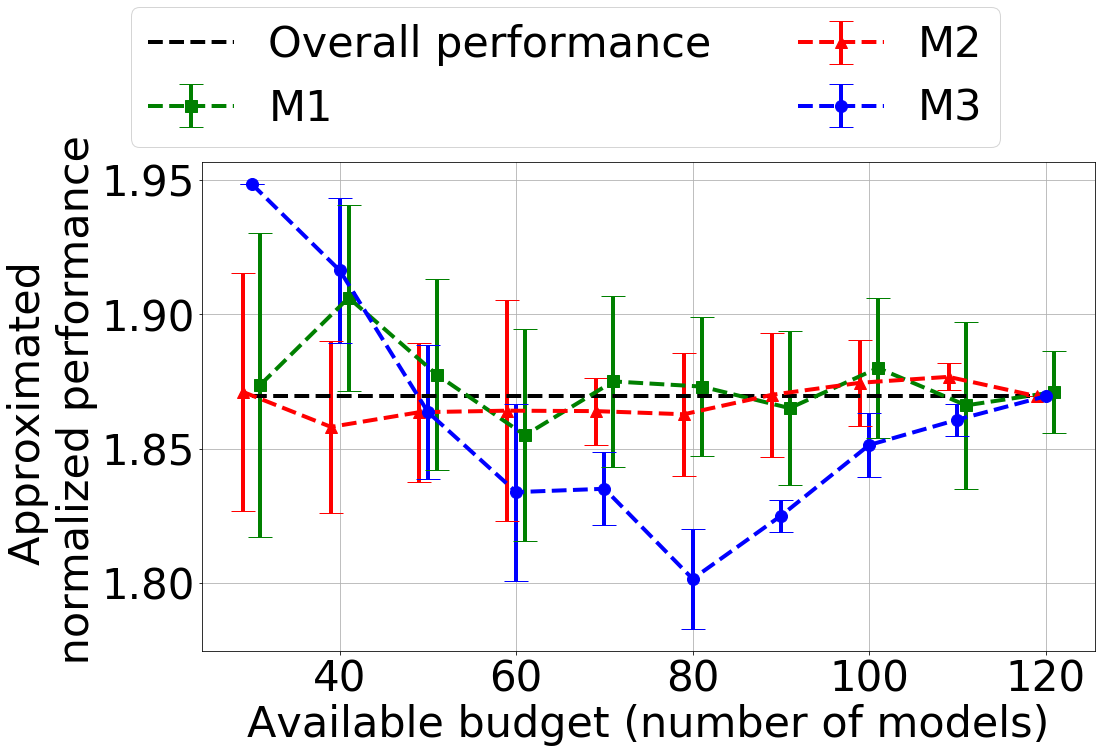}
  \caption{SAC}
  \label{fig11b}
\end{subfigure}
\caption{Evaluations over a family of MDPs in half cheetah task with varying budget limits and sampling techniques. Closer to the overall performance the better.}
\label{sampling-cheetah}
\end{figure}

Figure~\ref{sampling-pendulum}, Figure~\ref{sampling-cartpole} and Figure~\ref{sampling-cheetah} demonstrate the results of performance approximations in pendulum, cartpole, and half cheetah tasks, respectively. In all three control tasks, we see that the random sampling with replacements method has a high variance overall (M1). In comparison, the random sampling without replacements method (M2) and k-means clustering-based method (M3) produce more accurate performance approximations. Unlike in the traffic signal control case study, where the random sampling without replacements method had high variation and significant inaccuracies in approximated performance, here in the three control tasks, we see it demonstrates better performance approximations. K-means clustering-based method consistently produces good approximated performances. We note that even though the random sampling without replacements method produces on average good approximated performances, k-mean based clustering method has comparatively low variance and therefore is preferred for most of the cases (except for a few cases where it seems to have a slightly high error in approximations). Therefore, in general, we recommend using the k-means-based clustering method as it consistently gives us better performance approximations.

\subsection{Further experiments on approximation algorithms}
\label{extra-approximations}

In this section, we perform a sensitivity analysis of the proposed approximation algorithms in Section~\ref{approx_main-content}. Since the overall evaluations are subject to task failures, we investigate the sensitivity of the approximate evaluation methods to the prevalence of ``failure'' point MDPs. In particular, we look at the approximation accuracy as we change the underlying point MDP distribution for the same set of point MDPs used in the traffic signal control case study. We define a failure as any MDP with an average per-vehicle waiting time greater than the 20s. We conduct three sets of experiments. First, we look at the case where the failure MDPs are given low probabilities (compared to the rest of the point MDPs) in the point MDP distribution. Second, we look at the case where the failure MDPs are given high probabilities. Finally, to consider a fairly different distribution, we consider the case where the prevalence of each point MDP is uniformly random.

As a reminder, our notation for the three performance approximation techniques as introduced in Section~\ref{approx_main-content} are (1) \textbf{M1}: random sampling with replacements from the point MDP distribution, (2) \textbf{M2}: random sampling without replacements, and (3) \textbf{M3}: clustering point MDPs using k-means and assigning probability mass of all point MDPs that belong to same cluster to its centroid. Here, $k$ defines the number of point MDPs used for approximate evaluations. Further details of each method can be found in Appendix~\ref{performance approximations}. 

\subsubsection{Case 1: Failure point MDPs get low probabilities.}

\begin{figure*}[h]
\centering
\begin{subfigure}{0.32\linewidth}
  \centering
  \includegraphics[width=\linewidth]{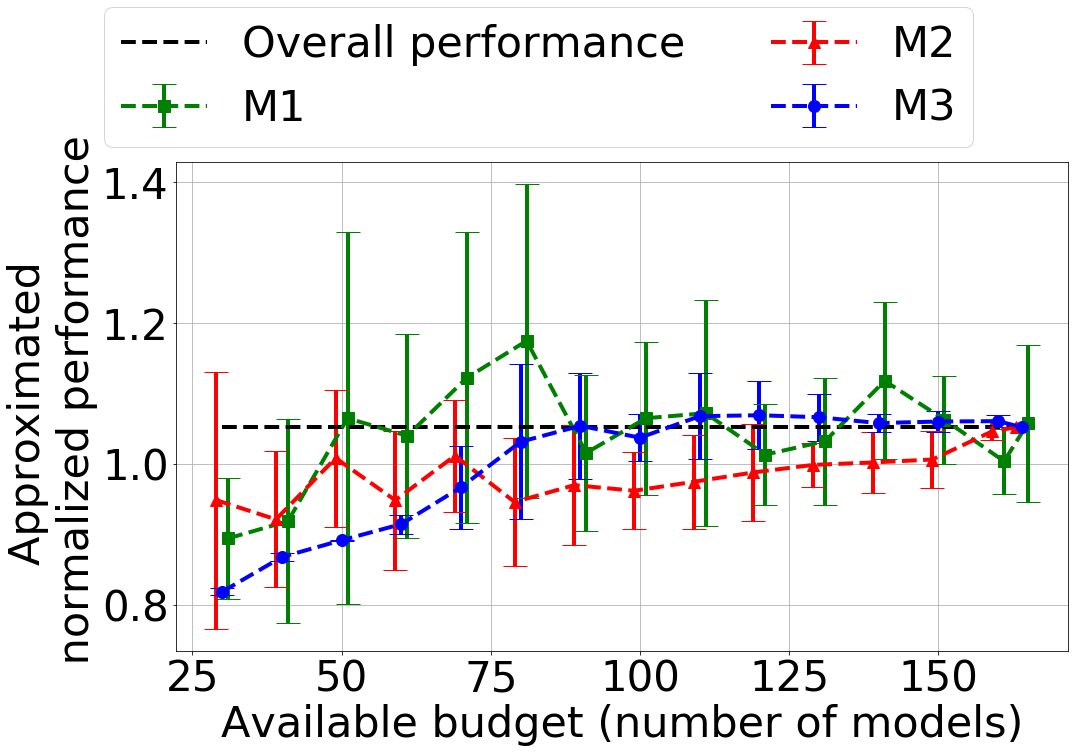}
  \caption{IDQN}
  \label{fig11a}
\end{subfigure}
\begin{subfigure}{0.32\linewidth}
  \centering
  \includegraphics[width=\linewidth]{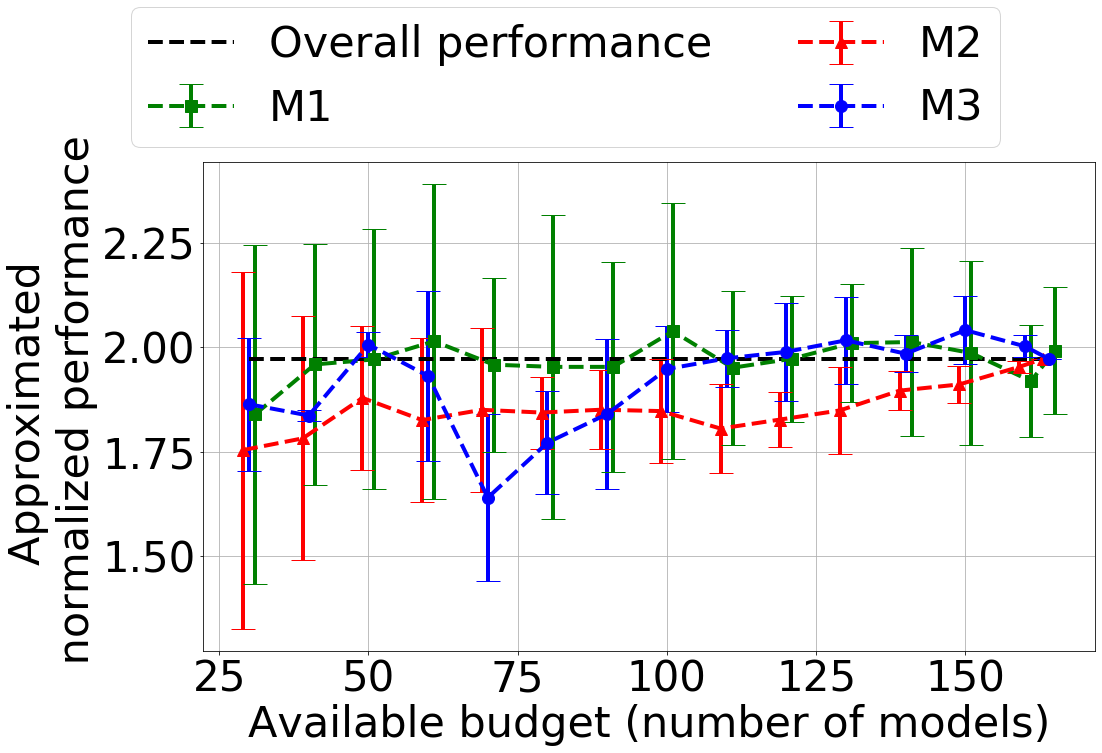}
  \caption{IPPO}
  \label{fig11b}
\end{subfigure}
\begin{subfigure}{0.32\linewidth}
  \centering
  \includegraphics[width=\linewidth]{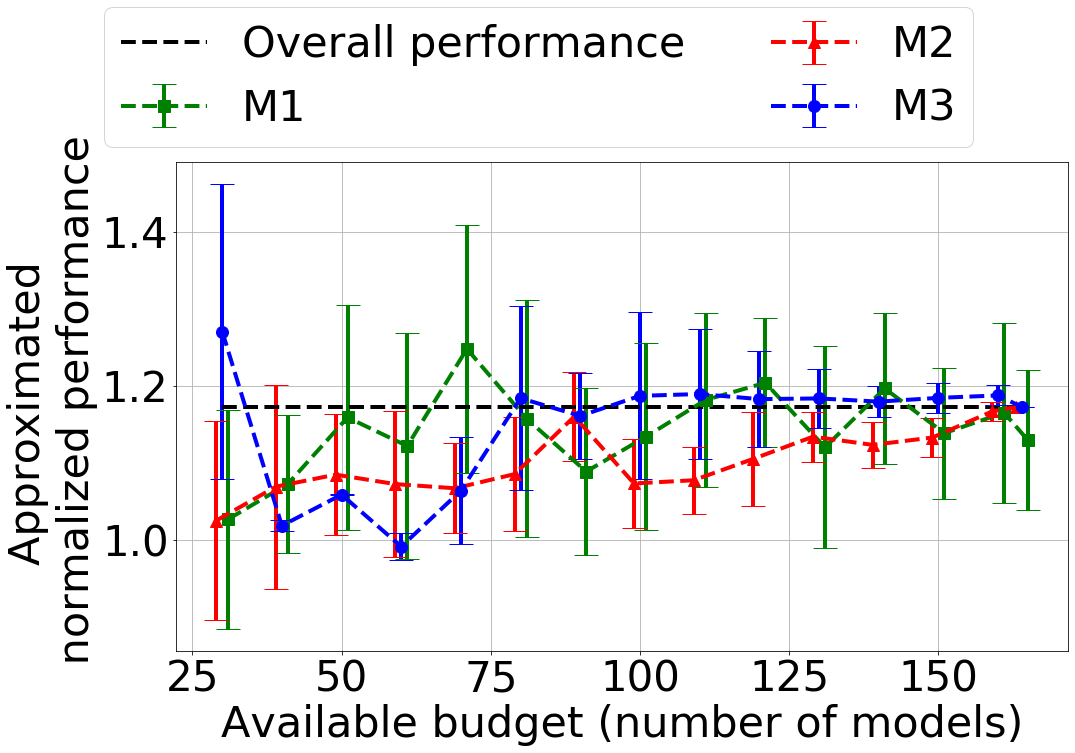}
  \caption{MPLight}
  \label{fig11b}
\end{subfigure}
\begin{subfigure}{0.32\linewidth}
  \centering
  \includegraphics[width=\linewidth]{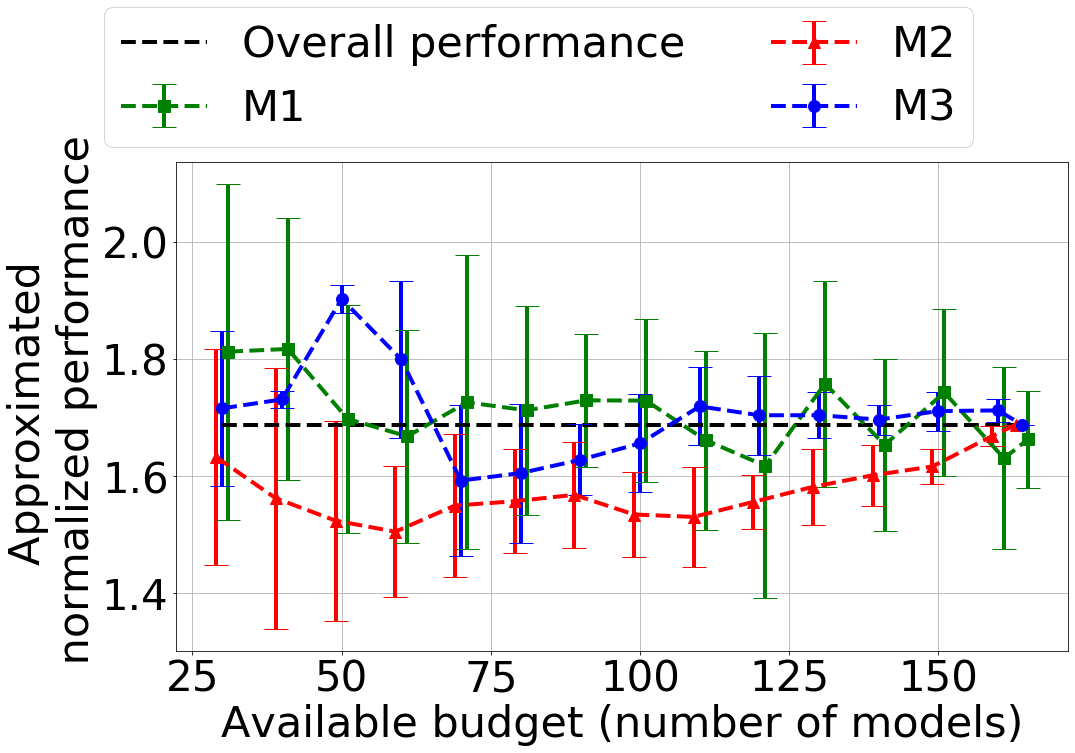}
  \caption{MPLight$^*$}
  \label{fig11b}
\end{subfigure}
\begin{subfigure}{0.32\linewidth}
  \centering
  \includegraphics[width=\linewidth]{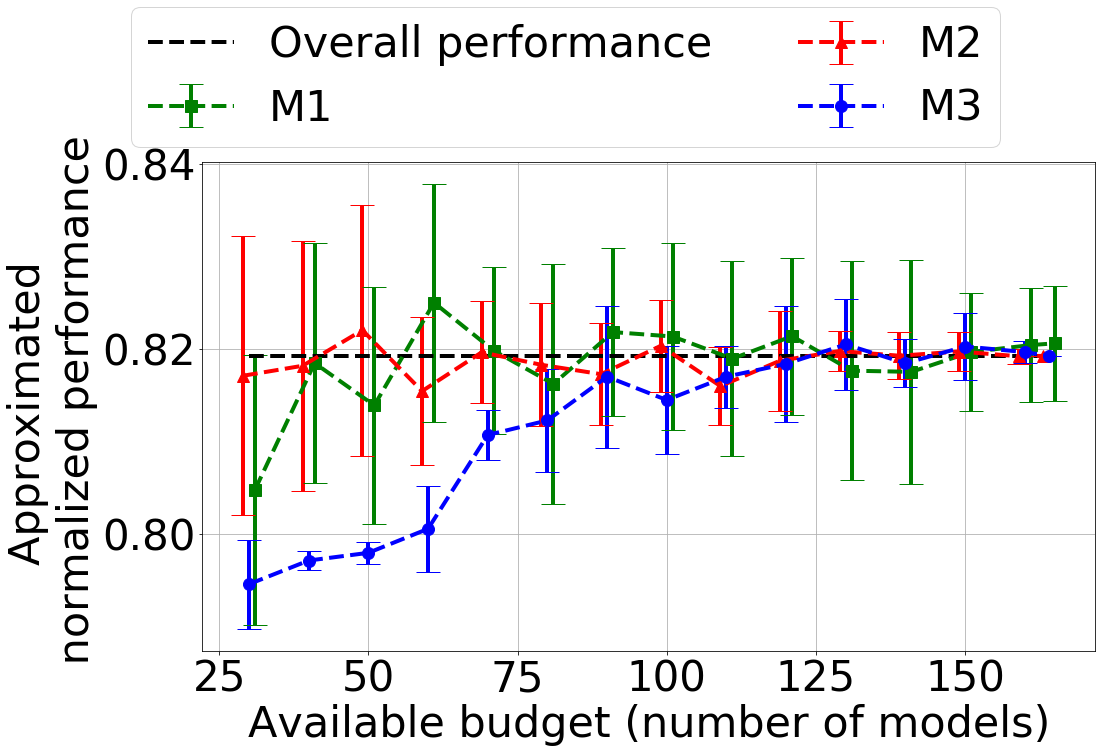}
  \caption{Fixed Time}
  \label{fig11b}
\end{subfigure}
\begin{subfigure}{0.32\linewidth}
  \centering
  \includegraphics[width=\linewidth]{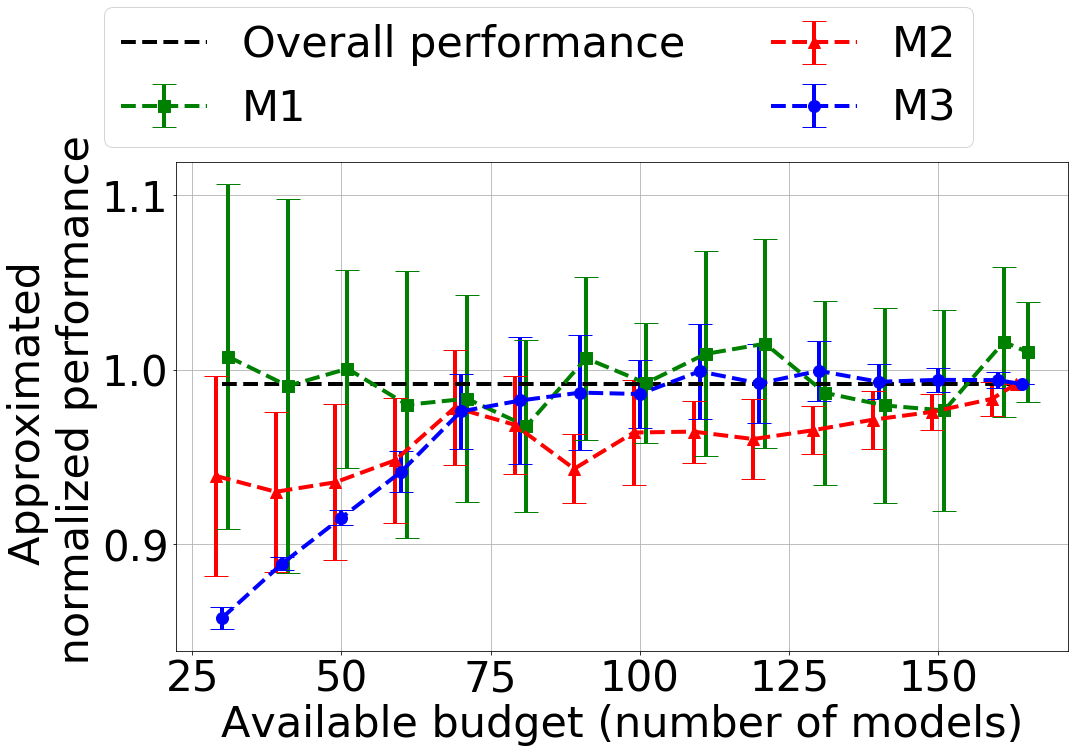}
  \caption{Max Pressure}
  \label{fig11b}
\end{subfigure}
\caption{Evaluations over a family of MDPs in traffic signal control with varying budget limits and sampling techniques when the failure point MDPs are assigned low probabilities than other point MDPs. Closer to the overall performance the better. }
\label{one-approx}
\end{figure*}

Figure~\ref{one-approx} denotes the performance approximations for each of the three methods with varying budget sizes when the failure point MDPs get low probabilities. It is clear from the figure that the k-means-based clustering approach performs the best in this case, specifically when the budget size is greater than half the size of total point MDPs. Random sampling with replacements has an acceptable mean performance but with a high variance. Finally, random sampling without replacements often overestimates the performance unless the budget size is closer to the total point MDPs count (note: lower the normalize score the better).

\subsubsection{Case 2: Failure MDPs get high probabilities.}
\begin{figure*}[h]
\centering
\begin{subfigure}{0.32\linewidth}
  \centering
  \includegraphics[width=\linewidth]{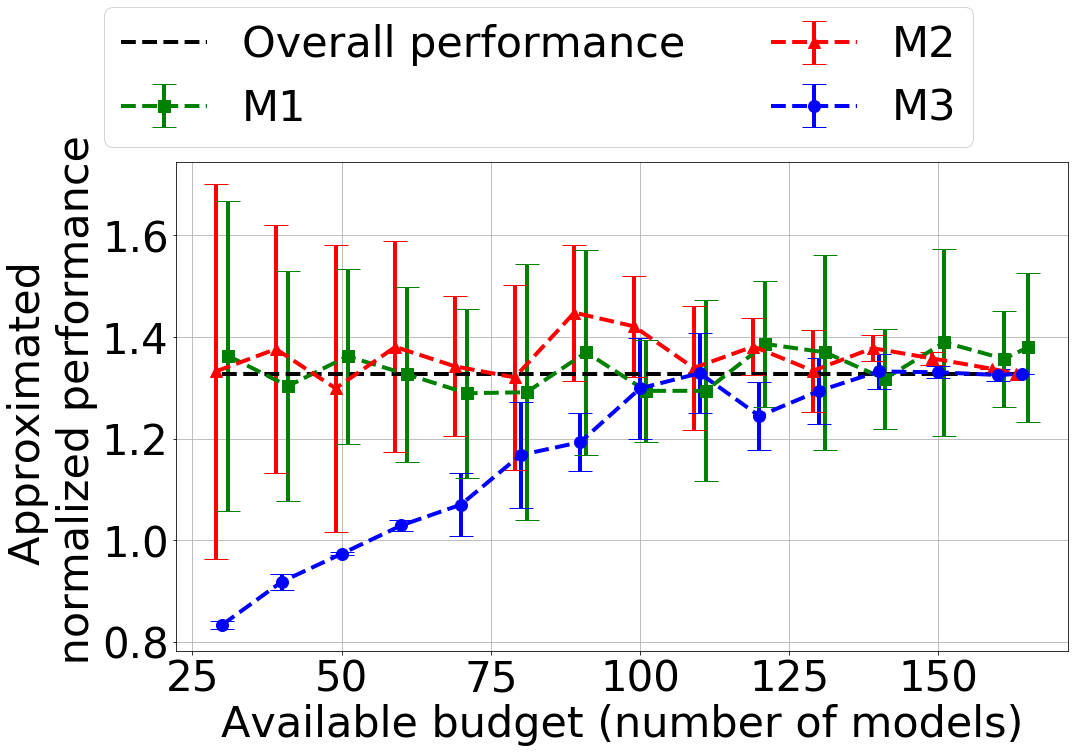}
  \caption{IDQN}
  \label{fig11a}
\end{subfigure}
\begin{subfigure}{0.32\linewidth}
  \centering
  \includegraphics[width=\linewidth]{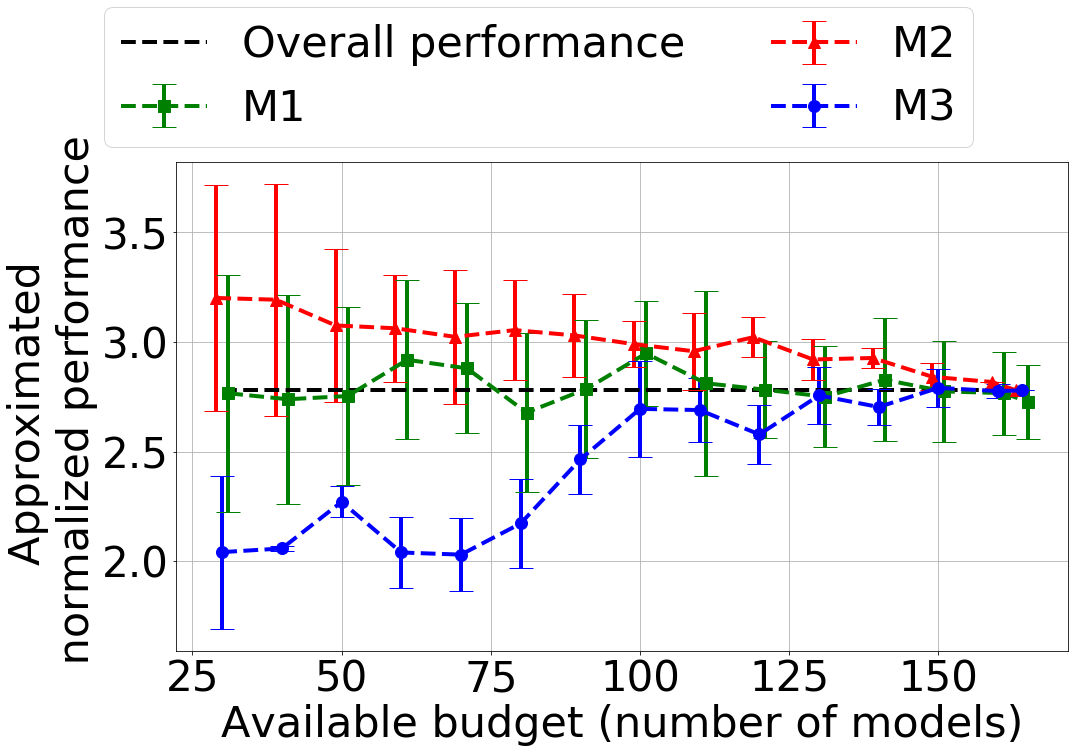}
  \caption{IPPO}
  \label{fig11b}
\end{subfigure}
\begin{subfigure}{0.32\linewidth}
  \centering
  \includegraphics[width=\linewidth]{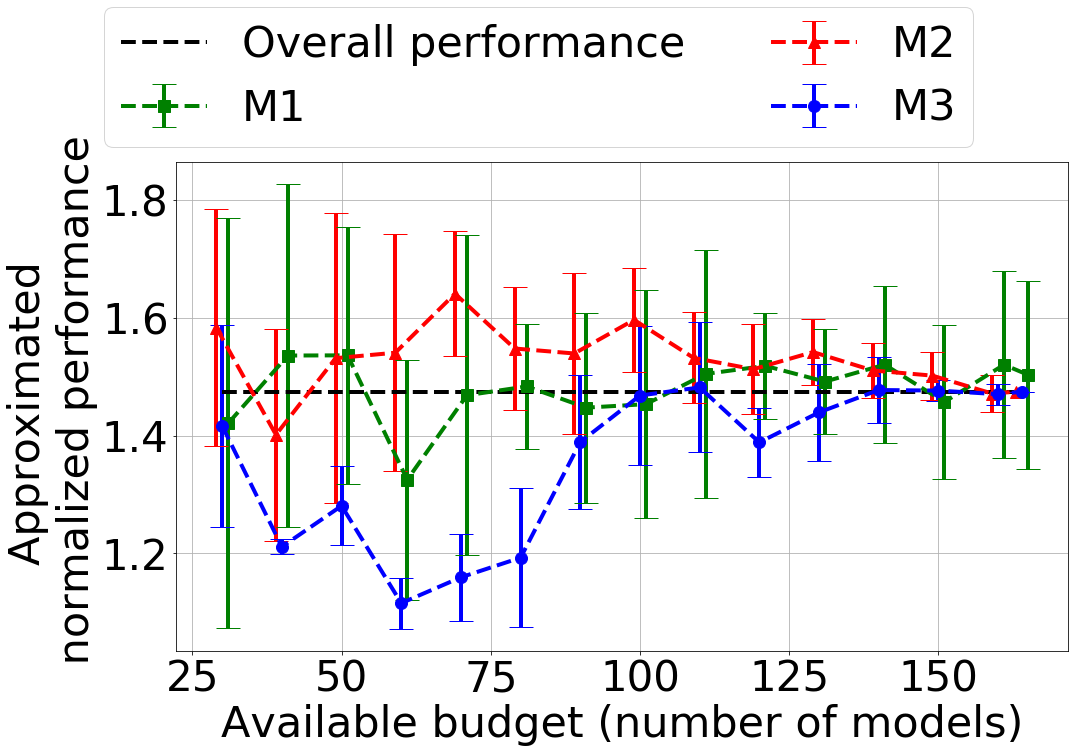}
  \caption{MPLight}
  \label{fig11b}
\end{subfigure}
\begin{subfigure}{0.32\linewidth}
  \centering
  \includegraphics[width=\linewidth]{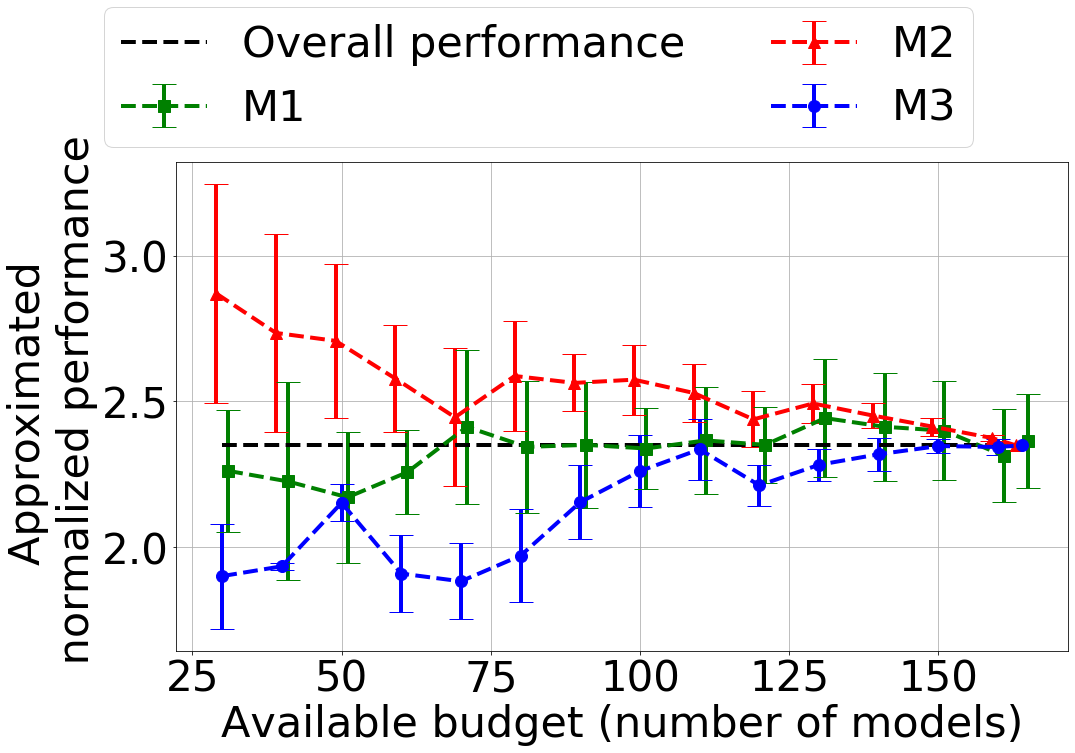}
  \caption{MPLight$^*$}
  \label{fig11b}
\end{subfigure}
\begin{subfigure}{0.32\linewidth}
  \centering
  \includegraphics[width=\linewidth]{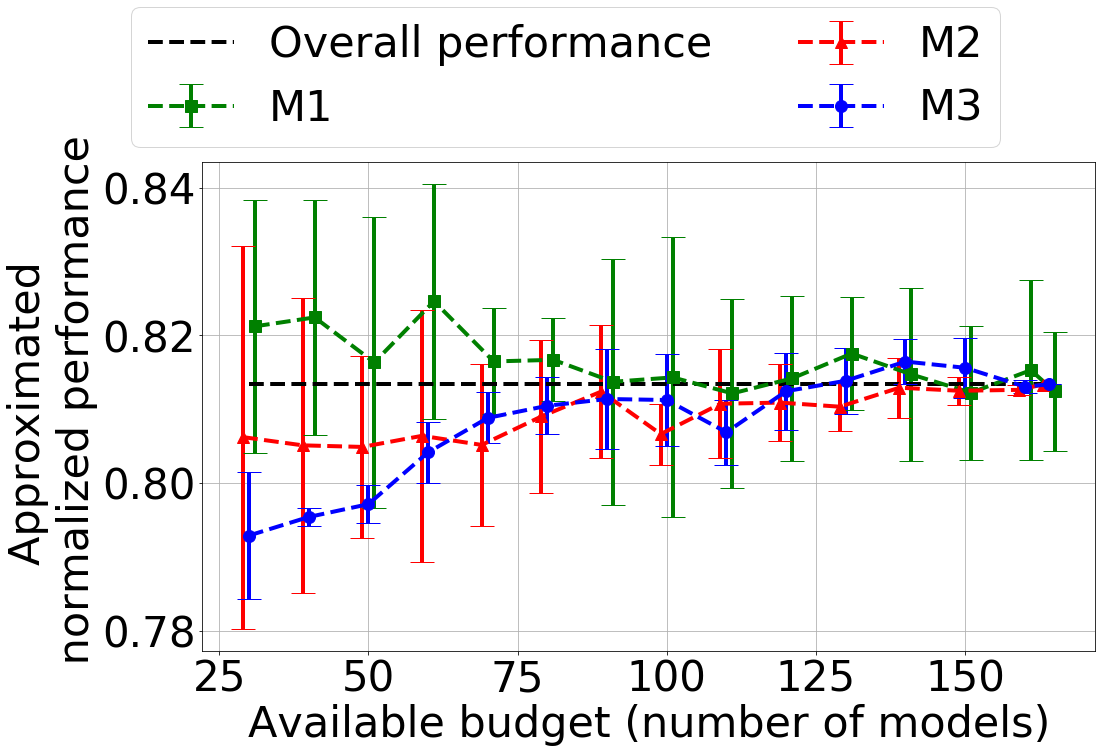}
  \caption{Fixed Time}
  \label{fig11b}
\end{subfigure}
\begin{subfigure}{0.32\linewidth}
  \centering
  \includegraphics[width=\linewidth]{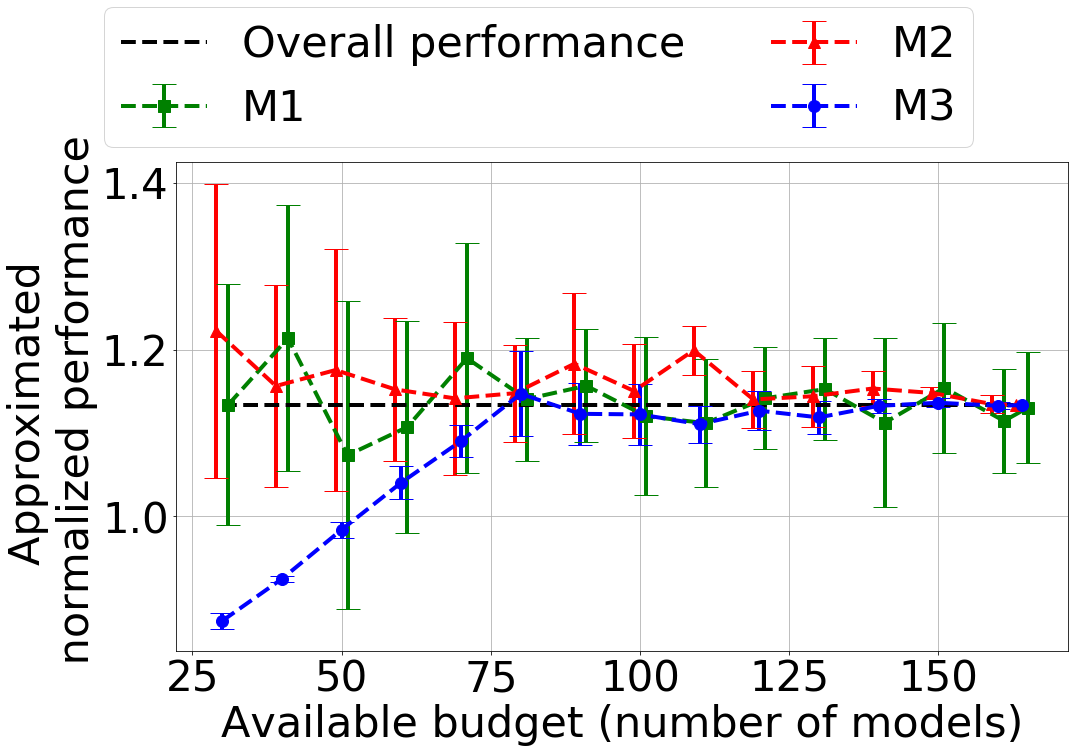}
  \caption{Max Pressure}
  \label{fig11b}
\end{subfigure}
\caption{Evaluations over a family of MDPs in traffic signal control with varying budget limits and sampling techniques when the failure point MDPs are assigned a high probabilities than other point MDPs. Closer to the overall performance the better.}
\label{two-approx}
\end{figure*}

As the second case, we look at the case where failure MDPs get high probabilities than the rest of the point MDPs in Figure~\ref{two-approx}. While k-means-based clustering may underestimate the performance if the budget size is small, it still is the best method when the budget size is greater than half the total MDP count. Random sampling with the replacement method seems to work quite well in this case. Specifically, when the budget size is small, it performs better than random sampling without the replacement method, which always seems to underestimate the performance (note: lower the normalize score the better). 

\subsubsection{Case 3: Point MDP prevalence is uniformly random.}

\begin{figure*}[h]
\centering
\begin{subfigure}{0.32\linewidth}
  \centering
  \includegraphics[width=\linewidth]{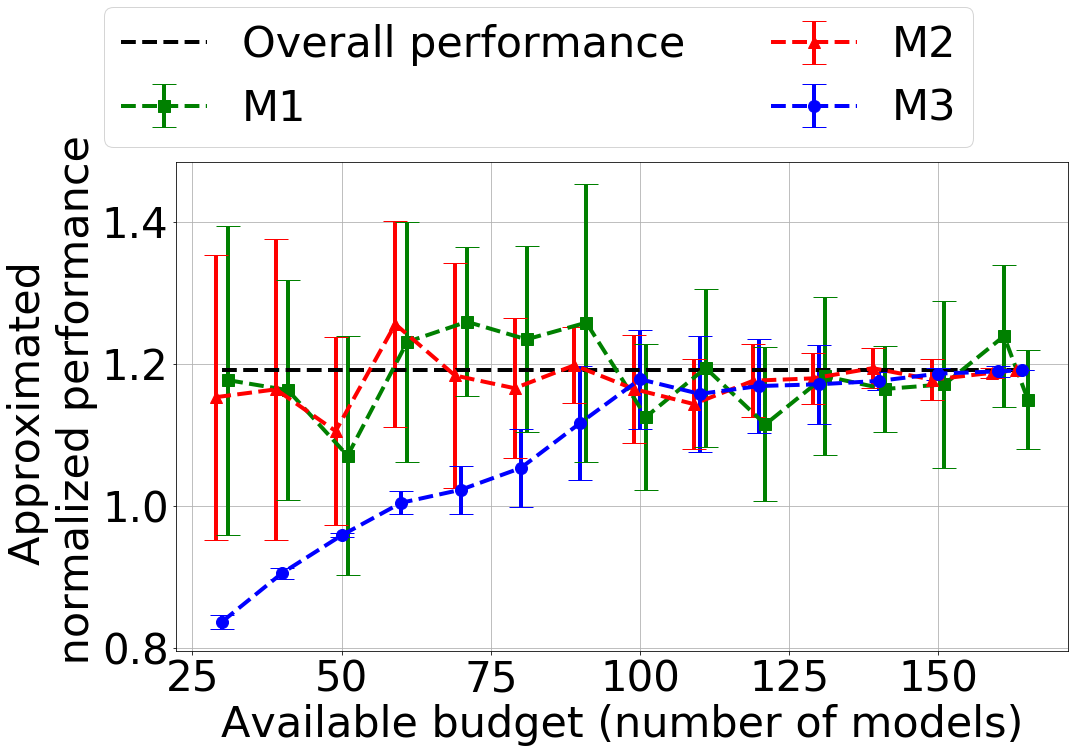}
  \caption{IDQN}
  \label{fig11a}
\end{subfigure}
\begin{subfigure}{0.32\linewidth}
  \centering
  \includegraphics[width=\linewidth]{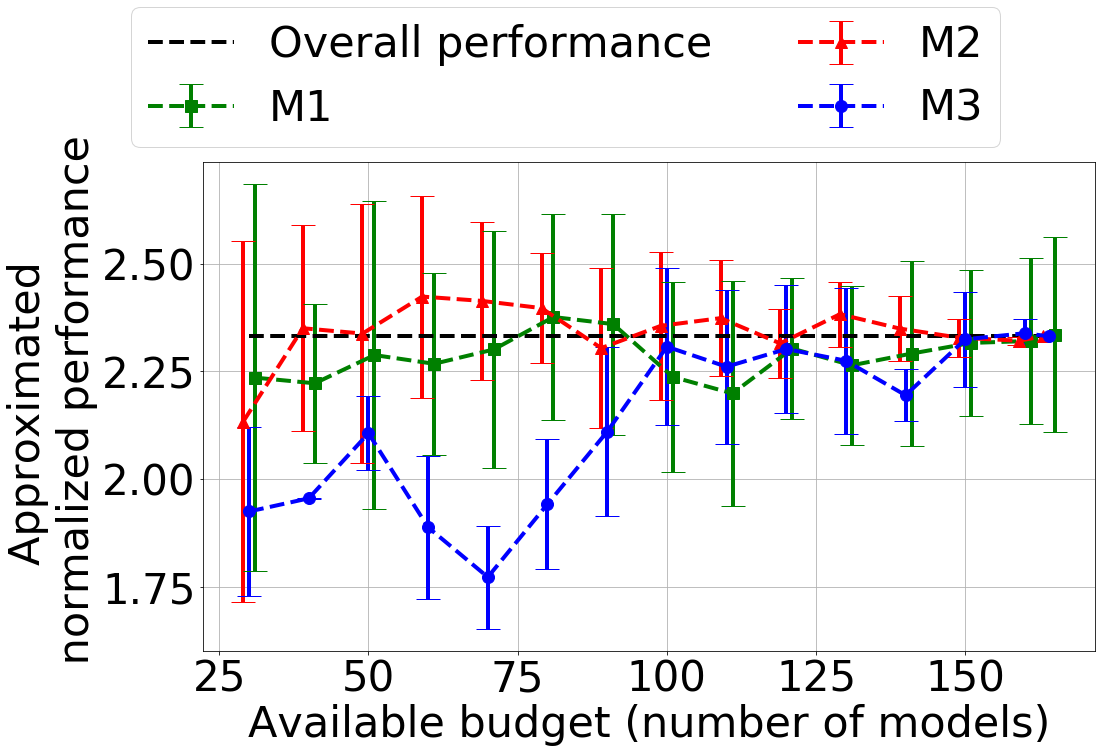}
  \caption{IPPO}
  \label{fig11b}
\end{subfigure}
\begin{subfigure}{0.32\linewidth}
  \centering
  \includegraphics[width=\linewidth]{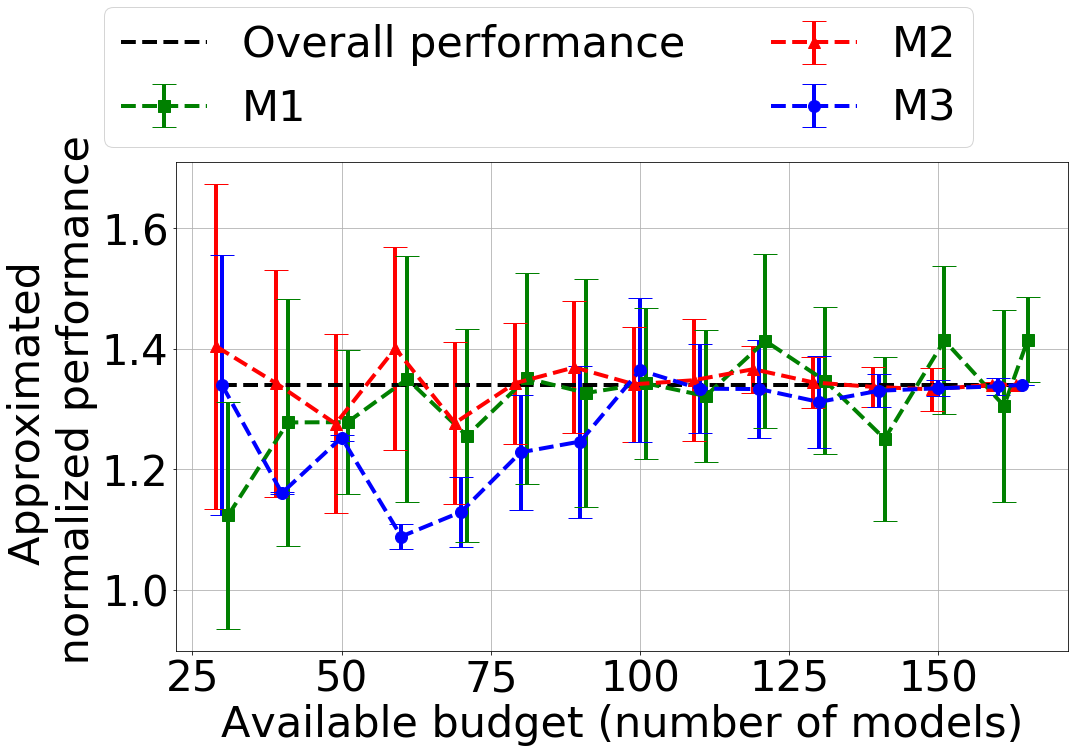}
  \caption{MPLight}
  \label{fig11b}
\end{subfigure}
\begin{subfigure}{0.32\linewidth}
  \centering
  \includegraphics[width=\linewidth]{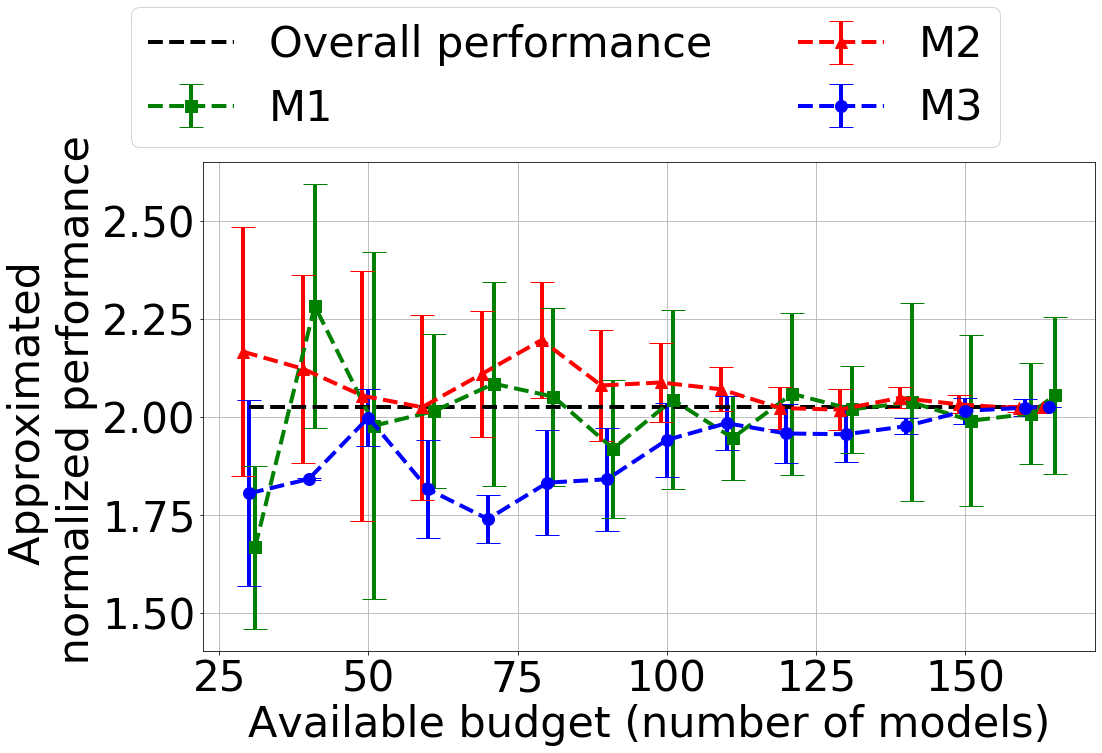}
  \caption{MPLight$^*$}
  \label{fig11b}
\end{subfigure}
\begin{subfigure}{0.32\linewidth}
  \centering
  \includegraphics[width=\linewidth]{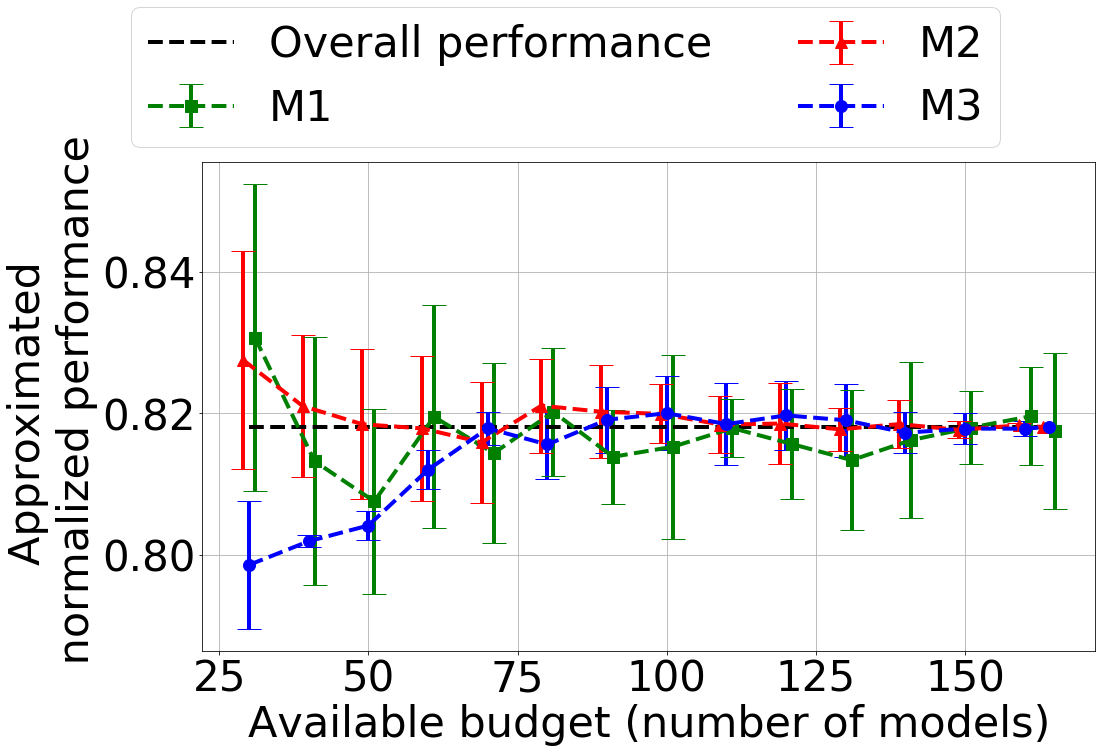}
  \caption{Fixed Time}
  \label{fig11b}
\end{subfigure}
\begin{subfigure}{0.32\linewidth}
  \centering
  \includegraphics[width=\linewidth]{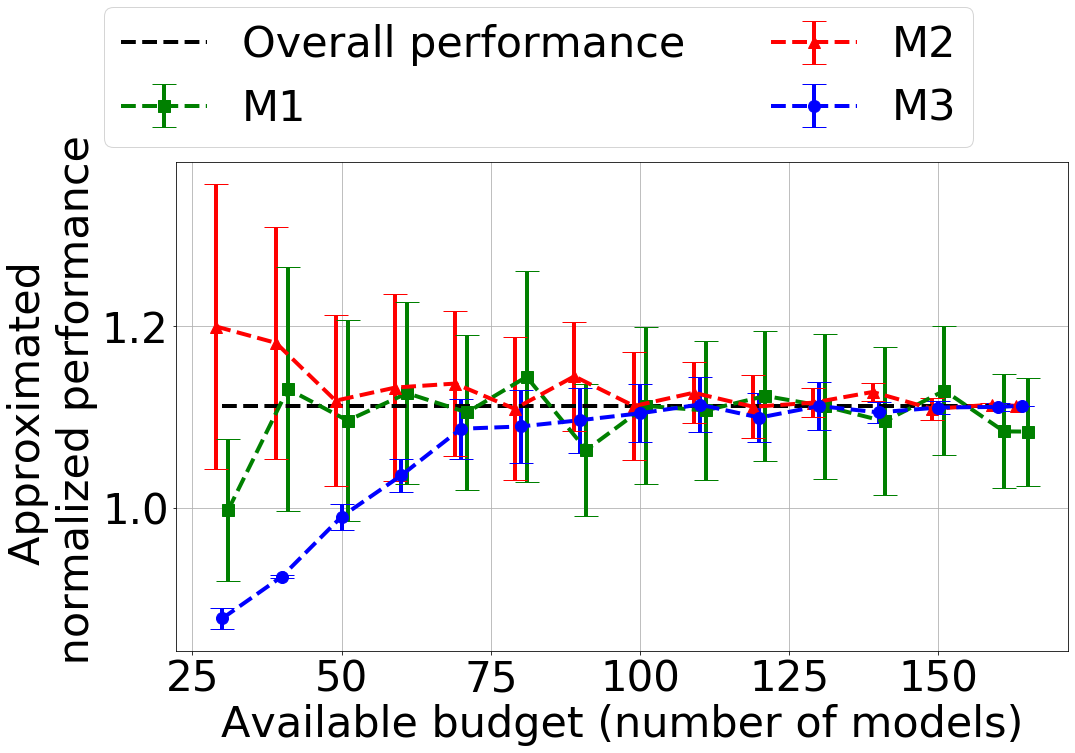}
  \caption{Max Pressure}
  \label{fig11b}
\end{subfigure}
\caption{Evaluations over a family of MDPs in traffic signal control with varying budget limits and sampling techniques when the support of the point MDP distribution is a discrete normal distribution. Closer to the overall performance the better.}
\label{three-approx}
\end{figure*}

In Figure~\ref{three-approx}, we denote the performance estimates when the prevalence of each point MDP is uniformly random. Here we observe an interesting result. The random sampling without replacements method performs quite well under this setting, yielding low variance. As in the other cases, the k-means-based clustering method performs well when the budget size is greater than half the total point MDP count. The random sampling with replacements method performs as usual but has a high variance. 

Overall, we see that the k-means-based clustering method (M3) performs consistently across three cases when the budget size is at least half the total MDP count. Random sampling with replacements (M1) method in expectation provides a reasonable performance estimate but may have a high variance. Random sampling without replacements method (M2) only performs well when all point MDPs are given a fair chance of getting any probability. It is clear that failure MDPs have an overall impact on which method suits best for a given use case. However, there is not enough information to know which point MDPs are failure MDPs as a prior because it poses a chicken-and-egg problem, as we pointed out in Section~\ref{formalism}. Therefore, in general, the k-means clustering-based method (M3) is recommended. However, k-means clustering can be inefficient when the point MDP distribution has a higher number of dimensions (since k-means is not robust to high dimensional data). If that is the case, we recommend using random sampling with replacements (M1). However, domain experts may have insights into which point MDPs are more susceptible to failure (e.g., in traffic signal control, intersections with short approaching lanes and high vehicle inflows can be challenging). If such domain knowledge can be incorporated into the process and thereby can be confident that point MDP distribution does not contain many failure point MDPs, the experiment designers may utilize the random sampling without replacements (M2) method since it performs better under such setting. 

\section{Impact of Point MDP Distribution on Evaluations}

\begin{figure*}[h]
\centering
\begin{subfigure}{0.32\linewidth}
  \centering
  \includegraphics[width=\linewidth]{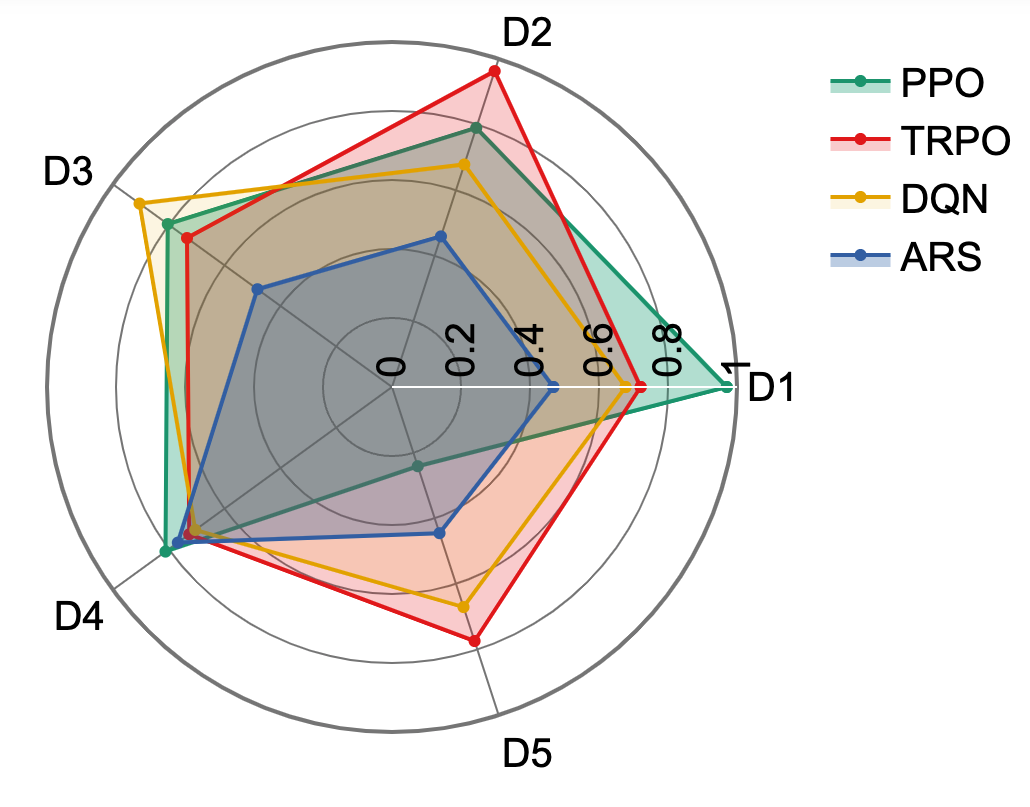}
  \caption{Cartpole (higher the scores the better)}
  \label{fig11a}
\end{subfigure}
\begin{subfigure}{0.32\linewidth}
  \centering
  \includegraphics[width=\linewidth]{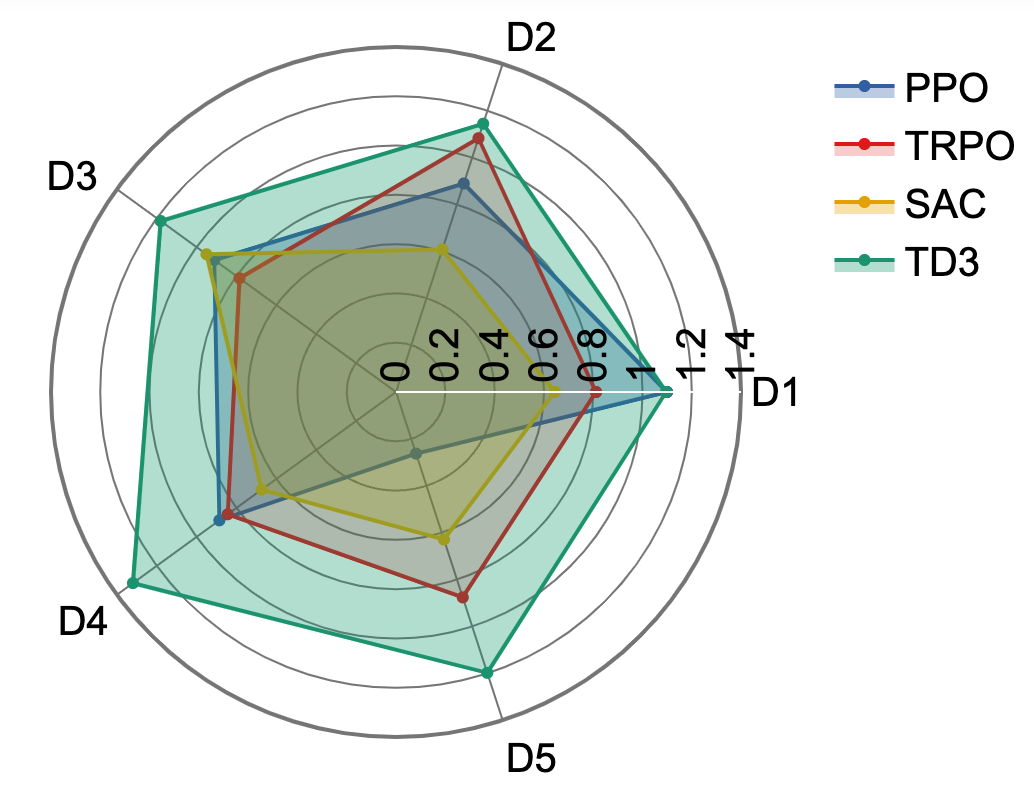}
  \caption{Pendulum (lower the scores the better)}
  \label{fig11b}
\end{subfigure}
\caption{Evaluations of cartpole and pendulum over a family of MDPs under different point-MDP distributions. $D_1 \dots D_5$ denotes the five random distributions and scores on the plots are normalized performance scores.}
\label{distribution-for-PE}
\end{figure*}

The choice of point MDP distribution can have a significant impact on the overall performance of an algorithm and, subsequently, on the ranking of a set of algorithms evaluated on the same task. Point MDP distribution essentially acts as a calibration distribution for evaluations according to some pre-defined criteria. For example, traffic engineers in New York may use a different point MDP distribution in comparison to what Salt Lake City engineers would use in traffic signal control benchmarking as they have different requirements. To demonstrate the impact of the point MDP distribution on overall evaluations, we perform a family of MDP-based performance evaluations using multiple point MDP distributions on cartpole and pendulum tasks in CARL benchmark suite~\cite{Benjamins2021CARLAB}. For cartpole, we create a family of MDPs that contain 576 point MDPs, and for the pendulum, we create a family of MDPs with 180 point MDPs. For illustration purposes, we randomly generate five point MDP distributions $D_1 \dots D_5$ and plot the overall performance of each algorithm under each distribution in Figure~\ref{distribution-for-PE}.

From Figure~\ref{distribution-for-PE}, it is clear that under both tasks, depending on the choice of underlying point-MDP distribution, the ranking of methods can significantly change. For example, in cartpole, under $D1$ distribution, PPO performs the best among the other methods, but under $D5$ distribution, PPO performs the worst, and TRPO performs way better than PPO. Similar results can be seen in the Pendulum task as well. In conclusion, it is evident that the choice of the point MDP distribution plays a major role in point MDP family-based evaluations and that it should be carefully set to fit the pre-defined requirements.

\section{Do Point MDP-based Evaluations Affect Standard DRL Evaluations Across Tasks? }

In this work, our primary focus is identifying and pointing out shortcomings of point MDPs-based evaluations when DRL methods are evaluated for algorithmic generalization within a task. However, we hypothesize such shortcomings due to point MDPs are not just isolated to this case but also can happen when DRL methods are evaluated for algorithmic generalization across tasks. The standard practice of evaluating general DRL methods is to consider a suite of tasks (one point MDP from each task) and evaluate the methods on all of them to report the robust performance of the method. In this section, we conduct experiments to provide insights that suggest that, even in this case, the choice of point MDPs can have an overall impact on how one would rank DRL methods based on this evaluation protocol. Therefore, we call for future work to further analyze this phenomenon.  

We used pendulum, half cheetah, and mountain car continuous control as a suite of tasks. We use the same point MDP families described in Table~\ref{drl-tasks-detailed-vv} for pendulum and half cheetah and generate the MDP family for mountain car by varying max speed, goal position, and power. We use four DRL methods: PPO, TRPO, SAC, and TD3. To mimic the standard evaluation procedure of evaluating DRL methods on one point MDP selected from each task, we generate all combinations of one point MDP from each task. Then we calculate the overall performance score for each DRL method evaluated on each point MDP combination. In Table~\ref{table:suite-of-tasks}, we report the percentage of the times each DRL algorithm obtained each rank.

\begin{table}[!h]
\begin{center}
\begin{tabular}{lcccc} \toprule
Method & Rank 1 & Rank 2 & Rank 3 & Rank 4\\ \midrule
PPO & 18.75\% & 29.54\% & 27.74\% & 23.97\%\\ \midrule
TRPO & 26.23\% & 24.99\% & 30.45\% & 18.33\%\\ \midrule
SAC & 42.56\% & 29.09\% & 18.50\% & 9.85\%\\ \midrule
TD3 & 12.46\% & 16.38\% & 23.31\% & 47.85\%\\ \midrule
Reported & SAC & TRPO & PPO & TD3 \\ \midrule
\end{tabular}
\caption{Percentage of the times each DRL algorithm obtained each rank}
\label{table:suite-of-tasks}
\end{center}
\end{table}

In Table~\ref{table:suite-of-tasks}, we see some interesting results. Despite what would otherwise conclude based on default point MDPs currently used in benchmark suits for the pendulum, half cheetah, and mountain car, we observe that the choice of the point MDP from each task matters a lot. In fact, different choices can lead to completely different ranking orders. We show that depending on which point MDP is used from each task, PPO is ranked first for 19\% times, TRPO for 26\%, SAC for 43\%, and TD3 for 12\%. For comparison, algorithms are ranked SAC, TRPO, PPO, and TD3 when trained on the default point MDPs. These findings provide insightful evidence that considering point MDP families may even benefit the evaluation practices for DRL across tasks because doing so may inform better selection of “default” point MDPs. We, therefore, call for future work in this direction to further analyze the phenomena in detail.

\end{document}